\documentclass{article}
     \usepackage[final]{neurips_data_2022}

\usepackage{natbib}
\usepackage[utf8]{inputenc} % allow utf-8 input
\usepackage[T1]{fontenc}    % use 8-bit T1 fonts
\usepackage{url}            % simple URL typesetting
\usepackage{booktabs}       % professional-quality tables
\usepackage{amsfonts}       % blackboard math symbols
\usepackage{nicefrac}       % compact symbols for 1/2, etc.
\usepackage{microtype}      % microtypography
\usepackage{xcolor}         % colors

\usepackage{graphicx}
\usepackage{wrapfig}
\usepackage{threeparttable}
\usepackage{caption}
\usepackage{subcaption}
\usepackage{multirow}
\usepackage{amsmath,amssymb}
\usepackage{siunitx}

\usepackage{xspace}
\usepackage{makecell}
\usepackage{pifont}% http://ctan.org/pkg/pifont
\usepackage[colorlinks=true, allcolors=blue]{hyperref} % hyperlinks
\usepackage{cleveref}
\usepackage[inline]{enumitem}

\usepackage[many]{tcolorbox}    	% for COLORED BOXES (tikz and xcolor included)

\definecolor{main}{HTML}{880808}    % setting main color to be used
\definecolor{sub}{HTML}{e97451}     % setting sub color to be used

\tcbset{
    sharp corners,
    colback = white,
    before skip = 0.2cm,    % add extra space before the box
    after skip = 0.5cm      % add extra space after the box
}

\newtcolorbox{boxH}{
    colback = sub, 
    colframe = main, 
    boxrule = 0pt, 
    leftrule = 6pt % left rule weight
}                           % setting global options for tcolorbox

\title{\textsc{AirfRANS}: High Fidelity Computational Fluid Dynamics Dataset for Approximating Reynolds-Averaged Navier–Stokes Solutions}
 \author{
     Florent Bonnet  \\
     Sorbonne Université, CNRS, ISIR \\
     Extrality \\
     Paris, France \\     
     \texttt{bonnet@isir.upmc.fr} \\
   \And
   Jocelyn Ahmed Mazari\\
   Extrality \\
   Paris, France \\
   \texttt{ahmed@extrality.ai} \\
   \And
   Paola Cinnella \\
   Sorbonne Université, Institut Jean Le Rond d’Alembert \\
   Paris, France \\
   \texttt{paola.cinnella@sorbonne-universite.fr} \\
   \And
   Patrick Gallinari \\
   Sorbonne Université, CNRS, ISIR \\
   Criteo AI Lab \\
   Paris, France \\
   \texttt{patrick.gallinari@sorbonne-universite.fr} \\
 }

\begin{document}

\maketitle

% \begin{abstract}
% Numerical simulations in Computational Fluid Dynamics (CFD) are crucial for modeling physical phenomena, which help design planes, rockets, wind turbines, etc. Numerical simulation in CFD is used to solve Navier-Stokes PDEs, which involve a complex dissipation process and need to be solved at all scales. However, fluid numerical simulations can take weeks to converge to an accurate solution. Recently, data-driven methods have been a surge to complement numerical solvers in reducing the computational cost. However, despite the fast-growing field of Deep Learning (DL) for physical processes, the corresponding real-world datasets are rare, and the existing ones are far from satisfying the physical requirements of real-world phenomena. In this work, we develop, \textsc{AirfRANS}, a 2-D graph-mesh dataset to study Reynolds-Averaged-Navier-Stokes (RANS) solutions over airfoils at a high Reynolds regime and angle of attack (AoA). We also introduce metrics on the stress forces at the surface of geometry to assess the capabilities of Geometric Deep Learning (GDL) models in accurately predicting the form of the boundary layer. Finally, we propose four Machine Learning tasks to study \textsc{AirfRANS} under different constraints for generalization considerations:  big and scarce data regime, Reynolds and AoA extrapolation.
% \end{abstract}

\begin{abstract}
 Surrogate models are necessary to optimize meaningful quantities in physical dynamics as their recursive numerical resolutions are often prohibitively expensive. It is mainly the case for fluid dynamics and the resolution of Navier–Stokes equations. However, despite the fast-growing field of data-driven models for physical systems, reference datasets representing real-world phenomena are lacking. In this work, we develop \textsc{AirfRANS}, a dataset for studying the two-dimensional incompressible steady-state Reynolds-Averaged Navier–Stokes equations over airfoils at a subsonic regime and for different angles of attacks. We also introduce metrics on the stress forces at the surface of geometries and visualization of boundary layers to assess the capabilities of models to accurately predict the meaningful information of the problem. Finally, we propose deep learning baselines on four machine learning tasks to study \textsc{AirfRANS} under different constraints for generalization considerations: big and scarce data regime, Reynolds number, and angle of attack extrapolation.

\end{abstract}

\section{Introduction}
Numerical simulations of physical dynamics are a consequent part of scientific research as it allows us to quantitatively study natural phenomena without requiring often complex and expensive experiments. Those dynamics are mainly governed by Partial Differential Equations (PDE) and are numerically solved with the help of discretization methods such as finite differences, finite elements, or finite volumes methods. Such techniques are accurate when used over sufficiently fine meshes but are often expensive in time and resources. Thus, the optimization of meaningful quantities with respect to the parameters of the studied dynamics is, most of the time, out of scope. In particular, the numerical resolution of Navier–Stokes equations for fluid dynamics analysis leads to computations that can last for thousands of CPU hours. Hence, the design of accurate surrogate models is at the core of engineering as they allow us to tackle the task of optimization via data-driven approaches. However, to be able to compare and validate such surrogate models we need datasets of reference and evaluation protocols. For physical systems, some efforts have already been done in this direction \cite{otness2021an,bonnet2022an} and our work is another contribution to those efforts. In \cite{bonnet2022an}, we developed the first version of this dataset to study Reynolds-Averaged Navier–Stokes (RANS) equations with Machine Learning (ML) models along with an appropriate evaluation protocol. In this paper, we propose an extension of this work by introducing a new high-fidelity version of the dataset. This high-fidelity version is built over finer meshes than the previous one which helps to fight numerical diffusion and allows us to recover more accurate fields and the trail of airfoils. Moreover, it allows us to accurately compute the force coefficients acting over geometries.

% We focus on the classical aerodynamics task of predicting the steady-state two-dimensional fields, and the force over airfoils in a subsonic regime \cite{Leishman1988ValidationOA,HerbertAcero2015AnEA,app11114845,aerospace8100275}. The ultimate goal of this task is to be able to find the best airfoil in terms of lift over drag ratio (see chapter .. of \cite{aero}) in addition to the associated velocity and pressure fields. It is already a non-trivial problem in Computational Fluid Dynamics (CFD) \cite{DEFRAEYE20102281,dyke_album_1982,JAMESON2001197}, where turbulence modeling\cite{Sagaut_2010,1997SciAm.276a..62M,spalart,LUMLEY1979123,CATALANO2003493} and mesh engineering \cite{Sosnowski2018TheIO,10.1115/1.4007649,Kanade2009RapidMF,Steijl2006SlidingMA,ZHAO2015177} are required to find accurate force coefficients. Moreover, different Machine Learning (ML) frameworks can be used to build surrogate models \cite{doi:10.1146/annurev-fluid-010719-060214,10.5555/3360217,en15041513,https://doi.org/10.48550/arxiv.2110.02083,Brunton2021DataDrivenAE,gin_lusch_brunton_kutz_2021,loiseau:hal-02398729}. Deep Learning (DL) \cite{imagenet} is among the successful candidates and has recently gained popularity \cite{thuerey2020deepFlowPred} for fluid simulation. Remarkably, the emerging field of Geometric Deep Learning (GDL) \cite{DBLP:journals/spm/BronsteinBLSV17} models allows us to achieve learning directly on unstructured point clouds and meshes \cite{pfaff2021learning} (non-Euclidean data), which makes them good candidates for our task.

We focus on the classical aerodynamics task of predicting the steady-state two-dimensional fields and the force acting over airfoils in a subsonic regime. The ultimate goal of this task is to be able to find the best airfoil in terms of lift over drag ratio (see chapter 1 of \cite{aero}) in addition to the associated velocity and pressure fields. It is already a non-trivial problem in Computational Fluid Dynamics (CFD) as turbulence is involved and mesh engineering is required to find accurate force coefficients. To accelerate the resolution process, different ML frameworks can be used to build surrogate models \cite{surrogate2020, surrogate2008}. Deep Learning (DL) is among the successful candidates and has recently gained popularity for fluid simulation \cite{thuerey2020deepFlowPred}. Moreover, the emerging field of Geometric Deep Learning (GDL) \cite{DBLP:journals/spm/BronsteinBLSV17} models allows us to achieve learning directly on unstructured data \cite{pfaff2021learning} which, in this particular case, allows us to compute accurately meaningful quantities at the surface of geometries.

In this work, we present a high fidelity aerodynamics dataset of RANS solutions around airfoils. In Section~\ref{sec:dataset} we present the RANS equations, the chosen design space for the airfoils generation, the meshes construction, and the dataset generation procedure. We also present the two force coefficients of interest, namely the drag and the lift coefficients. In Section~\ref{sec:setup} we introduce the different sub-tasks of the problem in addition to the evaluation protocol and the setup for our GDL baselines. In particular, the evaluation protocol contains metrics and visualizations for the force coefficient ranks and the accuracy of the surrogate models over boundary layers. We finally present, in Section~\ref{sec:results} the results of our baselines on the main sub-task and let the remaining ones in Appendix~\ref{ap:results}. All the values of the constant used in this work and the definition of dimensionless quantities are given in Appendix~\ref{ap:dimensionless}.

\section{Related Work}\label{sec:related_works}
Although several research directions are established to come up with efficient surrogate models to tackle physics problems, from physically guided methods \cite{de2018deep,Chen2020,dgm,pmlr-v80-long18a,brandstetter2022lie} to neural operators \cite{mgno,Kovachki2021NeuralOL,fno,lu2021learning} and Physics Informed Neural Networks (PINN) \cite{pinns}, the lack of standard benchmarking datasets, and common evaluation protocols impede making rigorous comparisons between the different families of methods for a given task. Benchmarking datasets and common evaluation protocols are shown to be the key components for making progress as it is observed in neighboring fields such as, for example, computer vision \cite{imagenet,Smaira2020ASN} and speech recognition \cite{pmlr-v48-amodei16}. Though, few physics-based datasets have been proposed  such as: 1D Burger’s equation and  2D Darcy flow PDE \cite{mgno}, structural mechanics \cite{pfaff2021learning}, incompressible fluid in vorticity form \cite{fno}, reaction-diffusion, wave-equations and damped pendulum \cite{yin:hal-03137025}, heat transfer equation \cite{DBLP:journals/corr/abs-2010-02011}, Lorenz system \cite{dubois:hal-02475962}. More recently, few standard benchmarks datasets on complex chemical and physical systems \cite{atom3d,du2021graphgt,freitas2021a,freeman2021brax,gilpin2021chaos} have been proposed. More interestingly, \cite{otness2021an} suggests a framework to study a set of representative physics problems with appropriate evaluation protocols, namely a single oscillating spring, a one-dimensional linear wave equation, a Navier–Stokes flow problem, as well as a mesh of damped springs. We follow those efforts by proposing a dataset on a steady-state aerodynamics task with dynamics that can be found in realistic flight scenarios. We also focus the validation of models on meaningful parts of the dynamic instead of only regarding the mean square loss of regressed fields.

Most of the works proposed in the literature to tackle tasks represented by Navier–Stokes equations are grid-based approaches \cite{um2020sol,thuerey2020deepFlowPred,mohan2020embedding,wandel2021learning,10.1145/3392717.3392772,gupta2021multiwaveletbased, tfnet} which rely on Convolutional Neural Networks (CNN). Other architectures such as Fourier Neural Operator \cite{fno} act in the frequency domain and require a regular grid to perform a Fast Fourier Transform of the input data. Those models are not designed to directly operate on unstructured data like CFD meshes, resulting in inaccurate predictions of the physical fields at the surface of geometries. However, recent progress in learning on unstructured data \cite{DBLP:journals/spm/BronsteinBLSV17} has enabled learning on graphs and manifolds by designing geometrical inductive bias in DL \cite{1555942,4700287,Li2016GatedGS,Kipf:2016tc,pmlr-v80-sanchez-gonzalez18a}. This framework is particularly useful to achieve learning on arbitrary shapes and frees us from the constraint of data voxelization as required by CNN. One successful attempt at learning Navier–Stokes (or RANS) equations with Graph Neural Networks (GNN) can be found in \cite{pfaff2021learning}. Finally, let us emphasize that PINN, as defined in \cite{pinns} can act on unstructured data but are not suited for surrogate modeling as they are designed to solve one and only one PDE.

\section{Dataset Presentation} \label{sec:dataset}
\paragraph{Design-oriented dataset.} This dataset is mainly motivated by a realistic shape optimization problem. We choose a classical aerodynamics problem for this purpose: airfoil design optimization. The goal is to accurately predict force coefficients in addition to the different fields of the fluid in a subsonic flight regime with a reduced quantity of data as is often the case in practice. The design space is chosen from NASA's early works on airfoils via the 4 and 5 digits series \cite{naca} as they are easy to handle and already rich families of shapes.

We aim to resolve the air dynamic around a two-dimensional (2D) airfoil in a steady-state subsonic regime at sea level and \SI{298.15}{\kelvin}. More precisely, we study airflows at a Reynolds number between 2 and 6 million, which leads to turbulent behavior of the fluid. It corresponds to a Mach number smaller than 0.3 which allows us to assume incompressible flow behavior (see chapter 8 of \cite{aero}), and a velocity greater than \SI{30}{\meter\per\second} which is a reasonable lower bound in subsonic flight conditions. Moreover, as the flow is turbulent in certain areas, we use Reynolds-Averaged-Simulations (RAS) with a sufficiently high number of cells in our meshes to accurately compute the force acting over airfoils. This method solves the RANS equations widely used in Computational Fluid Dynamics (CFD) to control the numerical complexity of the resolutions.

We rely on the Turbulence Modeling Resource (TMR) of the Langley Research Center of the National Aeronautics and Space Administration (NASA) \cite{TMR, NACA0012-1, NACA0012-2, NACA4412} to generate our dataset. In what follows, we present the different steps to build the dataset, and we define the relevant physical quantities of the problem.

\paragraph{Reynolds-Averaged Navier–Stokes equations.} At a high Reynolds number, untidy patterns emerge in fluid flows; we call this phenomenon turbulence. In CFD, turbulence resolution is a crucial problem as it implies transient simulations on prohibitively fine meshes most of the time. Different strategies have been developed to tackle this problem, one of them being RAS. In RAS, we solve mean-field equations similar to Navier–Stokes equations but with an effective viscosity representing the diffusion added through turbulent processes. Those equations, called incompressible RANS equations, are given by:
\begin{align}
    \partial_i \Bar{u}_i &= 0 \\
    \partial_j(\Bar{u}_i\Bar{u}_j) &= -\partial_i\left(\frac{\Bar{p}}{\rho}\right) + (\nu + \nu_t)\partial^2_{jj}\Bar{u}_i, \quad i\in\{1,2\}
\end{align}
where $\Bar{\cdot}$ denotes an ensemble-averaged quantity, $\partial_i$ the partial derivative with respect to the $i^{th}$ spatial components, $u$ the fluid velocity, $p$ an effective pressure, $\rho$ the fluid specific mass, $\nu$ the fluid kinematic viscosity, $\nu_t$ the fluid kinematic turbulent viscosity and where we used the Einstein summation convention over repeated indices. Often, in the incompressible case, the effective pressure is replaced by the reduced pressure, abusively denoted by the same symbol via the transformation $p \rightarrow p/\rho$, which allows us to write RANS equations without explicit dependence on $\rho$. From now on, we will only discuss in terms of the reduced pressure. Finally, the dynamics of the turbulent viscosity is driven by a set of supplementary equations. In this work, we choose to use the well-known $k-\omega$ SST turbulence model \cite{SST} which is well suited for aerodynamics problems. Details on the RANS equations, the definition of the different quantities, the ensemble average, and the choice of the turbulent model are given in Appendix~\ref{ap:RANS}.

\paragraph{Airfoil design space.} In the first half of the twentieth century, teams of the National Advisory Committee for Aeronautics (NACA) worked on several airfoil families. Two of them called the 4 and 5 digits series, are entirely parameterized and allow us to generate a broad spectrum of airfoils quickly. Both series define a camber line and an envelope around this camber line. An airfoil of the 4 digits one is defined by a sequence MPXX where M is the maximum ordinate of the camber line in hundredth of chords\footnote{One chord is the characteristic length of the airfoil, in our case \SI{1}{\meter}}, P is the position of this maximum from the leading edge in tenth of chords and XX the maximum thickness in hundredth of chords. For the 5 digits series, each airfoil is defined by a sequence LPQXX. Digits L and P define in a more sophisticated manner than the 4 digits sequence the maximum camber of the camber line, Q is a boolean that switches between a single-cambered airfoil and a double-cambered one which allows in the latter case to achieve a theoretical pitching moment of 0. The last two digits XX have the same definition as in the 4 digits case. 

Each simulation is first defined by an airfoil drawn in the 4 and 5 digits series families. The sampling strategy in those two series is given in Table~\ref{tab:sample-NACA}. In our previous work \cite{bonnet2022an}, we chose to use the UIUC Airfoil Database \cite{UIUC} to build the dataset but we decide here to restrict our airfoil design space to the NACA 4 and 5 digits series. Those series are already rich families of airfoils that have been widely used historically and they are easier to handle for the automation of the mesh generation due to their explicit parametrization. Moreover, in the 4 digits series, we choose to sample the parameter P between 0 and 7 and to set the drawn parameters in the interval $(0, 1.5]$ to 0. We motivate this choice as airfoils with P in the range $(0, 1.5]$ have their maximum camber close to the trailing edge which can lead to unusable airfoils.

Examples of different airfoils, details on the generation of such airfoils, and empirical statistics of the drawn parameters are given in Appendix~\ref{ap:airfoil}.   
% \begin{figure}
%   \centering
%   \begin{subfigure}{0.49\textwidth}
%       \centering
%       \includegraphics[width = \linewidth]{Dataset/naca_(2.271, 1.839, 8.599).png}
%   \end{subfigure}
%   \begin{subfigure}{0.49\textwidth}
%       \centering
%       \includegraphics[width = \linewidth]{Dataset/naca_(3.408, 6.127, 10.424).png}
%   \end{subfigure}
  
%   \begin{subfigure}{0.49\textwidth}
%       \centering
%       \includegraphics[width = \linewidth]{Dataset/naca_(5.916, 1.801, 11.06).png}
%   \end{subfigure}
%   \begin{subfigure}{0.49\textwidth}
%       \centering
%       \includegraphics[width = \linewidth]{Dataset/naca_(6.066, 2.446, 14.67).png}
%   \end{subfigure}
  
%   \begin{subfigure}{0.49\textwidth}
%       \centering
%       \includegraphics[width = \linewidth]{Dataset/naca_(1.505, 6.78, 0.0, 10.288).png}
%   \end{subfigure}
%   \begin{subfigure}{0.49\textwidth}
%       \centering
%       \includegraphics[width = \linewidth]{Dataset/naca_(3.729, 6.903, 1.0, 10.309).png}
%   \end{subfigure}
  
%   \begin{subfigure}{0.49\textwidth}
%       \centering
%       \includegraphics[width = \linewidth]{Dataset/naca_(1.548, 7.354, 0.0, 16.511).png}
%   \end{subfigure}
%   \begin{subfigure}{0.49\textwidth}
%       \centering
%       \includegraphics[width = \linewidth]{Dataset/naca_(2.341, 6.162, 1.0, 19.459).png}
%   \end{subfigure}
%   \caption{Example of NACA airfoils. Digits are given in the form (M, P, XX) and (L, P, Q, XX) for the four and five-digits series, respectively. The $x$ and $y$ coordinates are given in unit of chord $c$.}
%   \label{fig:NACA_airfoil}
% \end{figure}

\begin{table}
  \caption{Sampling strategy for generating airfoils from the 4 and 5 digits series. An interval or a discrete set means that the sampling is uniform over this set. In the 4 digits case, the sampling for P has been a uniform sampling on the interval $[0, 7]$, and all the samples smaller than 1.5 have been set to 0 to get rid of geometries that have their maximum camber too close from the leading edge.}
  \label{tab:sample-NACA}
  \centering
  \begin{tabular}{ccc}
    \toprule
    \multicolumn{3}{c}{4-digits}                  \\
    \midrule
    M & P & XX \\
    \midrule
    $[0, 7]$ & $\{0\}\bigcup [1.5, 7]$ & $[5, 20]$      \\
    \bottomrule
  \end{tabular} \hspace{1cm}
  \begin{tabular}{cccc}
    \toprule
    \multicolumn{4}{c}{5-digits}                    \\
    \midrule
    L & P & Q & XX \\
    \midrule
    $[0, 4]$ & $[3, 8]$ & $\{0, 1\}$ & $[5, 20]$      \\
    \bottomrule
  \end{tabular}
\end{table}

\begin{figure}
  \centering
  \begin{subfigure}{0.49\textwidth}
      \centering
      \includegraphics[width = \linewidth]{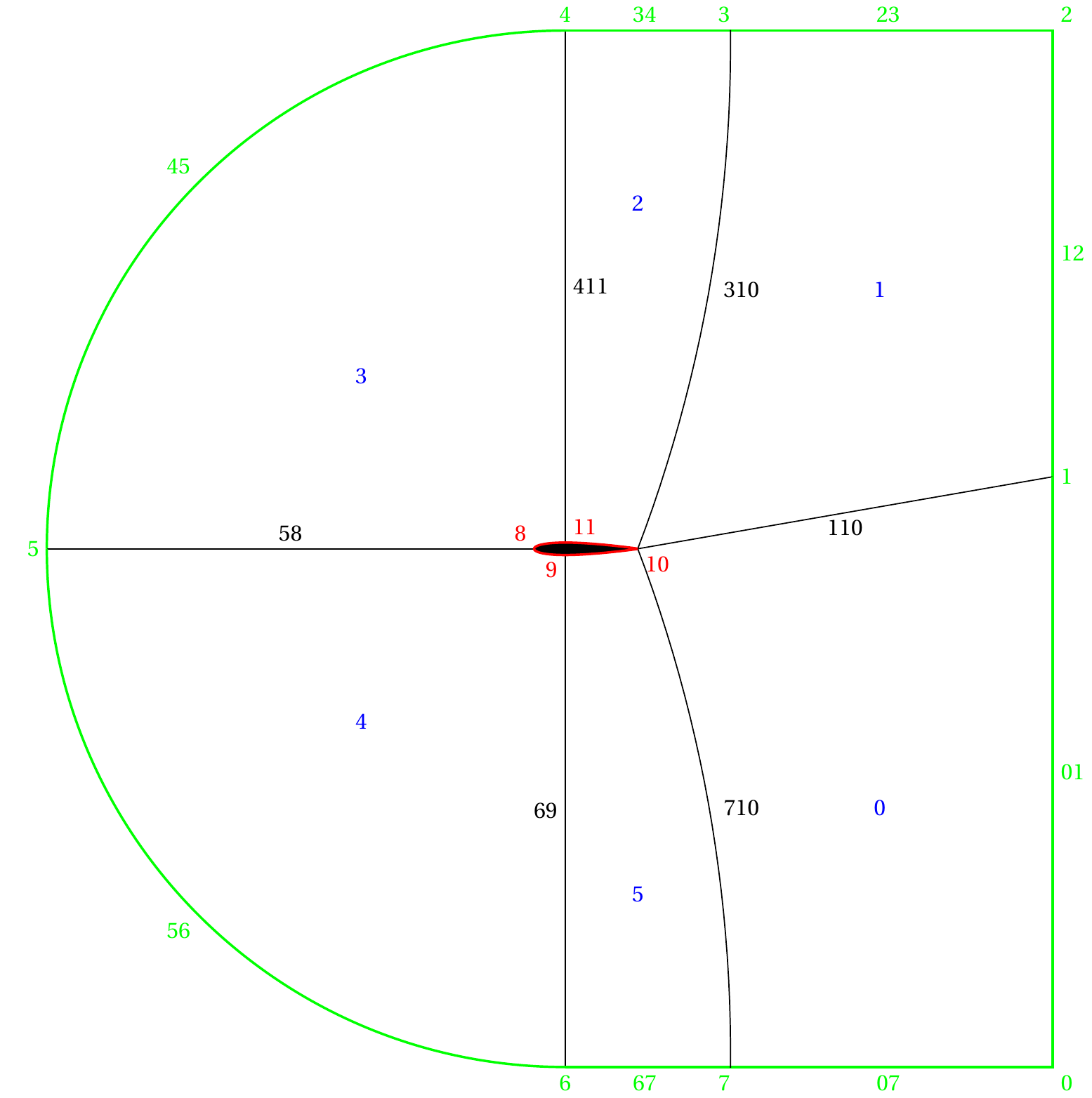}
      \caption{}
   \end{subfigure}
   \begin{subfigure}{0.49\textwidth}
      \centering
      \includegraphics[width = \linewidth]{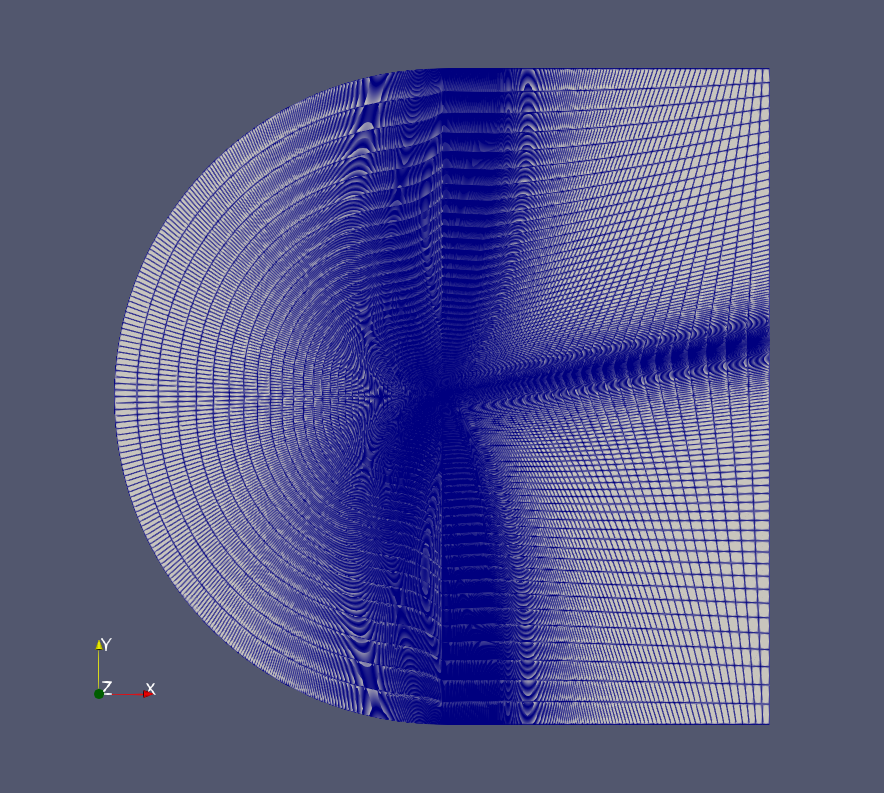}
      \caption{}
  \end{subfigure}
  \caption{Example of mesh for the NACA 0012 at an angle of attack of \SI{10}{\degree}. (a) Scheme of the multi-block mesh. Point number 1 moves following the angle of attack; all other points are fixed. The contour in red (airfoil) and green (freestream) are the domain's boundaries. (b) The entire domain of a ready-to-use mesh.}
  \label{fig:mesh_scheme}
\end{figure}

\paragraph{Mesh generation.} As airfoils are pretty simple geometries, we use the multi-block hexahedral mesh generator \emph{blockMesh} from OpenFOAM v2112 \cite{OpenFOAM} to mesh our shapes. We build a C-Grid mesh for each airfoil, mimicking the mesh developed by NASA for the NACA 0012 and 4412 cases \cite{TMR}. Boundaries are at 200 chords of the airfoil to reduce the impact of boundary conditions on simulations. In Figure~\ref{fig:mesh_scheme}, we show the different block definitions with an example of a ready-to-use mesh and a final mesh on a classical airfoil. As we aim for accuracy in the computation of the global forces over the airfoil surface, such as the wall shear stresses, we mesh the boundary layer such that the first cells of the surface are of height \SI{2}{\micro\meter} leading to a $y^{+}$ of around 1 in the worst case of our design space. This leads to meshes from \SI{250000}{} to \SI{300000}{} cells. All the technical details and definitions of the meshing procedure are given in Appendix~\ref{ap:meshing}.

\begin{figure}
  \centering
  \begin{subfigure}{0.49\textwidth}
      \centering
      \includegraphics[width = \linewidth]{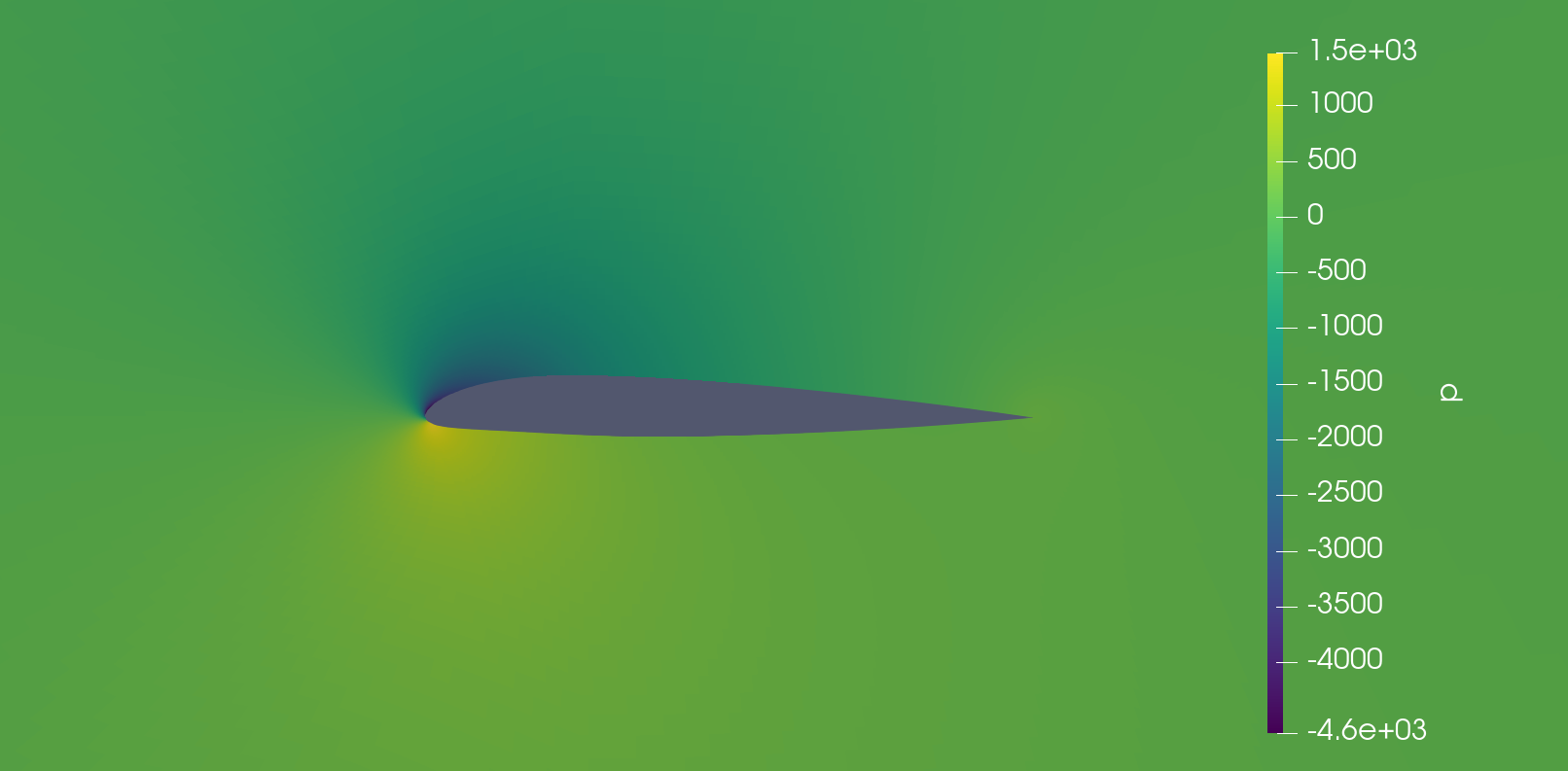}
   \end{subfigure}
   \begin{subfigure}{0.49\textwidth}
      \centering
      \includegraphics[width = \linewidth]{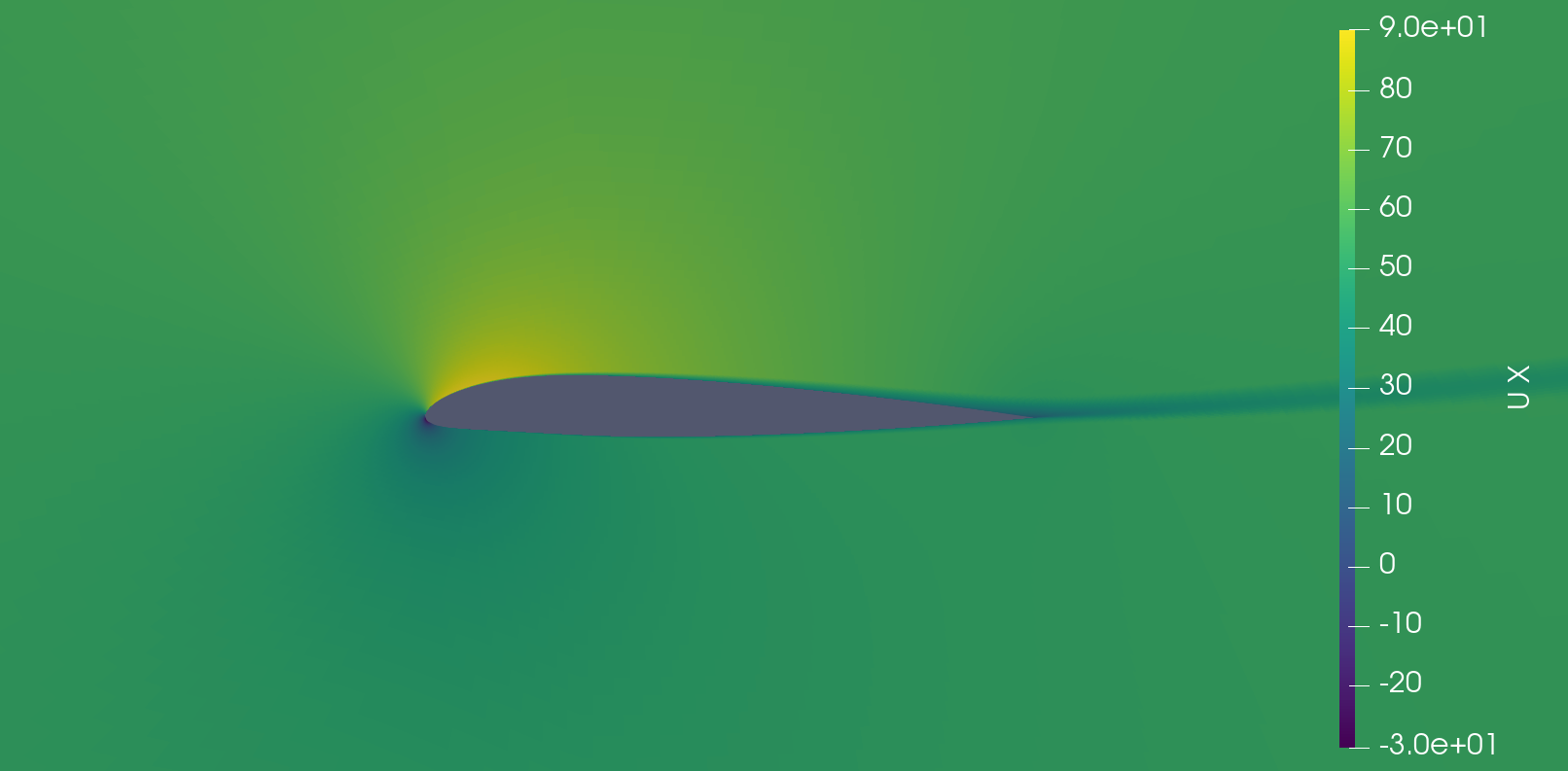}
  \end{subfigure}
  \caption{Example of a pressure (left) and $x$-component of the velocity (right) fields around a NACA (2.123, 3.832, 1, 9.902) at a velocity of \SI{54.238}{\meter\per\second} and at an angle of attack of \SI{7.911}{\degree}.}
  \label{fig:NACA_field}
\end{figure}

\paragraph{Dataset generation.} For the generation of the dataset, we run \SI{1000}{} simulations, each defined by an airfoil, a Reynolds number, and an angle of attack. We choose to only run \SI{1000}{} simulations as one of the goals of this dataset is to be close to real-world settings, \emph{i.e.} limited quantity of data. The airfoil is sampled from the distribution given in Table~\ref{tab:sample-NACA}. We motivate the design space of the initial conditions to reproduce the panel of flight conditions encountered in subsonic flights. We stop at a Mach number of 0.3 (Reynolds number of roughly 6 million) to keep the incompressible assumption valid and we start at a Reynolds number of 2 million as it is a correct lower bound of flight velocity (around 60 knots). The lower bound for angles of attack, \SI{-5}{\degree}, is chosen such as cambered airfoils have a lift coefficient of roughly 0 and the upper bound, \SI{15}{\degree}, is chosen to prevent stall and unsteady patterns in the trail of airfoils. Those ranges are tighter than the one chosen in our previous work \cite{bonnet2022an} but better represent the classical ranges of velocity and angle of attack encountered in subsonic flight conditions. We then run the simulations with the help of the steady-state RANS solver \emph{simpleFOAM} via the SIMPLEC algorithm \cite{SIMPLE, SIMPLEC} and with the $k-\omega$ SST turbulence model \cite{SST} until convergence of drag and lift coefficients. Simulations are done on 16 CPU cores of an AMD Ryzen™ Threadripper™ 3960X. Figure~\ref{fig:NACA_field} shows a near view of the pressure and $x$-component of the velocity fields. Boundary conditions types and values for each simulation are given in Appendix~\ref{ap:boundary}.

\paragraph{Force coefficients.} One of the important quantities when simulating the fluid flow around a geometry is the force acting on it (see chapter 4 of \cite{aero}). This force is made by the contribution of the pressure and the viscous stresses at the surface of the geometry (called wall shear stress). The force component collinear to the free-stream velocity is called the drag $D$, and the one orthogonal to the free-stream velocity is the lift $L$. If divided by $q_\infty := \rho U_\infty^2 A/2$, where $\rho$ is the fluid specific mass, $U_\infty$ the inlet velocity and $A$ the characteristic area of the geometry (in our case we take the chord as the 1D surface characteristic "area", \emph{i.e.} $A = \SI{1}{\square\meter}$), those components give the dimensionless drag coefficient $C_D := D/q_\infty$ and lift coefficient $C_L := L/q_\infty$. Other quantities such as the pitching moment can also be computed with force acting on the body, but we will only focus on the drag and lift coefficients in this work. To compute the wall shear stress, we need to compute the velocity gradient at the surface of the airfoil. We compute it discretely with the help of the \emph{gradient filter} in ParaView \cite{paraview} over the mesh.

\section{Benchmarking Setup}\label{sec:setup}
The machine learning task consists in predicting the different spatial fields such as the mean-field velocity and reduced pressure $\Bar{u}$ and $\Bar{p}$ and the force acting over airfoils. The turbulent viscosity $\nu_t$ is not mandatory in the regression task as we do not need it to compute the force over an airfoil\footnote{the boundary condition of the turbulent viscosity on the airfoil is 0 in our case}. Still, it can give insights into the local intensity of turbulence in the volume. Regarding the force coefficients, and more precisely, the drag and lift coefficients, from a design process standpoint, we are more interested in the rank correlation with the true values than with the Mean Squared Error (MSE). Indeed, if the rank of the coefficients is accurately approximated, the optimization process will lead to the same airfoil. We can also add linear correlation plots for qualitative prediction information. Moreover, we are dealing in this dataset with hexahedral meshes with more than \SI{250,000}{} cells which represent roughly the same number of nodes. These are already big meshes for 2D simulations but it is nothing compared to 3D cases where the number of cells can be of tens of million. To train a model on such simulations, we need to find a way to reduce the numerical complexity of the problem. Cropping close to the geometry is one way to do so, but as most cells lie in the vicinity of the geometry, it is not sufficient. Also, as we want to infer the force acting over the airfoil accurately, we cannot treat the cropped simulation as an image and work on a sub-sampled regular grid.

In this work, we choose the strategy which consists of regressing the four fields $\Bar{u}_x$, $\Bar{u}_y$, $\Bar{p}$ and $\nu_t$ and compute the wall shear stress and the associated forces as post-processing to follow the form of RANS equations. We present a method to take into account the remarks mentioned above. Finally, we use the 1000 simulations to build four different setups:
\begin{itemize}
    \item \emph{Full data regime:} 800 samples for the training set and 200 for the test set
    \item \emph{Scarce data regime:} 200 samples for the training set and the same test set as in the full data regime
    \item \emph{Reynolds extrapolation:} the training set is composed of the samples with a Reynolds of three to five million, and the test set is formed by the samples with a Reynolds of two to three and five to six million
    \item \emph{Angle of attack extrapolation:} the training set is composed by the samples with an angle of attack between \SI{-2.5}{\degree} and \SI{12.5}{\degree} and the test set is composed by the samples with an angle of attack from \SI{-5}{\degree} to \SI{-2.5}{\degree} and \SI{12.5}{\degree} to \SI{15}{\degree}.
\end{itemize}

%\textbf{Ahmed: add data statistics and compare in appendix the previous dataset set and the current one and say for the sake of simplicity we follow the same processing as figure 2 in workshop paper.}
% As we said earlier, we are dealing in this dataset with hexahedral meshes with more than \SI{250000}{} cells which represent more than \SI{500000}{} nodes. This are already big meshes for 2D simulations but it is nothing compare to the often tens of millions of cells required to correctly treat a real-life 3D case. In order to be able to train model on such simulations, we need to find a way to reduce the numerical complexity of the problem. Cropping close to the geometry is one way to do so but as the majority of cells lie in the vicinity of the geometry it is not sufficient. Moreover, as we want to accurately infer the force on the airfoil, we cannot treat the cropped simulation as an image and work on a sub-sampled regular grid. Hence, we create a sampling procedure described in the following to control the numerical complexity without giving up on the force inference over airfoils.
%Short intro. 1/4 page

\paragraph{Preprocessing.} As we do not need the far-field to get rid of the boundary conditions impact on the simulation as in CFD, we crop all the simulations to a rectangle of size $[-2, 4]\times[-1.5, 1.5]$ meters. It allows us to limit our point clouds' size and make the network focus on the interesting part of the simulations. Moreover, data normalization is important in DL to make the optimization process easier or feasible. We use normalization with the means and the standard deviations of the training set field components.
% We transform velocities $u$, pressures $p$, turbulent viscosities $\nu_t$ and positions $x$ into dimensionless quantities following:
% \begin{align}
%     u\rightarrow \frac{u}{\Tilde{U}_\infty}, \quad p\rightarrow \frac{p}{\frac{1}{2}\Tilde{U}_\infty^2}, \quad \nu_t \rightarrow \frac{\nu_t}{\Tilde{U}_\infty c}, \quad x\rightarrow \frac{x}{L}
% \end{align}
% where $\Tilde{U}_\infty$ is the mean inlet velocity magnitude of the training set, $c$ the airfoil chord (\emph{i.e.} \SI{1}{\meter}) and $L$ the characteristic length of the domain (\emph{i.e.} \SI{6}{\meter}). In Appendix (ref), we present the normalization process, and we compare the scores of the models presented later trained with dimensionless quantities or not.
% Check dimensionless or not quantities. Cropping, normalization. 1/2 page

\paragraph{Loss, sampling, and graph construction.} At the end of the cropping procedure, we still have to deal with roughly \SI{150,000}{} cells in each simulation giving about the same number of nodes. To train a model on such simulations, we need to find a way to reduce the numerical complexity even more. 

To handle this numerical complexity, at each epoch, we choose to sample uniformly on the cropped mesh \SI{32.000}{} nodes and, when necessary, reconstruct a radius graph of radii \SI{5}{\centi\meter} with a maximum number of neighbors of 64\footnote{which means that we connect nodes only very locally compared to the characteristic length of \SI{1}{\meter} of airfoils}. This approach has several advantages, it allows us to directly control the numerical complexity with the number of sub-sampled nodes, the radii of the graph, and the maximum number of neighbors inside it. Moreover, during the inference, it is straightforward to infer fields on each node of the initial mesh by making multiple forward passes with different sub-sampling until every node has been seen and averaging the outputs on the nodes that have been seen multiple times. It allows us to compute the velocity gradient on the airfoil with the help of the initial mesh and ParaView's pythonic interface PyVista \cite{pyvista} to be able to compare with the surface force targets. However, one downside effect of the method is that the mesh densities bias the learning procedure, and the learned models may not generalize well to different types of meshes. In this dataset, we are not facing this problem as all the meshes are generated via the same procedure. Additionally, we can not sample independently on the airfoil and on the volume to bias models to be more accurate on the airfoil as the force highly depends on the pressure field at the surface. 
%An attempt at a uniform sub-sampling over the volume and the airfoil is presented in Appendix (ref). In the following, we use this mesh-biased sub-sampling method. 
Finally, The loss $\mathcal{L}$ used in this work is the sum of two terms, a loss over the volume $\mathcal{L}_{\mathcal{V}}$ and a loss over the surface $\mathcal{L}_{\mathcal{S}}$:
\begin{equation}\label{eq:global_loss}
\centering
    \mathcal{L}:={\underbrace{\frac{1}{|\mathcal{V}|}\sum_{i\in \mathcal{V}} \|f_\theta(x_i) - y_i\|_2^2}_{\text{$\mathcal{L}_\mathcal{V}$}}}+  \lambda{\underbrace{\frac{1}{|\mathcal{S}|}\sum_{i\in \mathcal{S}} \|f_\theta(x_i) - y_i\|_2^2}_{\text{$\mathcal{L}_\mathcal{S}$}}}
     %&:=% \mathcal{L}_\mathcal{V} + \lambda\mathcal{L}_\mathcal{S}
\end{equation}
where $\mathcal{V}$, $\mathcal{S}$ are respectively the set of the indices of the nodes that lie in the volume and on the airfoil, $x_i \in \mathbb{R}^{7}$ is the input at node $i$ containing the spatial coordinates, the inlet velocity, the Euclidean distance function between the node and the airfoil, and the unit surface outward-pointing normal for points on the airfoil (filled with zeroes otherwise). The targets $y_i \in \mathbb{R}^{4}$ at node $i$ contain the velocity, the pressure, and the turbulent kinematic viscosity at this node. And $f_\theta$ represents the model used. The coefficient $\lambda$ is used to balance the weight of the error at the surface of the geometry and over the volume\footnote{In this work $\lambda$ is set to 1.}. We have to emphasize that this loss is not necessarily a good proxy when it comes, for instance, to infer the wall shear stress accurately or ensuring that the inferred fields satisfy the RANS equations.

\begin{boxH}
\textbf{Disclaimer, May 26, 2023 update.} Due to an implementation error in the training phase, the results given in the main part of the paper have been generated with models trained via a classical MSE loss (without the separation between the points over the surface and the volume). Results generated with models trained with the loss presented above are given in Appendix \ref{ap:corrected_results}. Qualitative insights given in the main paper are still valid.
\end{boxH}

\paragraph{Metrics and visualizations.} One of the challenges of this dataset is to build models that manage to predict the form of the boundary layer accurately. To evaluate the performance of the models, we define qualitative and quantitative metrics.

To qualitatively check this accuracy, we propose to plot the components of the velocity and the turbulent viscosity (if regressed) in the boundary layer of airfoils at different chord lengths. Also, we propose to check the accuracy of the prediction for the pressure and skin friction coefficients on the airfoil as they carry important information on the wing's behavior in flight conditions. Finally, we propose to plot the predicted force coefficients with respect to the true coefficients which gives us qualitative information about the correlation between both variables. 
% Figure (ref) shows an example of such qualitative checkups.

In terms of quantitative metrics, we use the MSE for each field on the volume and over the airfoil to measure the accuracy of our models. Moreover, we compute the mean and standard deviation of the relative error on the drag and lift coefficients. Finally, we compute Spearman's rank correlation coefficient between the true and predicted force coefficients. From a design process point of view, the last coefficient is the most crucial quantity to maximize as it quantifies the monotonic correlation between the true and predicted force coefficients. If this coefficient is close to 1, we can expect our model to be able to find the best airfoil maximizing or minimizing the chosen force coefficient even if the inferred value is not close to the true value.

% Pressure, skin friction coefficients, MSE on it + visualization. Relative error on drag and lift coefficient. drag/drag, lift/lift and drag/lift plot 1 page

\section{Benchmarking Results}\label{sec:results}
To propose baselines for the problem, we train three standard GDL models and a Multi-Layer Perceptron (MLP) in the full data regime. The associated results for the three other tasks are given in Appendix~\ref{ap:results}. Each model is preceded by an encoder and followed by a decoder, both defined by a MLP and trained together with the model. We follow the setup defined in the previous section for the training and testing procedures. Each model is trained 5 times to compute a mean and a standard deviation for the different metrics. For each metric, we bold the best-performing method. We choose as baselines a GraphSAGE \cite{gsage}, a PointNet \cite{qi2016pointnet}, a Graph U-Net \cite{gunet} and a MLP. Those baselines have been chosen as they access different types of information. The MLP only has access to the features of the nodes, the GraphSAGE has in addition access to local neighborhood information, the PointNet conditioned a deep MLP with global features, and the Graph U-Net access from local to global neighborhood information via its multi-scale architecture. Models are trained in the same conditions and the details of architectures and hyperparameters can be found in Appendix~\ref{ap:models}.

\begin{figure}
  \centering
  \includegraphics[width = \linewidth]{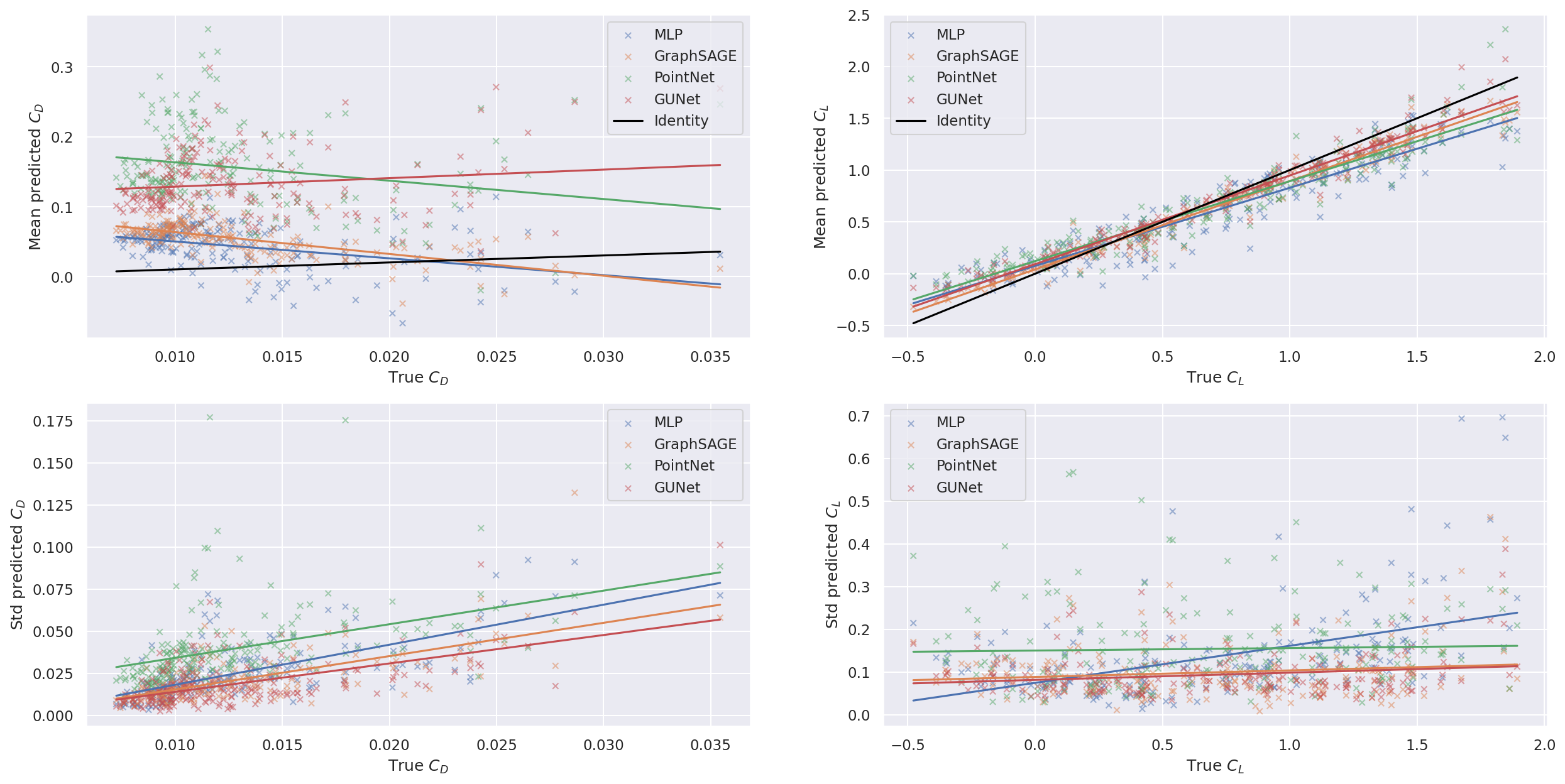}
  \caption{Predicted drag (left) and lift (right) coefficients with respect to the true ones. The mean (top) and standard deviation (bottom) of each point on the five copies of the trained models are separated for sake of readability. Linear regression is done for each point cloud to highlight linear trends. On the top plots, the Identity graph is given in black for comparison.}
  \label{fig:coefs}
\end{figure}

In Table~\ref{tab:MSE}, we give the MSE over the volume and at the surface of airfoils for the different regressed fields. In Table~\ref{tab:spear} we give the mean relative errors on the force coefficient and the Spearman's rank correlation coefficient. In Table~\ref{tab:time} we compare the computational time to run a simulation, train a model and infer on a new example. In Figure~\ref{fig:coefs} we plot the predicted force coefficients with respect to the true coefficients. Plots of the velocity and turbulent viscosity profiles in the boundary layer and surface coefficients for randomly chosen test geometries are given in Appendix~\ref{ap:results}. 

The Graph U-Net model significantly outperforms other models for the pressure at the surface, this correlates with its performance on the relative error and Spearman's correlation for the lift coefficient. However, it struggles to learn the velocity field compared to local models such as GraphSAGE and MLP. The GraphSAGE model seems to be a good trade-off, in this setting, between complexity and performance as it achieves almost equivalent performance with the Graph U-Net, is almost twenty times faster to call, and has half of the number of parameters of the Graph U-Net.

\begin{table}
  \caption{Mean squared error on the different normalized fields for a MLP and standard GDL baselines on the full data regime test set. Only the reduced pressure is given on the surface as the other quantities are null via the boundary conditions. Those quantities are directly regressed by the models.}
  \label{tab:MSE}
  \centering
  \begin{tabular}{cccccc}
    \toprule
    Model & \multicolumn{4}{c}{Volume} & Surface  \\
    & $\Bar{u}_x$ ($\times 10^{-2}$) & $\Bar{u}_y$ ($\times 10^{-2}$) & $\Bar{p}$ ($\times 10^{-2}$) & $\nu_t$ ($\times 10^{-2}$) & $\Bar{p}$ ($\times 10^{-1}$) \\
    \midrule
    MLP & 0.95 $\pm$ 0.06 & \textbf{0.98 $\pm$ 0.17} & 0.74 $\pm$ 0.13 & 1.90 $\pm$ 0.10 & 1.13 $\pm$ 0.14 \\
    GraphSAGE & \textbf{0.83 $\pm$ 0.01} & 0.99 $\pm$ 0.05 & \textbf{0.66 $\pm$ 0.05} & 1.60 $\pm$ 0.21 & 0.66 $\pm$ 0.10 \\
    PointNet & 3.50 $\pm$ 1.04 & 3.64 $\pm$ 1.26 & 1.15 $\pm$ 0.23 & 2.92 $\pm$ 0.48 & 0.93 $\pm$ 0.26 \\
    Graph U-Net & 1.52 $\pm$ 0.34 & 2.03 $\pm$ 0.39 & 0.66 $\pm$ 0.08 & \textbf{1.46 $\pm$ 0.14} & \textbf{0.39 $\pm$ 0.07} \\
    \bottomrule
  \end{tabular}
\end{table}

% \begin{table}
%   \caption{Relative errors for the unnormalized predicted pressure $\Bar{p}$ and components of the wall shear stress $\tau$. The wall shear stress is computed as a post processing from the unnormalized regressed velocity.}
%   \label{tab:rel_err}
%   \centering
%   \begin{tabular}{cccc}
%     \toprule
%     Model & \multicolumn{3}{c}{Relative error} \\
%     & $\Bar{p}$ & $\tau_x$ & $\tau_y$ \\
%     \midrule
%     MLP & \textbf{3.92}$\pm$\textbf{0.53} & 0.65$\pm$0.15 & \textbf{3.92}$\pm$\textbf{0.54}  \\
%     GraphSAGE & \textbf{3.92}$\pm$\textbf{0.53} & 0.65$\pm$0.15 & 4.44$\pm$0.64 \\
%     PointNet & \textbf{3.92}$\pm$\textbf{0.53} & 0.65$\pm$0.16 & 12.9$\pm$2.1 \\
%     Graph U-Net & \textbf{3.92}$\pm$\textbf{0.53} & 0.54$\pm$0.56 & 10.3$\pm$0.8 \\
%     \bottomrule
%   \end{tabular}
% \end{table}

\begin{table}
  \caption{Relative errors (Spearman's rank correlation) for the predicted drag coefficient $C_D$ ($\rho_D$) and lift coefficient $C_L$ ($\rho_L$). We want Spearman's correlation to be close to one. Those quantities are computed as a post-processing from the unnormalized regressed fields.}
  \label{tab:spear}
  \centering
  \begin{tabular}{ccccccc}
    \toprule
    Model & \multicolumn{2}{c}{Relative error} & \multicolumn{2}{c}{Spearman's correlation}  \\
     & $C_D$ & $C_L$ & $\rho_D$ & $\rho_L$ \\
    \midrule
    MLP & 4.289 $\pm$ 0.679 & 0.767 $\pm$ 0.108 & -0.117 $\pm$ 0.256 & 0.913 $\pm$ 0.018 \\
    GraphSAGE & \textbf{4.050 $\pm$ 0.704} & 0.517 $\pm$ 0.162 & -0.303 $\pm$ 0.124 & 0.965 $\pm$ 0.011 \\
    PointNet & 14.637 $\pm$ 3.668 & 0.742 $\pm$ 0.186 & -0.022 $\pm$ 0.097 & 0.938 $\pm$ 0.023 \\
    Graph U-Net & 10.385 $\pm$ 1.895 & \textbf{0.489 $\pm$ 0.105} & -0.138 $\pm$ 0.258 & \textbf{0.967 $\pm$ 0.019} \\
    \bottomrule
  \end{tabular}
\end{table}

\begin{table}
  \caption{Running time for one simulation on 16 CPU cores of an AMD Ryzen™ Threadripper™ 3960X compared to training and inference time of the different models on an NVIDIA GEFORCE RTX 3090. The inference time is given for one call of a model on a batch of 32000 nodes, for one simulation we need around a hundred calls to get a result on the entire mesh as the nodes are chosen randomly on the CFD mesh. The number of parameters for each model is given as additional information.}
  \label{tab:time}
  \centering
  \begin{tabular}{ccccc}
    \toprule
    Model & \multicolumn{2}{c}{Running time} & \# Parameters \\
    & Training & Inference (\si{\micro\second}) & \\
    \midrule
    MLP & $\sim$\SI{2}{\hour}\SI{20}{\minute} & 13.3 $\pm$ 0.2 & 19988 \\
    GraphSAGE & $\sim$\SI{4}{\hour}\SI{20}{\minute} & 20.9 $\pm$ 2.3 & 29204 \\
    PointNet & $\sim$\SI{2}{\hour}\SI{40}{\minute} & 33.9 $\pm$ 3.5 & 75244\\
    Graph U-Net & $\sim$\SI{6}{\hour}\SI{50}{\minute} & 357.8 $\pm$ 36.9 & 65820\\
    \midrule
    Simulation & \multicolumn{2}{c}{$\sim$\SI{25}{\minute}} &  \\
    Dataset & \multicolumn{2}{c}{$\sim$20 days} &  \\
    \bottomrule
  \end{tabular}
\end{table}

\begin{table}
  \caption{Mean squared error on the different normalized fields, relative error, and Spearman's correlation for the force coefficients on the four different tasks for the GraphSAGE model. Only the reduced pressure is given on the surface ($\Bar{p}_s$) as the other quantities are null via the boundary conditions.}
  \label{tab:score_GSAGE}
  \centering
  \begin{tabular}{ccccc}
    \toprule
    Field / Coefficient & \multicolumn{4}{c}{Task}  \\
    & Full & Scarce & Reynolds & AoA \\
    \midrule
    $\Bar{u}_x$ ($\times 10^{-2}$) & 0.832 $\pm$ 0.015 & 1.457 $\pm$ 0.125 & 7.558 $\pm$ 1.046 & 4.435 $\pm$ 0.334 \\
    $\Bar{u}_y$ ($\times 10^{-2}$) & 0.994 $\pm$ 0.052 & 1.454 $\pm$ 0.123 & 3.498 $\pm$ 0.613 & 9.400 $\pm$ 2.167 \\
    $\Bar{p}$ ($\times 10^{-2}$) & 0.661 $\pm$ 0.050 & 4.696 $\pm$ 0.804 & 3.826 $\pm$ 0.248 & 10.908 $\pm$ 2.164 \\
    $\nu_t$ ($\times 10^{-1}$) & 0.160 $\pm$ 0.021 & 0.611 $\pm$ 0.079 & 1.694 $\pm$ 0.383 & 5.178 $\pm$ 0.365 \\
    $\Bar{p}_s$ ($\times 10^{-1}$) & 0.662 $\pm$ 0.103 & 1.945 $\pm$ 0.336 & 1.797 $\pm$ 0.338 & 7.638 $\pm$ 0.945 \\
    \midrule
    $C_D$ & 4.050 $\pm$ 0.704 & 3.504 $\pm$ 0.998 & 8.971 $\pm$ 1.278 & 5.589 $\pm$ 1.090 \\
    $C_L$ & 0.517 $\pm$ 0.162 & 0.385 $\pm$ 0.097 & 0.616 $\pm$ 0.124 & 0.818 $\pm$ 0.300 \\
    $\rho_D$ & -0.303 $\pm$ 0.124 & -0.139 $\pm$ 0.175 & 0.013 $\pm$ 0.064 & 0.055 $\pm$ 0.171 \\
    $\rho_L$ & 0.965 $\pm$ 0.011 & 0.981 $\pm$ 0.006 & 0.927 $\pm$ 0.027 & 0.908 $\pm$ 0.019 \\
    \bottomrule
  \end{tabular}
\end{table}

From the plots of the different examples of boundary layers and skin friction coefficients given in Appendix~\ref{ap:results}, we conclude that the models have difficulties to predict the wall shear stresses as the velocity values at the closest nodes from the geometry are often largely overestimated. This particularly affects the accuracy of the drag coefficient as we can see with the Spearman's correlation $\rho_D$, the relative error on the drag coefficient, and the plot of the predicted drag with respect to the true drag coefficients (see Figure~\ref{fig:coefs} left). However, the wall shear stress has a small impact on the lift coefficient compared to the pressure at the surface of airfoils. Hence, as the pressure is more accurately inferred compared to the wall shear stress, as we can see by looking at the plots of the pressure coefficient at the surface of airfoils in Appendix~\ref{ap:results}, the inferred lift coefficient is also more accurately inferred and the rank is better predicted (see Figure~\ref{fig:coefs} right) leading to a Spearman's correlation close to one.

\paragraph{Difficulties of the different tasks.} In Table~\ref{tab:score_GSAGE} we give the scores of the GraphSAGE model on the four different tasks. The scores are given on the test set associated with each task. This makes difficult the direct comparison as the test set for the Reynolds and angle of attack extrapolation regimes are both different from the test set of the full and scarce regimes.

In the full and scarce regime, the MSE over the different fields shows that the GraphSAGE model is performing better in interpolation when more data is available, as expected. However, the scores on the force coefficients are slightly higher in the scarce regime which tells us that the MSE over the different fields is not necessarily a good proxy for the accuracy of the force coefficients. This can be understood as the computations of the force coefficients involve an integration over the surface and can lead to the accumulation or compensation of local errors.

In the Reynolds and angle of attack extrapolation regimes, the MSE scores are significantly higher than in the full and scarce data regimes. As expected, the extrapolation tasks are more difficult than the interpolation tasks. The Spearman's correlation for the lift coefficient is also significantly lower for both extrapolation tasks than for the interpolation ones, supporting the previous observation.

Finally, in Table~\ref{tab:time}, we confirm that even for a two-dimensional case, the training cost of models is rapidly amortized (after, in the worst case, a dozen of simulations).
% Present the different models. 1/2 page

\section{Conclusion}\label{sec:conclusion}
In this work, we presented a high-fidelity dataset of solutions of the two-dimensional RANS equations around NACA airfoils. Simulations have been done at Reynolds of the order of magnitude of what we find in subsonic flight regimes mimicking classical aerodynamics setups. We defined four ML tasks highlighting the different challenges of surrogate models, from scarce data regimes to extrapolation. We proposed different metrics focusing not only on the velocity and pressure fields but also on the force coefficients. Those metrics quantify the ability of ML models to accurately predict fields and force coefficients in addition to their ability to rank the latter, for example, for shape optimization. Different baselines have been introduced from the GDL framework, highlighting the need for models that can handle unstructured point clouds in order to be able to accurately predict force coefficients. Those baselines have shown in the proposed setting, as expected, that the prediction of the drag coefficient is more challenging than the prediction of the lift coefficient as the wall shear stress is derived from the velocity field and not directly regressed like the pressure.

\paragraph{Metrics hierarchy.} From a design-oriented perspective we may set a hierarchy for the proposed metrics. The Spearman's correlation for the force coefficients is the main metrics to maximize the recovery of the best airfoils in terms of lift-over-drag ratio as it quantifies the ability of models to preserve the force coefficient ranks. In addition to this metric, the plots of the predicted with respect to true force coefficients give qualitative information on the accuracy of the model for each simulation of the test set. Then, the relative errors for the force coefficients are important but not crucial from a design perspective and their minimization is secondary. We may say that a model is effective if it has Spearman's correlations close to one and accurate if it has low relative errors and low MSE on the fields over the volume and the airfoil. The goal here is an effective and accurate model. For relative errors, the bound of \SI{5}{\percent} is often used to state that a model is accurate enough.

\paragraph{Limitations.} Concerning the dataset itself, we restricted the design space of airfoils to NACA 4 and 5 digits for the sake of simplicity and meshing automation but we can expect that models trained on this dataset will have difficulties generalizing to more exotic shapes. Following this, we did not propose an extrapolation test on out-of-distribution airfoils. Also, even though the problem proposed is a classical aerodynamics one, it is two-dimensional and does not reflect entirely the complexity of three-dimensional natural phenomena which implies that models working on this dataset could not necessarily be extended to more generic three-dimensional cases.

In terms of baselines, we proposed four architectures of different types, an MLP, a GNN, a network acting on point clouds, and a multi-scale GNN architecture. The GNN approach suffers from its heaviness when dealing with large graphs leading to the necessity of building a downsampling strategy. In addition to that, auto differentiation with respect to positions is not possible with GNN as an entire graph is given in inputs. This leads to difficulties in the inference process and the need to rely on input CFD meshes to compute derivatives using numerical schemes. On the other hand, the MLP approach allows more flexibility and does not require a subsampling strategy or the input CFD mesh to compute derivatives. Uniform sampling is also possible in this case and several models have already shown their ability to fit complex signals \cite{siren,bacon}. The main downside of such approaches is the generalization capacity requiring conditioning or hypernetworks to work on multiple examples and generalize to unseen ones. Moreover, let us mention equivariant networks \cite{brandstetter2022geometric} as another promising direction to handle the lack of data often encountered in such tasks by leveraging symmetries of the problem. Finally, neural operators handling unstructured data such as DeepONet \cite{lu2021learning} are by definition well suited for such tasks and could be mixed with previous techniques to achieve high-performance surrogate models.

In total, we consider this work as a first step towards the generic treatment of real-world physical phenomena in ML. We hope that it will lower the potential barrier for entering the field of ML applied to physical systems and that it will encourage the construction of models that are not only good on the predicted fields but also on meaningful derived quantities.

% 3/4 page
% \paragraph{Limitation of our works.}\textcolor{red}{\bf demandé dans la section 7: paper checklist}
% \paragraph{Future works.}\begin{itemize*}
%     \item Blog (could be submitted to ICLR / Neurips blog track on published papers) 
%     \item Detailed documentation in a dedicated website for maintainability 
%     \item include AirfoilsPDE dataset in largely used ML software such as PyG (Pytorch Geometric
%     \item Extension to 3D cases, compressible RANS, unsteady processes
% \end{itemize*}\textcolor{red}{\bf Je te laisse MAJ cette partie en fonction de l'espace qui te reste. J'ai mis quelques idées. Feel free to add ideas, remove the ones l added. ...}

% \newpage
\newpage
\bibliographystyle{plainnat}
\bibliography{neurips_data_2022}

\newpage

\section{Paper Checklist}
\begin{itemize}
    \item (a) Do the main claims made in the abstract and introduction accurately reflect the paper's contributions and scope?  \textbf{Yes}
    \item (b) Have you read the ethics review guidelines and ensured that your paper conforms to them? \textbf{Yes}
    \item (c) Did you discuss any potential negative societal impacts of your work? \textbf{Military applications like weapons design. However, the license that we have chosen is open source. See Appendix \ref{sec:licence}.}
    \item (d) Did you describe the limitations of your work? Yes, in Section~\ref{sec:conclusion}
    \item If you are including theoretical results:
    \begin{itemize}
        \item (a) Did you state the full set of assumptions of all theoretical results? N/A
        \item (b) Did you include complete proofs of all theoretical results? N/A
    \end{itemize}
    \item If you ran experiments:
    \begin{itemize}
        \item (a) Did you include the code, data, and instructions needed to reproduce the main experimental results (either in the supplemental material or as a URL)? \textbf{Yes. We provide a URL to download the dataset with the main paper deadline publicly.} 
        
        \textbf{We release the code through a \href{https://github.com/Extrality/AirfRANS}{GitHub repository} to run all the experiments: including data preprocessing, reproduction of ML tasks/models, and our quantitative and qualitative results. }
        
        \textbf{We provide another \href{https://github.com/Extrality/NACA_simulation}{GitHub repository} for running new simulations over 4 and 5 digits series NACA airfoils.}
        
        \textbf{Finally, we provide the \href{https://airfrans.readthedocs.io/en/latest/index.html}{AirfRANS Python library} to easily manipulate simulations from the dataset and its associated \href{https://github.com/Extrality/airfrans_lib}{GitHub repository}.}
        \item (b) Did you specify all the training details (e.g., data splits, hyperparameters, how they were chosen)? \textbf{Yes. In section \ref{sec:results} of the main paper and in Appendix~\ref{ap:results}}
        \item (d) Did you include the amount of compute and the type of resources used (e.g., type of GPUs, internal cluster, or cloud provider)? Yes, in Section~\ref{sec:dataset} and Section~\ref{sec:setup}.
    \end{itemize}
    \item If you are using existing assets (e.g., code, data, models) or curating/releasing new assets:
    \begin{itemize}
        \item (a) If your work uses existing assets, did you cite the creators? \textbf{Yes. We cite all the works in the appropriate sections, either in the main paper or the supplementary part. The models we took from the existing works are adapted before training them with our dataset.}
        \item (b) Did you mention the license of the assets? \textbf{N/A}
        \item (c) Did you include any new assets either in the supplemental material or as a URL? \textbf{N/A}
        \item Did you discuss whether and how consent was obtained from people whose data you're using/curating?  \textbf{N/A}
        \item (e) Did you discuss whether the data you are using/curating contains personally identifiable information or offensive content? \textbf{N/A}

    \end{itemize}
        \item If you used crowdsourcing or conducted research with human subjects: \textbf{N/A}

\end{itemize}

\newpage
\appendix
\tableofcontents
\newpage

\section{Resources Availability and Licensing}\label{sec:licence}

This work is open-source under \href{https://choosealicense.com/licenses/odbl-1.0/}{Open Data Commons Open Database License v1.0}. For both dataset generation (mesh generation, OpenFoam simulations) and ML experiments/baselines. The Open Database License (ODbL) is a license agreement that allows users to freely share, modify, and use a database while maintaining this same freedom for others. The \textbf{Permissions} include:
\begin{itemize*}
 \item Commercial use
\item  Distribution
 \item Modification
 \item Private use
\end{itemize*}
. The \textbf{limitations} are:
\begin{itemize*}
     \item Liability
 \item Patent use
 \item Trademark use
 \item Warranty
\end{itemize*}
 and finally the \textbf{conditions} are:
\begin{itemize*}
   \item Disclose sourceƒ
\item License and copyright notice
 \item Same license
\end{itemize*}. This license is the same than the one used in our first version of this work \cite{bonnet2022an}.

\section{Broader impact}
This work could be used to:
\begin{enumerate}
    \item experiment new Geometric Deep Learning (GDL) models in this area,
    \item study the capabilities of Deep Learning (DL) to capture physical phenomena relying on our physics-based evaluation protocol,
    \item give insights to establish new research directions for combining numerical simulation and Machine Learning (ML) following the behaviors that will be observed in our quantitative and qualitative results w.r.t the targeted physical metrics that we developed in purpose for this work,
    \item extend graph-mesh, and point clouds benchmarking datasets and applications of GNNs to real-world-like physics problems,
    \item build surrogate solvers to help Computational Fluid Dynamics (CFD) engineers optimize design cycles and iterate as efficiently as needed on their designs,
    \item  reduce the cost (time and materials) of prototyping new designs of planes while avoiding dangerous studies in the real world, as well as enabling the test of several configurations,
    \item study the generalization/extrapolation abilities of DL to large/unseen domains as the industrial world demands, including wide ranges of initial and boundary conditions.
%    \item extend this dataset to include exotic meshes\footnote{thanks to the simple mesh process generation that we make open-source} to study extrapolation test on out-of-distribution/non-uniform airfoil shapes

\end{enumerate}
\section{Reproducibility statement}

We provide a  \href{https://github.com/Extrality/AirfRANS}{GitHub repository} to reproduce all the experiments and a \href{https://data.isir.upmc.fr/extrality/NeurIPS_2022/Dataset.zip}{link} to download the preprocessed dataset and \href{https://data.isir.upmc.fr/extrality/NeurIPS_2022/OF_dataset.zip}{another one} for the raw OpenFOAM data. The experiments have been conducted with a single NVIDIA RTX 3090 24Go. The repository the preprocessed/raw datasets include:
\begin{itemize*}
    \item code to reproduce the ML experiments
    \item code to generate the figures.
\end{itemize*}

We also provide a \href{https://github.com/Extrality/NACA_simulation}{GitHub repository} to run new simulations and to be able to reproduce the generation settings of the dataset. The simulations have been done with 16 CPU cores of an AMD Ryzen™ Threadripper™ 3960X. The codes in the repository include:
\begin{itemize*}
    \item (extensible) code to generate the meshes
    \item code to run new simulations and/or build the dataset.
\end{itemize*}

Finally, we provide the \href{https://airfrans.readthedocs.io/en/latest/index.html}{AirfRANS Python library} with its associated \href{https://github.com/Extrality/airfrans_lib}{GitHub repository} to easily manipulate simulations from the dataset.

\section{Description Of Software}
In this section, we describe the tools that we have used in this work to build the dataset\footnote{similar to our first version of our dataset \cite{bonnet2022an}}, make the visualizations, and train the models. This work makes use of computational fluid dynamics (CFD) and ML tools.

\textbf{OpenFOAM} \cite{OpenFOAM} stands for \emph{Open-source Field Operation And Manipulation}, a C++ software for developing custom numerical solvers to study continuum mechanics and CFD problems. In this work, we have used version v2112 of OpenFOAM to make our simulations. OpenFOAM is released as free and open-source software under the \emph{GNU General Public Licence}.

\textbf{ParaView} \cite{paraview} is an open-source visualization tool designed to explore and visualize efficiently large data using quantitative and qualitative metrics. ParaView runs on distributed and shared memory parallel and single processor systems. In this work, we have used it to visualize the following: point clouds, meshes, the predicted (as well as the ground truth) physical fields. We have used version 5.10.0 of ParaView in this work. ParaView  is released  as free and open-source software under the \emph{Berkeley Software Distribution License}.

\textbf{PyVista} \cite{pyvista} is an open-source tool based on a handy interface for the Visualization ToolKit (VTK). It is simple to use in interaction with NumPy \cite{numpy} and other Python libraries. It is mainly used for mesh analysis. In this work, we use PyVista to build the inputs of our DL models. We have used version 0.36.1 of PyVista in this work. PyVista is released as free and open-source software under the \emph{MIT License}.

\textbf{PyTorch} \cite{NEURIPS2019_9015} is an open-source library for DL using GPUs and CPUs. In this work, we use PyTorch to build our training protocol. In this work, we have used version 1.11.0 of Pytorch along with CUDA 11.3 and Python 3.9.12. PyTorch is released as free and open-source \emph{Berkeley Software Distribution License}.

\textbf{PyTorch Geometric} (PyG) \cite{fey2019graph} is an open-source library for GDL built upon PyTorch which targets the training  of geometric neural networks, including point clouds, graphs and meshes. We use PyG to design our message passing schemes. In this work, we have used version 2.0.4 of PyG along with CUDA 11.3. PyG is released as free and open-source software under the \emph{MIT License}.

% In Figure \ref{fig:working_pipline}, we illustrate the whole pipeline to make the experiments. Starting from mesh generation to model training and output visualizations. We show the connection between all the aforementioned tools to perform our task. 

% \begin{figure}
%     \centering
%     \includegraphics[width = \linewidth]{Dataset/pipline.pdf}
%     \caption{The pipeline to make the experiments and the connection between the different tools. CFD, VTK, GNNs stand respectively for Computational Fluid Dynamic, Visualization Toolkit, Graph Neural Networks.}
%     \label{fig:working_pipline}
% \end{figure}

\section{Constant and Dimensionless Quantities}\label{ap:dimensionless}
The fluid used in this study is the air at \SI{298.15}{\kelvin} (\SI{25}{\celsius}) and at sea level on earth. In Table~\ref{tab:air_properties} we give the different values of the constant associated with this fluid.

\begin{table}
  \centering
  \begin{threeparttable}
      \caption{Properties of air at \SI{298.15}{\kelvin} (\SI{25}{\celsius}) and at sea level on earth.}
      \label{tab:air_properties}
      \begin{tabular}{ccc}
        \toprule
        Name & Symbol & Value                  \\
        \midrule
        Kinematic viscosity & $\nu$ & \SI{1.56e-5}{\square\meter\per\second} \\
        Specific mass\tnote{$\star$} & $\rho$ & \SI{1.184}{\kilogram\per\cubic\meter} \\
        Thermal diffusivity\tnote{$\star$} & $\alpha$ & \SI{2.25e-5}{\meter\per\second} \\
        Specific heat\tnote{$\star$} & $c_p$ & \SI{1005}{\joule\per\kilogram\per\kelvin} \\
        Atmospheric pressure\tnote{$\star$} & $p_0$ & \SI{1.013e5}{\newton\per\square\meter} \\
        Atmospheric temperature\tnote{$\star$} & $T_0$ & \SI{298.15}{\kelvin} \\
        Speed of sound in the fluid\tnote{$\star$} & $c$ & \SI{346.1}{\meter\per\second} \\
        \bottomrule
      \end{tabular} \hspace{1cm}
      \begin{tablenotes}
        \item [$\star$] Those values are important only in the compressible case, they are given for comparison with compressible simulations. Especially, the absolute pressure is set to \SI{0}{\newton\per\square\meter} for the incompressible case as it only depends on the differential pressure.
      \end{tablenotes}
  \end{threeparttable}
\end{table}

The only dimensionless quantity for the incompressible case is the Reynolds number $Re$. We add the Mach number $Ma$ and the Prandtl number $Pr$ in the compressible case. Those quantities are defined by:
\begin{align}
    Re = \frac{UL}{\nu}, \quad Ma = \frac{U}{c}, \quad Pr = \frac{\nu}{\alpha}
\end{align}
where $U$ is the characteristic velocity of the problem, $L$ its characteristic length, $c$ the speed of sound in the fluid, $\nu$ its kinematic viscosity and $\alpha$ its thermal diffusivity. The Reynolds number compares the order of magnitude of the convective term with respect to the diffusive term in the Navier–Stokes equations, the Mach number quantifies flow compressibility and the Prandtl number compares the order of magnitude of the variation of energy via momentum with respect to the variation of energy via heat transfer in the compressible form of Navier–Stokes equations.

\section{Reynolds-Averaged Navier–Stokes Equations}\label{ap:RANS}
Under certain assumptions, the dynamics of a fluid is governed by the Navier–Stokes equations. Those equations are composed of a continuity equation, three momentum equations and an energy equation (see §49 of \cite{landau} and §5.3 of \cite{wilcox}):
\begin{align} \label{eq:NS}
    \partial_t\rho + \partial_i(\rho u_i) &= 0 \\
    \partial_t(\rho u_i) + \partial_j (\rho u_ju_i) &= -\partial_i p + \partial_j\sigma_{ji}, \quad i\in\{1, 2, 3\} \\
    \partial_t\left(\rho \left(\epsilon + \frac{1}{2}u^2\right)\right) + \partial_j \left(\rho u_j \left(h + \frac{1}{2}u^2\right)\right) &= \partial_j(u_i\sigma_{ij}) - \partial_j q_j \label{eq:energy_eq}
\end{align}
where $\partial_t$ denotes a partial derivative with respect to time, $\partial_i$ a partial derivative with respect to the $i^{th}$ space coordinate, $\rho$ the fluid specific mass, $u$ the fluid velocity, $p$ the fluid pressure, $\sigma$ the viscous stress tensor, $\epsilon$ the fluid specific energy, $h$ the fluid specific enthalpy ($h := \epsilon + p/\rho$) and $q$ the heat flux density due to thermal conduction. Moreover, we used the Einstein summation convention over repeated indices. Finally, fluid dynamics theory, thermodynamic relations, Fourier law and the perfect gas law give us:
\begin{align}
    \sigma_{ij} &= \mu \left(\partial_iu_j + \partial_ju_i - \frac{2}{3}\partial_ku_k\delta_{ij}\right), \quad i,\,j\in\{1,2,3\} \\
    \epsilon &= c_v T \\
    h &= c_p T \\
    q_i &= -\kappa\partial_iT, \quad i\in\{1,2,3\} \\
    p &= \rho RT \label{eq:state}
\end{align}
where $\delta_{ij}$ is the kronecker tensor, $T$ the fluid temperature, $\mu$ the fluid dynamic viscosity (function of $T$), $\kappa$ the fluid thermal conductivity (function of $T$), $R$ the fluid specific constant, $c_v$ and $c_p$ the fluid specific heat coefficient for constant volume and pressure respectively (taken constant here). This leads to a close set of partial differential equations with 6 unknowns ($\rho$, $p$, $u$, $T$) and 6 equations:
\begin{align}
    \partial_t\rho &+ \partial_i(\rho u_i) = 0 \\
    \partial_t(\rho u_i) &+ \partial_j (\rho u_ju_i) = -\partial_i p + \partial_j\left(\mu\left(\partial_iu_j + \partial_ju_i - \frac{2}{3}\mu\partial_ku_k\delta_{ij}\right)\right) , \quad i\in\{1, 2, 3\} \\
    \partial_t\left(\rho \left(c_vT + \frac{1}{2}u^2\right)\right) &+ \partial_j \left(\rho u_j \left(c_pT + \frac{1}{2}u^2\right)\right) = \partial_j\left(\mu u_i\left(\partial_iu_j + \partial_ju_i - \frac{2}{3}\partial_ku_k\delta_{ij}\right)\right) + \partial_j (\kappa\partial_jT)
\end{align}
together with the state equation \ref{eq:state}. We chose the perfect gas law for the state equation as we are going to treat the case of air but this equation can be replaced by any state equation better suited for the problem.

In certain cases, we can decently assume that the fluid is incompressible with constant density $\rho$. We then need only 4 equations to close the problem. We get rid of the energy and state equations and find the incompressible Navier–Stokes equations:
\begin{align}
    \partial_i u_i &= 0 \\
    \partial_tu_i + \partial_j(u_iu_j) &= -\partial_i\left(\frac{p}{\rho}\right) + \nu\partial^2_{jj}u_i, \quad i\in\{1,2,3\}
\end{align}
where $\nu := \mu/\rho$ is the fluid kinematic viscosity, taken constant in this case. In order to explicitly write an important dimensionless quantity in fluid mechanics, we can rewrite last equations with dimensionless variables. Let us define, $T$, $U$, $L$ and $P$ characteristic time, velocity, length and pressure of the problem, respectively. We write:
\begin{align}
    t = T\hat{t}, \quad u = U\hat{u}, \quad x = L\hat{x}, \quad p = P\hat{p}
\end{align}
where x is the cartesian position, $\hat{t}$, $\hat{u}$, $\hat{x}$ and $\hat{p}$ are dimensionless quantities. If we write $P = \rho U^2$ and $T = L/U$, we find for the incompressible case:
\begin{align}
    \partial_{\hat{i}} \hat{u}_i &= 0 \\
    \partial_{\hat{t}}\hat{u}_i + \partial_{\hat{j}}(\hat{u}_i\hat{u}_j) &= -\partial_{\hat{i}}\hat{p} + \frac{1}{Re}\partial^2_{\hat{j}\hat{j}}\hat{u}_i, \quad i\in\{1,2,3\}
\end{align}
where $Re := UL/\nu$ is the Reynolds number. This dimensionless number quantifies the importance of the convective term with respect to the diffusive term (in order of magnitude):
\begin{align}
    \frac{\|\partial_j (u_iu_j)\|}{\nu\|\partial^2_{jj}u_i\|} \approx \frac{UL}{\nu} = Re
\end{align}
When the Reynolds number tends to $0$, diffusion term are dominant, we call it a Stokes flow. When the Reynolds number tends to $\infty$, the equations get closer to the Euler equations for inviscid fluid. At high Reynolds, new chaotic patterns can emerge close to walls and the different fields get untidy. This transition is the transition between what we call laminar (tidy) and turbulent (untidy) flows. Turbulence is a process that emerges at high Reynolds number and allows more dissipation than expected with laminar flows via the emergence of eddies of different length scales (see §33 of \cite{landau}). Theoretical resolution of such dynamics is an open problem and direct numerical simulations (DNS) are highly challenging because of their huge computational costs. Hence, different technologies have been developed in order to reduce the computational complexity of the task, for example, large eddy simulations (LES) try to filter in space the pressure and velocity fields and model the smallest eddies by adding dissipation. Another one, that we will use here, try to model all the scales of eddies by doing an ensemble average on the velocity and pressure fields. An ensemble average is a theoretical average over multiple equivalent experiments, this can also be equivalently replaced by a time averaging on a time scale big compared to turbulent fluctuations rate and small compared to the macroscopic evolution rate of the fluid. We write:
\begin{align}
    u = \Bar{u} + u', \quad p = \Bar{p} + p'
\end{align}
where $\Bar{\cdot}$ denotes an ensemble averaged quantity and $\cdot '$ its fluctuations. If we set those expressions into the incompressible Navier–Stokes equations and take the ensemble averaging of the equations, we get:
\begin{align}
    \partial_i \Bar{u}_i &= 0 \\
    \partial_t\Bar{u}_i + \partial_j(\Bar{u}_i\Bar{u}_j) &= -\partial_i\left(\frac{\Bar{p}}{\rho}\right) + \nu\partial^2_{jj}\Bar{u}_i - \frac{1}{\rho}\partial_j (\sigma_t)_{ij}, \quad i\in\{1,2,3\}
\end{align}
where $(\sigma_t)_{ij} := -\rho\overline{u'_i u'_j}$ is called the Reynolds stress tensor. We now have a new unknown in our equations and the problem is not close anymore. A common assumption known as the Boussinesq hypothesis is to write the Reynolds stress tensor in the same way as the viscous stress tensor:
\begin{align}
    (\sigma_t)_{ij} &= \rho\nu_t\left(\partial_i\Bar{u}_j + \partial_j\Bar{u}_i\right) - \frac{2}{3}\rho k \\
    k &= \frac{1}{2}\overline{u'_i u'_i}
\end{align}
where $\nu_t$ is called the turbulent kinematic viscosity and $k$ the specific kinematic energy of turbulence. The term $-2\rho k/3$ is set to ensure the null value of the trace of $\sigma_t$. By defining an effective pressure, abusively denoted by the same symbol, $\Bar{p}$, $\Bar{p} := \Bar{p} + 2\rho k$, we find:
\begin{align}
    \partial_i \Bar{u}_i &= 0 \\
    \partial_t\Bar{u}_i + \partial_j(\Bar{u}_i\Bar{u}_j) &= -\partial_i\left(\frac{\Bar{p}}{\rho}\right) + \partial_{j}\left[(\nu + \nu_t)\partial_{j}\Bar{u}_i\right], \quad i\in\{1,2,3\}
\end{align}
This set of equations is known as the Reynolds-Averaged Navier–Stokes (RANS) equations. In order to close our set of equations, we need a last equation for $\nu_t$. Such equation is called a turbulence model and plenty of them have been developed in the last decades to recover experimental results in certain environments. We very briefly present two turbulence models that we are going to use in our experiments and let the details of those model in the references given. The Spalart-Allmaras model \cite{spalart} is a one-equation model designed for aerodynamics problems, it involves a modified viscosity called $\Tilde{\nu}$. The $k-\omega$ SST model \cite{SST} is the blending of two two-equations turbulence model, namely the $k-\varepsilon$ and the $k-\omega$ models \cite{kepsilon, komega}, and it extends the domain of application of both by switching models where it is more relevant to use one instead of another. It involves two quantities, the specific kinematic turbulent energy $k$ and the specific turbulence dissipation rate $\omega$.

In the compressible case, a mass-average is applied to the Navier–Stokes equations, for example in the case of the velocity, we use:
\begin{align}
    \Tilde{u} = \frac{1}{\Bar{\rho}}\overline{\rho u}
\end{align}
and we can decompose the velocity in a mass-averaged term and a fluctuation term:
\begin{align}
    u = \Tilde{u} + u''
\end{align}
By doing this for different quantities such as specific energy, specific enthalpy, temperature etc... we can write a new set of equations in a similar form as \ref{eq:NS} - \ref{eq:energy_eq}. This is called the Favre-Averaged Navier–Stokes equations, details can be found in chapter 5 of \cite{wilcox}.

Finally, the RANS equations are the equations solved by the \emph{simpleFoam} solver and the Favre-Averaged Navier–Stokes equations the ones solved by the \emph{rhoSimpleFoam} solver in the OpenFOAM suite. We compare our results with compressible simulations and the results given in the TMR \cite{TMR} in Appendix~\ref{ap:validation}.

\section{Force Coefficients}

The stress force $df$ acting on a face of area $dS$ and normal $n$ is:
\begin{align}
    df = -pn + 2\mu S\cdot n
\end{align}
We can conclude that for a geometry of surface $\mathcal{S}$, the stress force $F$ acting on it can be computed via:
\begin{align}
    F &= \oint_\mathcal{S} \sigma\cdot n dS \\
    &= -\oint_\mathcal{S} pndS + \oint_\mathcal{S} 2\mu S\cdot n dS
\end{align}
We call the term $P := -pn$ the wall pressure and the term $\tau := 2\mu S\cdot n$ the wall shear stress. Ultimately, we call drag $D$ and lift $L$ the component of $F$ that are respectively parallel and orthogonal to the main direction of the flow. If $u_{\parallel}$ is the unit direction of the velocity vector and $u_{\perp}$ its orthogonal unit direction, we have:
\begin{align}
    D &= \left(\oint_\mathcal{S} p dS + \oint_\mathcal{S} \tau dS\right)\cdot u_{\parallel} \\
    L &= \left(\oint_\mathcal{S} p dS + \oint_\mathcal{S} \tau dS\right)\cdot u_{\perp}
\end{align}

In the case of RANS equations, we add terms that take in account the effect of turbulence over the geometry. The pressure $p$ is replaced by an effective mean-field pressure $\Bar{p}$ and the wall shear stress is given by $\tau = 2(\mu + \mu_t)S\cdot n$ where $\mu_t$ is the dynamic turbulent viscosity. However, as the turbulent viscosity is null over the airfoil, we recover $\tau = 2\mu S\cdot n$.

For incompressible fluids we also often divide those quantities by $\rho$ the density of the fluid and solvers often express the results in terms of reduced pressure $\Bar{p} \rightarrow \Bar{p}/\rho$ and kinematic (turbulent) viscosity $\nu := \mu/\rho$ ($\nu_t := \mu_t/\rho$). We use this convention in this work.

\section{Airfoil Generation and Statistics}\label{ap:airfoil}
In this section, we review the construction of the NACA 4 and 5 digits \cite{naca}. Both of them are built in the same manner and rely only on 3 or 4 parameters for the 4 or 5 digits respectively. Each airfoil is defined via a camber lined and an envelope, the only difference between the 4 and 5 digits is the definition of the camber line.

\paragraph{NACA 4 digits.} Those profiles are defined by the name NACA followed by four digits MPXX where the first two digits M and P defined the camber line and the last two digits XX defined the maximum thickness of the profile in percentage of the chord (the total length of the airfoil). More precisely, M defines the maximum ordinate of the camber line in hundredth of the chord and P the position of this maximum from the leading edge in tenth of the chord. If we denote the chord $c$, the camber line of the NACA 4312 profile will have a maximum ordinate of camber of $y = 0.04c$, at $x = 0.3c$ and the profile will have a maximum thickness of $0.12c$. Also, the leading edge and the trailing edge of each airfoil are always taken at the points $(0, 0)$ and $(c, 0)$ in the $x-y$ plane respectively. From this point, all the abscissas and ordinates will be given in length per chord.

For the NACA 00XX, a symmetrical profile, the camber line is a straight line from $x = 0$ to $x = 1$ and the upper surface is defined by the graph of the function:
\begin{align}
    y_t(x) = \frac{t}{0.2}\left(0.2969\sqrt{x} - 0.126x - 0.3516x^2 + 0.2843x^3 - 0.1015x^4\right)
\end{align}
where $t := XX/100$ is the thickness defined with the two last digits. This definition involve a trailing edge with a thickness of $0.002c$. If, for example for numerical propose, we want to have a thickness of 0 at the trailing edge, we can change the coefficient of the fourth order term from $-0.1015$ to $-0.1036$. The lower surface is defined as $-y_t$.

For a generic NACA MPXX, The camber line is defined by the first two digits and follow the graph of the function:
\begin{align}
    y_c(x) =
    \begin{cases}
    m\frac{x}{p^2}(2p - x), \quad 0 \leqslant x\leqslant p \\
    m\frac{1 - x}{(1-p)^2}(1 + x - 2p), \quad p < x \leqslant 1
    \end{cases}
\end{align}
where $m := 0.01$M and $p:= 0.1$P.

Finally, the upper surface of a generic NACA MPXX is given by the set of coordinates $(x_u, y_u)$ defined as:
\begin{align}
    \begin{cases}
    x_u = x - y_t(x)\sin\theta(x) \\
    y_u = y_c(x) + y_t\cos\theta(x) \qquad \textrm{for } x\in [0,1]\\
    \theta(x) = \arctan y_c'(x)
    \end{cases}
\end{align}
where $y_c'$ is the derivative of $y_c$. The lower surface is given by a similar set of coordinates $(x_l, y_l)$ defined as:
\begin{align}
    \begin{cases}
    x_l = x + y_t(x)\sin\theta(x) \\
    y_u = y_c(x) - y_t\cos\theta(x) \qquad \textrm{for } x\in [0,1]\\
    \theta(x) = \arctan y_c'(x)
    \end{cases}
\end{align}

\paragraph{NACA 5 digits.} As we already stated earlier, the only difference between the NACA 4 and 5 digits is the definition of the camber line. In the case of the 5-digits LPQXX, the 3 first parameters defining the camber line are less explicit than for the 4 digits case but allow more complex shapes. The last two are the same as in the 4-digits case (\emph{i.e.} maximum thickness in hundredth of chord). The first digit L controls the camber implicitly via an optimal lift coefficient $C_L$, \emph{i.e.} it will give you the camber of the airfoil such that $C_L = 0.15$L. The second digit P is almost the same as in the 4-digits case, \emph{i.e.} it defines the position of the maximum of camber of the camber line in twentieth of chord. Lastly, the third digit Q is either 0 or 1 and represent a standard camber (similar to the 4-digits case) for 0 and a reflex camber for 1. The reflex camber is a double cambered line that makes the profile more stable (by setting the pitch coefficient to 0).

For a generic NACA LPQXX, the standard camber line is defined via the graph of the function:
\begin{align}
    y_c(x) = 
    \begin{cases}
    K_1\left(m^2(3-m)x - 3mx^2 + x^3\right), \quad 0\leqslant m \\
    K_1m^3(1-x), \quad m< x \leqslant 1
    \end{cases}
\end{align}
where $m$ is not the position of the maximum camber $p := 0.05$P but is related to it via the equation:
\begin{align}\label{eq:max_camb}
    p = m\left(1-\sqrt{\frac{m}{3}}\right)
\end{align}
and $K_1$ is related to $C_L$ via:
\begin{align}
    \begin{cases}
    K_1 = \frac{C_L}{Q} \\
    Q = \frac{3m - 7m^2 + 8m^3 - 4m^4}{\sqrt{m(1-m)}} - \frac{3}{2}(1-2m)\left(\frac{\pi}{2} - \arcsin(1-2m)\right)
    \end{cases}
\end{align}

The reflex camber line is defined via the graph of the function:
\begin{align}
    y_c(x) = 
    \begin{cases}
    K_1\left(m^3(1-x) - K_2(1-m)^3x + (x-m)^3\right), \quad 0\leqslant m \\
    K_1\left(m^3(1-x) - K_2(1-m)^3x + K_2(x-m)^3\right), \quad m< x \leqslant 1
    \end{cases}
\end{align}
where $m$ and $K_1$ are defined as in the standard case and $K_2$ is defined via:
\begin{align}
    K_2 = \frac{3(m - p)^2 - m^3}{(1-m)^3}
\end{align}
We use a standard Newton's method to numerically solve equation \ref{eq:max_camb} in $m$.

\begin{figure}
  \centering
  \begin{subfigure}{0.32\textwidth}
      \centering
      \includegraphics[width = \linewidth]{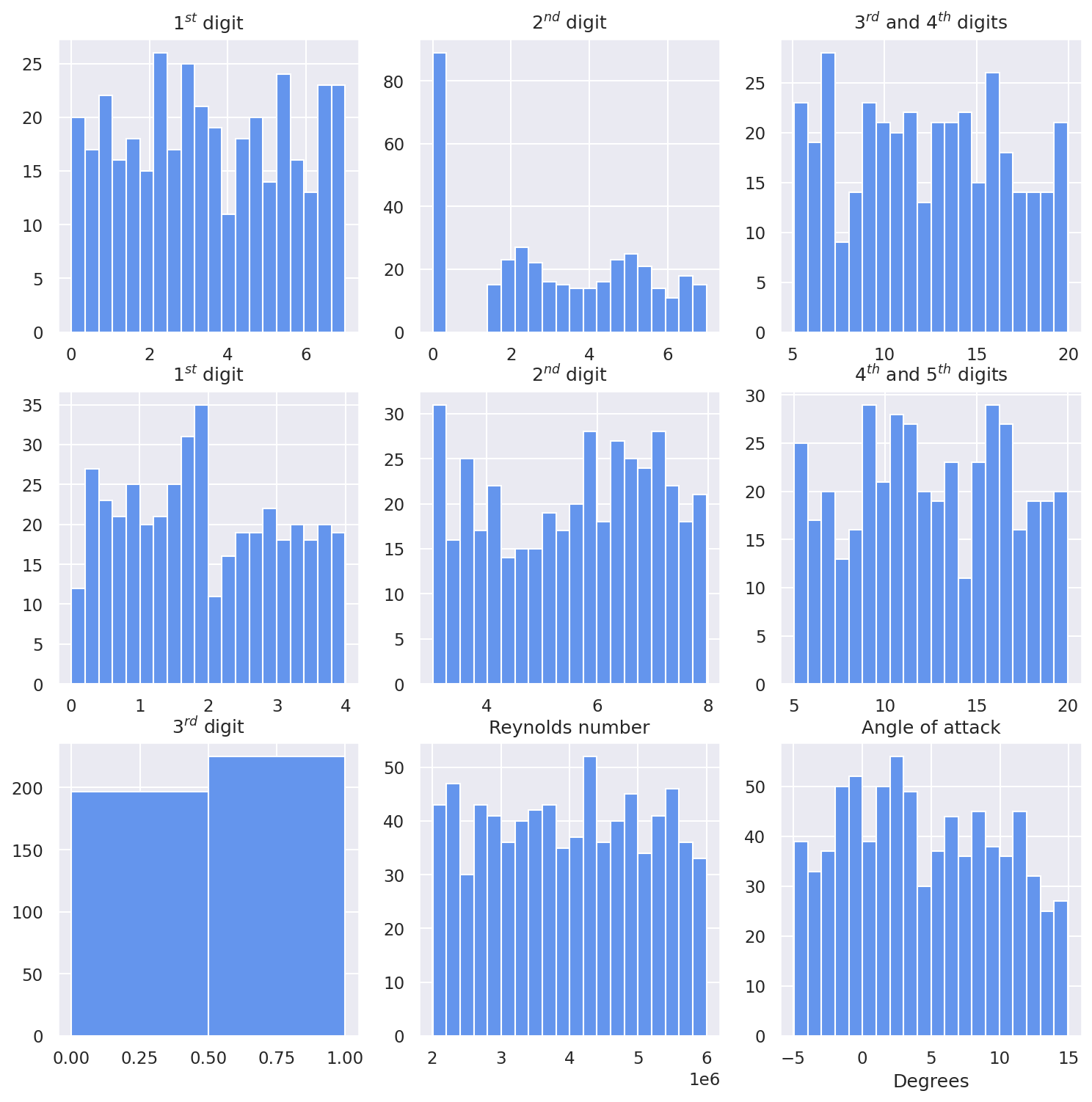}
      \caption{}
   \end{subfigure}
   \begin{subfigure}{0.32\textwidth}
      \centering
      \includegraphics[width = \linewidth]{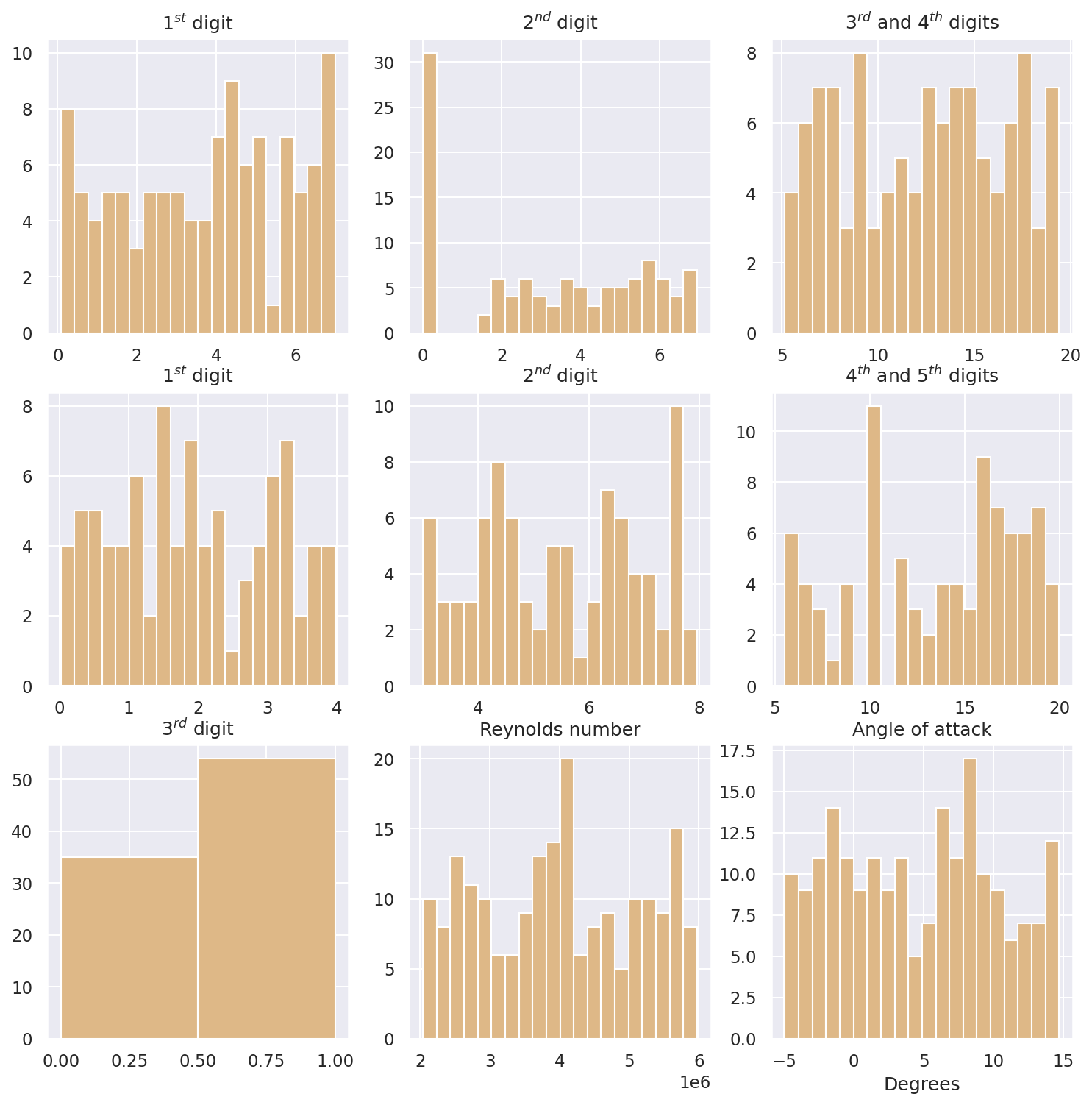}
      \caption{}
  \end{subfigure}
  \begin{subfigure}{0.32\textwidth}
      \centering
      \includegraphics[width = \linewidth]{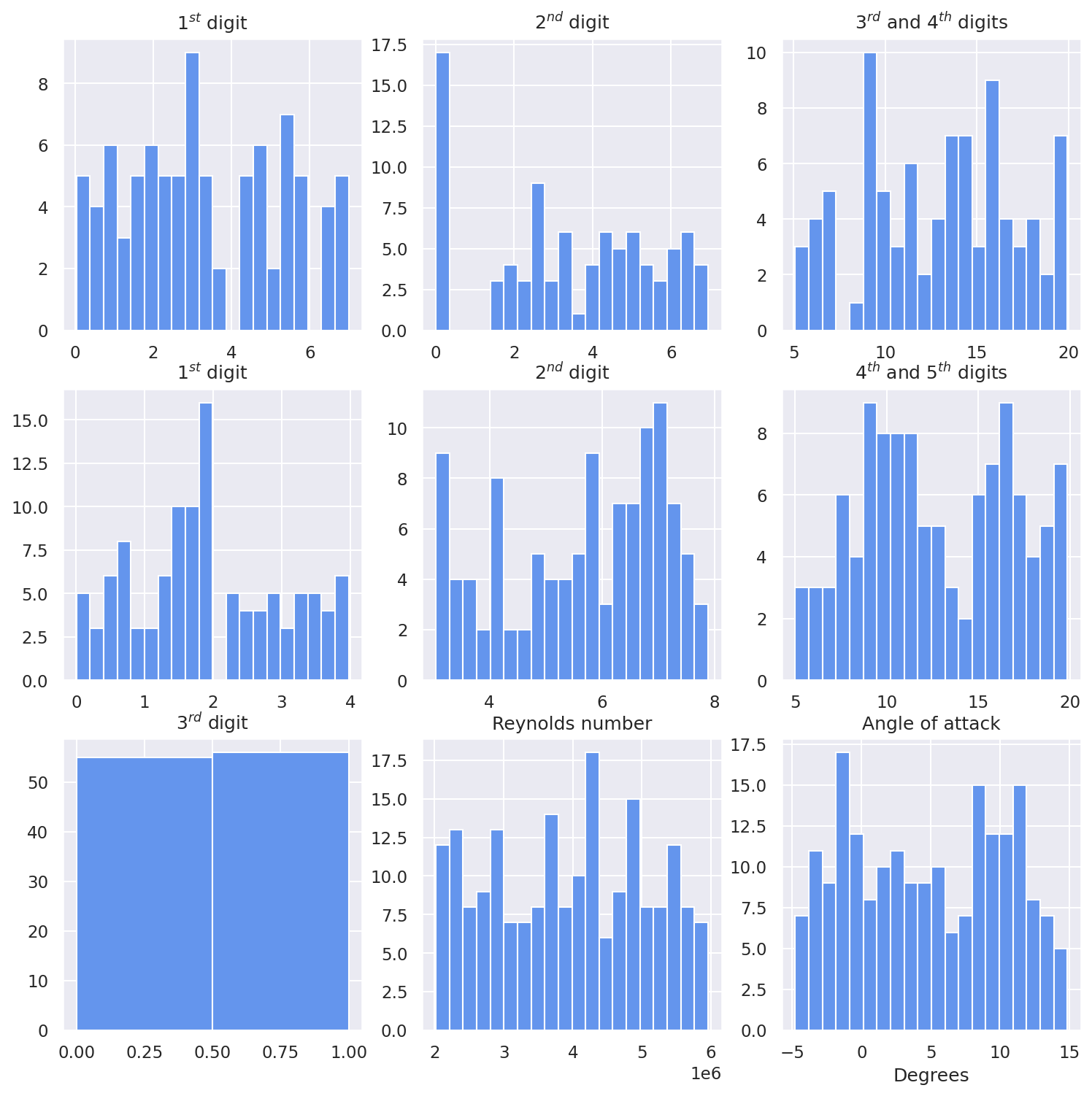}
      \caption{}
   \end{subfigure}
   
   \begin{subfigure}{0.32\textwidth}
      \centering
      \includegraphics[width = \linewidth]{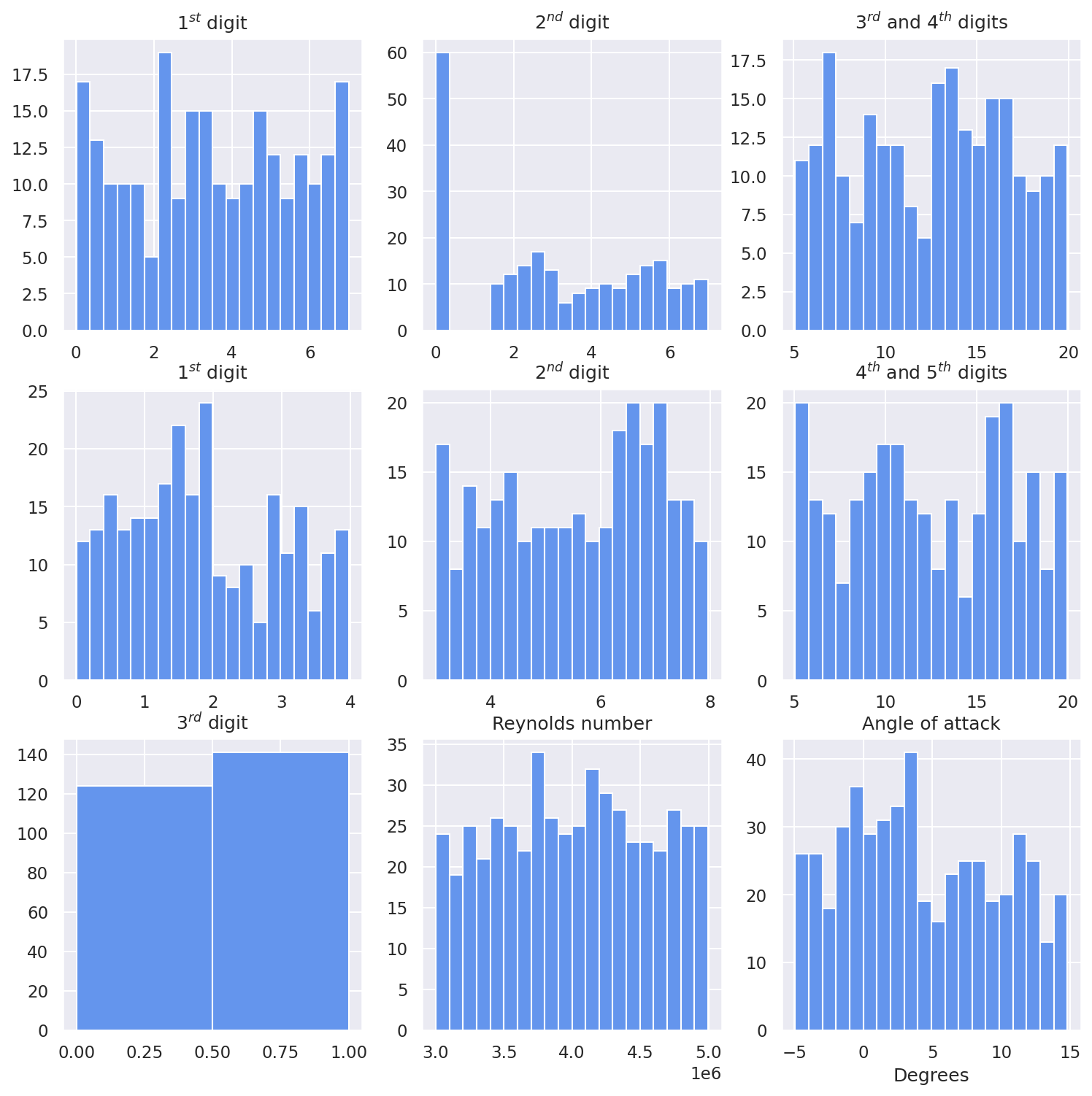}
      \caption{}
   \end{subfigure}
   \begin{subfigure}{0.32\textwidth}
      \centering
      \includegraphics[width = \linewidth]{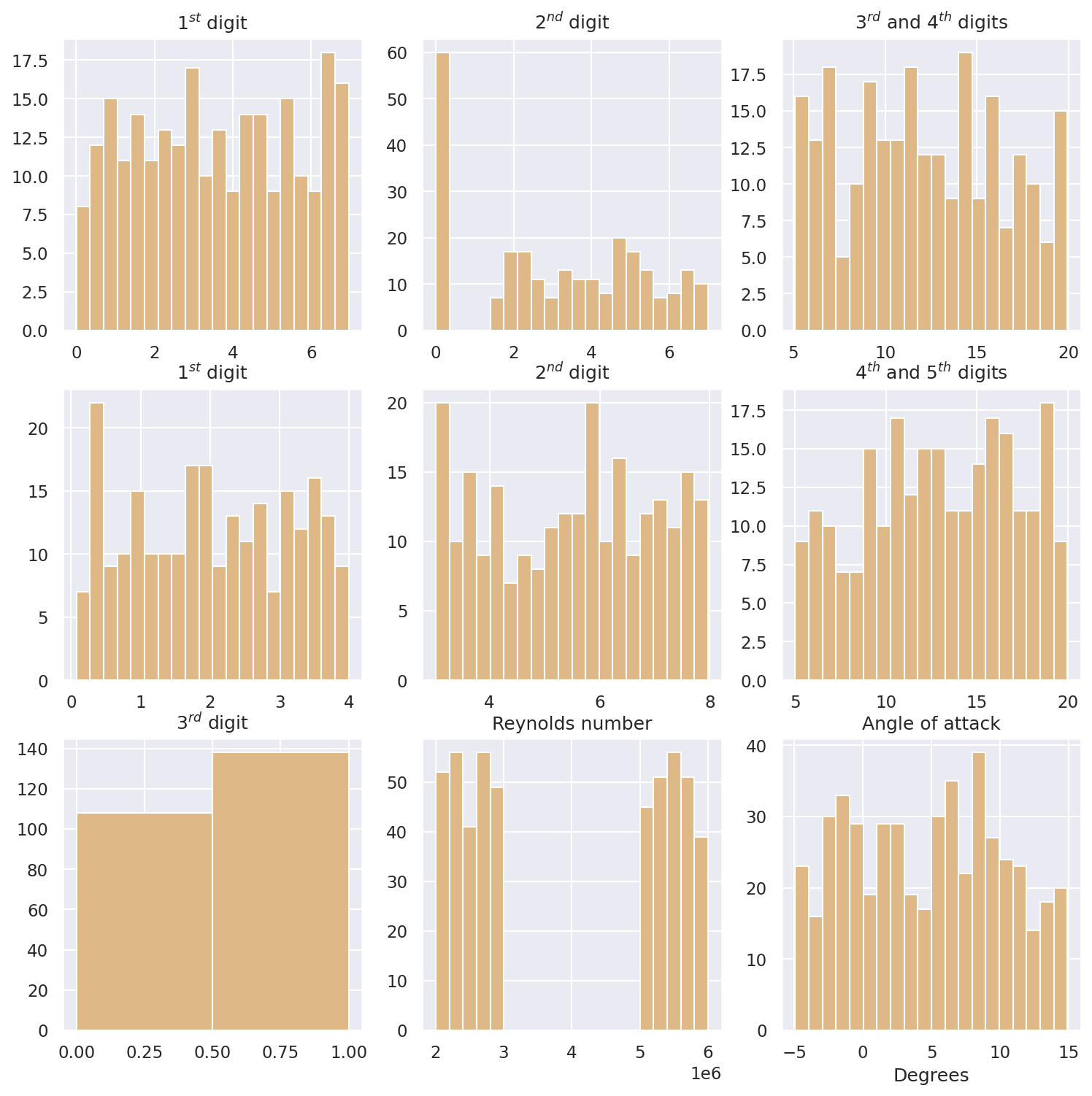}
      \caption{}
   \end{subfigure}
   
   \begin{subfigure}{0.32\textwidth}
      \centering
      \includegraphics[width = \linewidth]{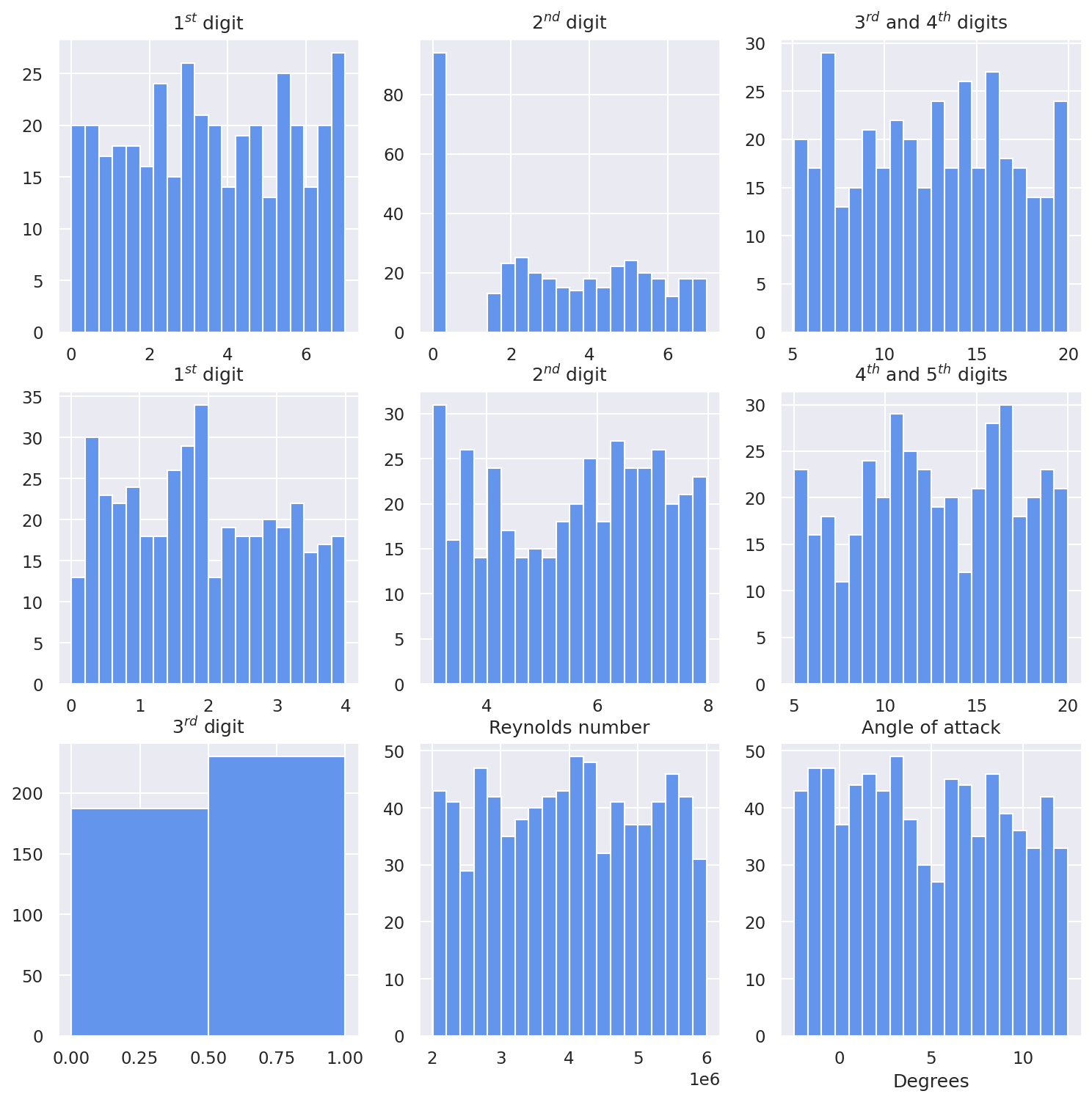}
      \caption{}
   \end{subfigure}
   \begin{subfigure}{0.32\textwidth}
      \centering
      \includegraphics[width = \linewidth]{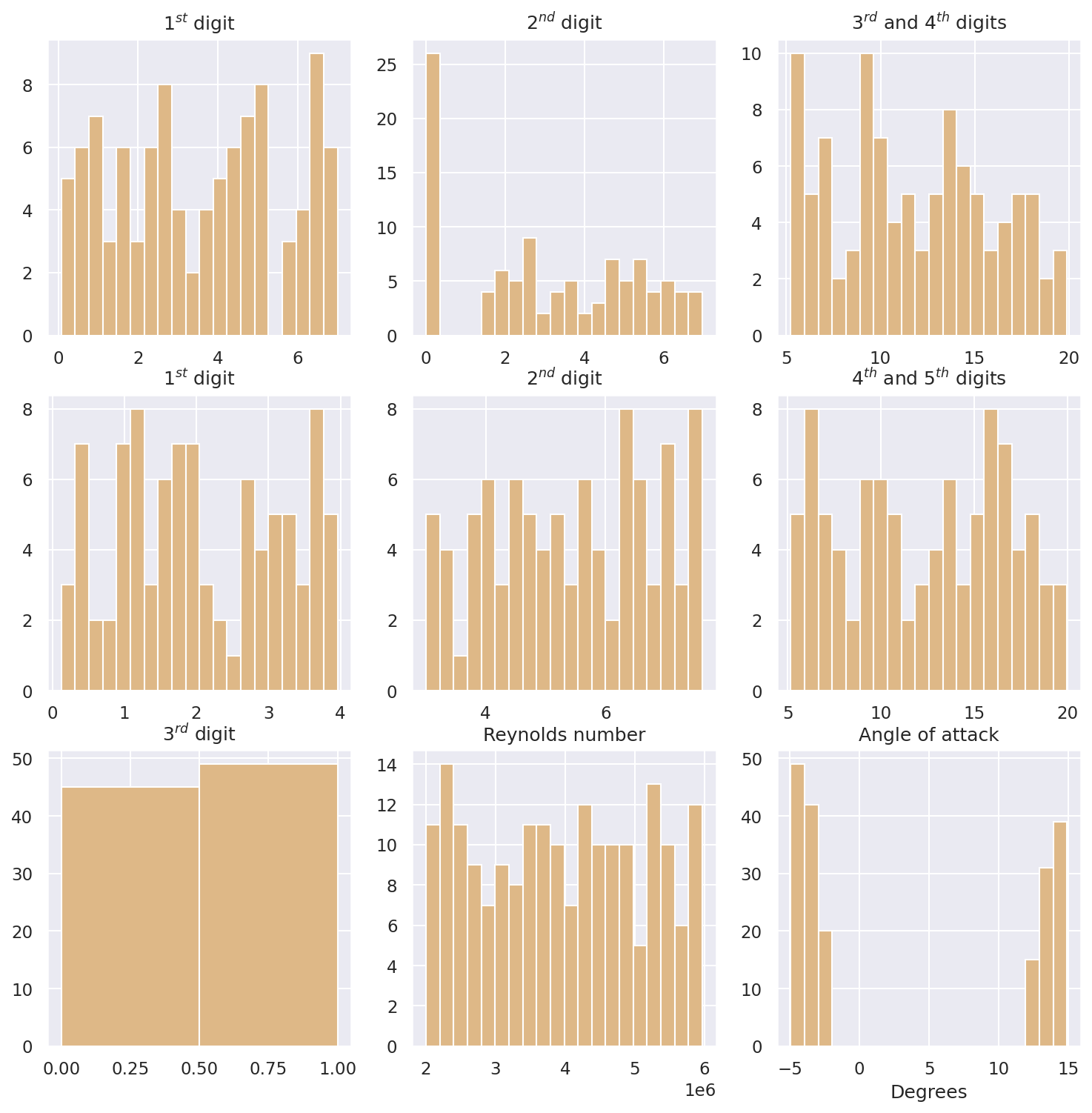}
      \caption{}
   \end{subfigure}
  \caption{Histogram of the different sampled parameters for the airfoils. For each subfigure, the top line represents the paramters of NACA 4-digits, the middle line, the parameters of the NACA 5-digits along with the left plot of the bottom line. The last two plots of the bottom lines are the histograms for the Reynolds number and the angle of attack. Each subfigure represents a different regime: (a)-(b) Full data regime (c) Scarce data regime (d)-(e) Reynolds extrapolation regime (e)-(f) Angle of attack extrapolation regime. Trainsets are in blue and testsets in yellow.}
  \label{fig:naca_stats}
\end{figure}

\paragraph{Parameter sets.}\label{sec:naca_sample} In Table~\ref{tab:sample-NACA} we give the parameter sets for the sampling. Let us underline that the digits are not necessarily integers. Also, for the values of P in the NACA 4-digits case, we actually uniformly sample in $[0, 7]$ and set the values of P strictly inferior to 1.5 to 0. This implies that the NACA 4-digits set is slightly biased towards symmetrical profiles. We assume that this bias is not of a great importance in the machine learning task. This bias could actually be leveraged to produce a smaller dataset of only symmetrical profiles where we can test the performance of invariant/equivariant models with respect to the plane symmetry of axis $y = 0$. Finally, we set the chord $c$ to \SI{1}{\meter} for all of the airfoils. In Figure~\ref{fig:naca_stats}, we show statistics of the airfoil parameters. 

\section{Meshing Procedure}\label{ap:meshing}
The construction of meshes is at the core of CFD tasks. A mesh completely determine the quality of a simulation and its characteristics. For CFD problems, the local size of cells gives the information of the resolved scales. For example, in turbulent regime, it would be impossible to simulate eddies of characteristic length smaller than your typical cell length scale. Unfortunately, it is often impossible to run direct numerical simulation (DNS) to correctly simulate all the length scale of a fluid dynamics problem as it implies a prohibitive quantity of cells in the mesh. Another technology, large eddy simulation (LES), tries to model the smallest length scales via a theoretical local filtering of the solution in order to reduce the characteristic length scale of the smallest cell needed to correctly simulate the phenomena studied. However, this can still lead to a prohibitive quantity of cells in the mesh. Finally, Reynolds-Averaged-Simulation (RAS), use the RANS equations described in section \ref{ap:RANS} to model all the length scales involved in turbulence via a theoretical ensemble averaging. This allows to recover a mean field solution which requires much less cells in the mesh. In this work, we chose to run steady-state RAS to recover mean steady-state fields around airfoils.

Another aspect of the construction of a mesh is the choice of a strategy to resolve the boundary layers close to an obstacle. There are two strategies but each are based on the value of the $y^+$. This quantity represents a local Reynolds close to the obstacle and is defined as:
\begin{align}
    \begin{cases}
        y^+ = \frac{yu_\tau}{\nu} \\
        u_\tau = \sqrt{\frac{\tau_w}{\rho}}
    \end{cases}
\end{align}
where $y$ is the distance from the wall, $\tau_w$ the magnitude of the wall shear stress, $\rho$ the density of the fluid and $\nu$ the kinematic viscosity of the fluid. In order to compute the $y^+$, experimental values on thin plates give the order of magnitude of the wall shear stress term \cite{boundary}. When we take $y$ as the height of the first cell close to the wall, we can define two strategies:
\begin{itemize}
    \item \emph{Low-Reynolds simulation:} resolve entirely the boundary layer, $y^+ < 1$
    \item \emph{High-Reynolds simulation:} model the boundary layer via a so-called wall function, $y^+ \sim 10^2$
\end{itemize}
%The second strategy makes use of wall functions that comes from a theoretical universal law of the wall. 
As we are interested in accurate force coefficients at the surface of our airfoils, we chose the first strategy to avoid modelling close to the wall. This implies that we need the maximum height $y$ of our first cells close to the wall to be smaller than $\nu/u_\tau$. In our case, we chose $y = \SI{2}{\micro\meter}$ which set the $y^+$ to be around 1 in the worst case of our design space.

\begin{figure}
    \centering
    \includegraphics[width = \textwidth]{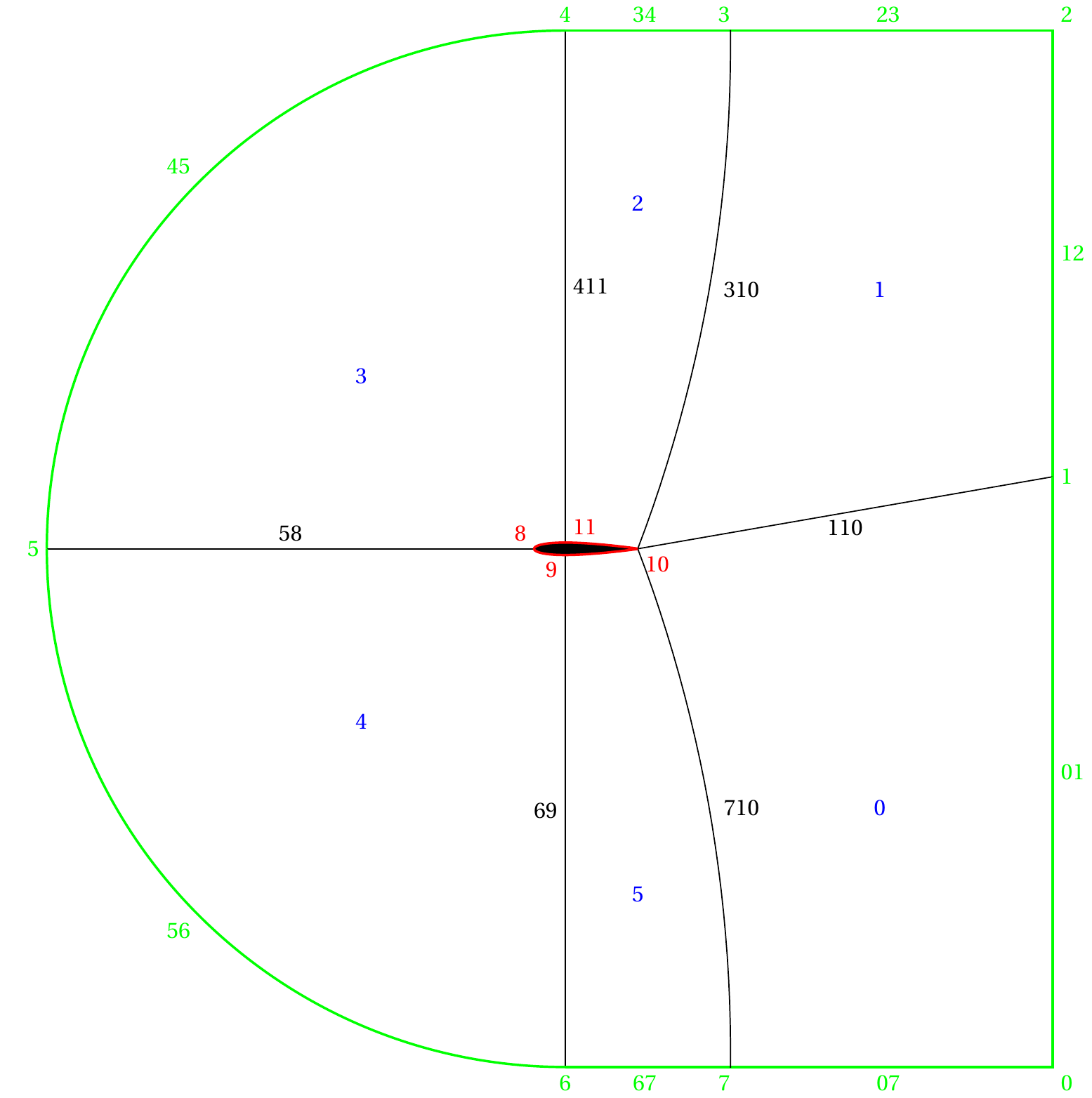}
    \caption{Scheme of the mesh template. This is a scheme for the NACA 0012 with an angle of attack of \SI{10}{\degree}. The aerofoil patch is highlighted in red, the freestream patch is highlighted in green and the internal patch is the union of the blocks 0 to 5 highlighted in blue. The indices of nodes and blocks are the same as in the \emph{blockMeshDict} file.}
    \label{fig:sch_mesh}
\end{figure}

Let us now present the mesh we use for our simulations. This is inspired by the National Aeronautics and Space Administration (NASA) mesh used in \cite{TMR} to recover experimental force coefficients on the NACA 0012 and 4412. We do not pretend to have the same quality of mesh as the NASA but we still argue that our mesh is well suited for our case. We show it in the next sections by comparing our results to experimental results. Meshes have been generated with the help of \emph{blockMesh}, a hexahedral mesh generator included in the OpenFOAM suite that works by defining blocks. With the help of a dictionnary (namely the \emph{blockMeshDict} file), we set the number of cells and the grading we want to fully determined the meshing inside each block. A scheme of how the domain is divided in multiple blocks and a result of the meshing procedure on the NACA 0012 with an angle of attack of \SI{10}{\degree} is given in Figure~\ref{fig:sch_mesh}. More precisely, as in the NASA mesh, a C-Grid domain is defined with a radius and a length of \SI{200}{\meter} (which means 200 chords here). This is smaller than the 500 chords length domain of the NASA but we found it big enough to be insensible to boundary conditions. We now use the index of the nodes, edges and blocks defined on Figure~\ref{fig:sch_mesh}. For the edges 310, 411, 58, 69 and 710, the smallest cell is of height \SI{2}{\micro\meter} as already said above and we set the expansion ratio (the length ratio between two consecutive cells) to 1.075. For the edges 01 and 12 the smallest cell is of height \SI{100}{\micro\meter} and the expansion ratio is also set to 1.075. At the upper surface of the airfoil, at the leading edge, the smallest cell is of width \SI{10}{\micro\meter} (at node 8) and we set the expansion ratio to 1.025 until roughly the maximum of camber of the airfoil (at node 11). From node 11 to node 10, an automatic expansion ratio is computed to fill the entire segment. Almost the same procedure is applied at the lower surface of the airfoil, the only difference is that, for consistency, the expansion ratio between node 9 and 10 is set such that the last cells (at node 10) is of the same width as the one at the upper surface. Edge 34 or 67 have a fix grading of 1 and edge 45 or 56 have a grading such that the width of the cell at the junction of blocks 2 and 3 or 4 and 5 are the same (this is only true at node 4 or 6 and a significant width difference can be seen at the center of edges 411 or 69). Finally edges 07, 110 and 23 have a smallest cell of the same width as for block 2 or 5 (with the same remark, this is true only at nodes 3, 7 and 10) and a grading of 1.075 is applied.

\begin{figure}
    \centering
    \includegraphics[width = \textwidth]{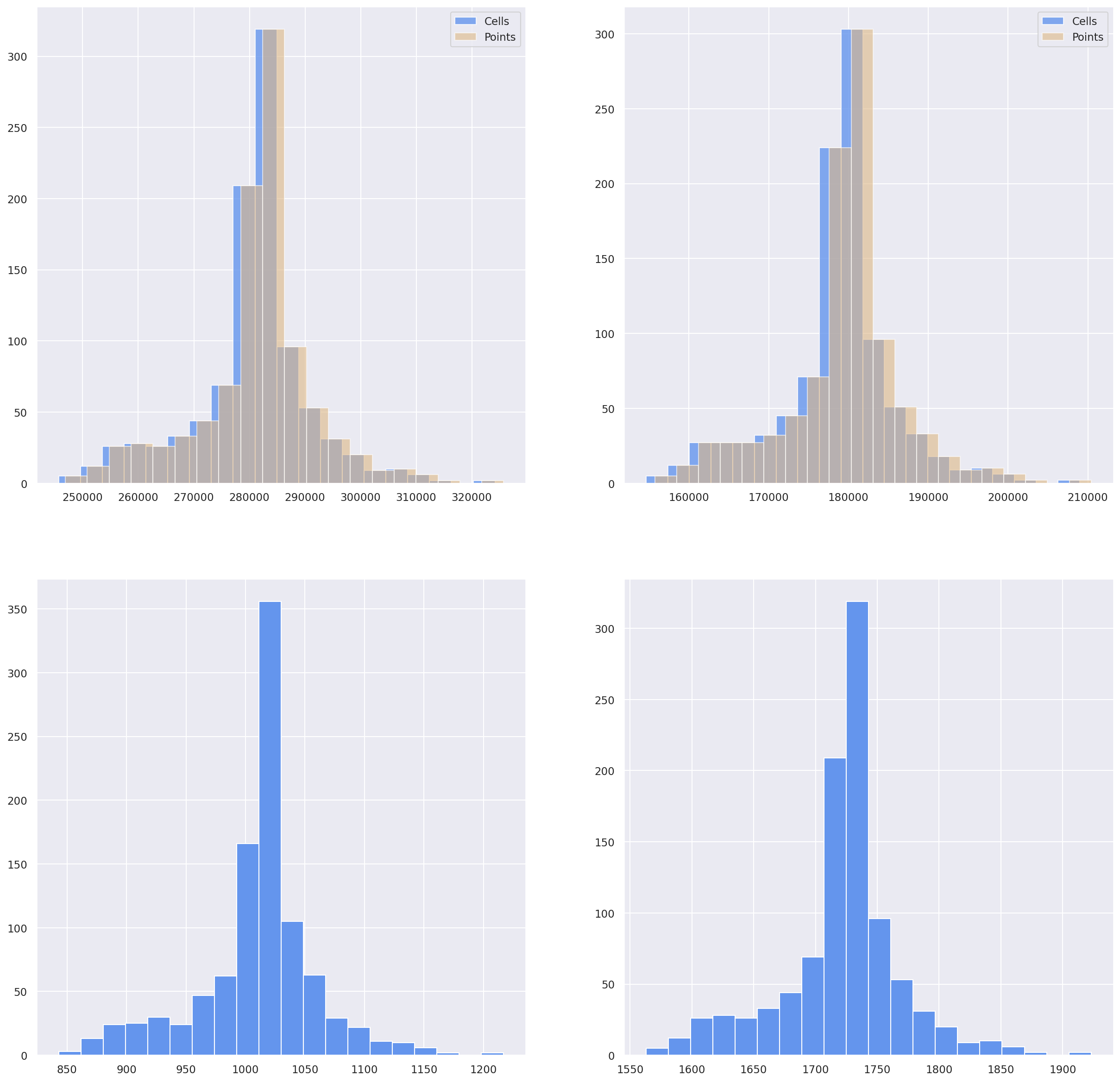}
    \caption{Histograms of the number of cells and nodes in simulations of the dataset. (top left) Number of cells and nodes in internal meshes for CFD simulations (top right) Numbers of cells and nodes in internal meshes for cropped simulations (bottom left) Number of nodes on airfoils patches (bottom right) Number of nodes on freestream patches. For the bottom plots, we only give the number of nodes as they are equal to the number of cells.}
    \label{fig:data_stats}
\end{figure}

In Figure~\ref{fig:data_stats} we give the number of cells and nodes in simulations of the dataset. We also give those quantities for the cropped simulations used for the ML tasks.

\renewcommand{\thefootnote}{\fnsymbol{footnote}}

\section{Boundary Conditions}\label{ap:boundary}
In this section, we explicit the different boundary conditions set on the different patches of the mesh. The quantities needed for a simulation depend on whether we run incompressible or compressible physics and on the turbulence model chosen. In Table \ref{tab:boundaries}, \ref{tab:bound_quant} and \ref{tab:bound_cond} we give the different OpenFOAM settings used for the different fields involved in the simulation, that are:
\begin{itemize}
    \item $U$ : ensemble averaged velocity in \si{\meter\per\second}
    \item $p$ : ensemble averaged effective pressure in \si{\pascal} (in the incompressible case the pressure is divided by the density $\rho$, $p\rightarrow p/\rho$)
    \item $\nu_t$ : kinematic turbulent viscosity in \si{\square\meter\per\second}
    \item $\Tilde{\nu}$\footnote{Only for the Spalart-Allmaras turbulent model.}: Spalart-Allmaras variable in \si{\square\meter\per\second}
    \item $k$\footnotemark[2]{} : turbulent kinetic energy in \si{\joule}
    \item $\omega$\footnotemark[2]{} : specific dissipation rate via turbulence in \si{\per\second}
    \item $T$\footnotemark[3]{} : temperature in \si{\kelvin}
    \item $\alpha_t$\footnotemark[3]{} : turbulent thermal diffusivity in \si{\square\meter\per\second}
\end{itemize}
\footnotetext[2]{Only for the $k-\omega$ SST turbulent model \cite{SST}.}
\footnotetext[3]{Only in the case of compressible simulations.}

\begin{table}
    \centering
    \begin{threeparttable}
        \caption{Boundary conditions set on the different patches of the mesh for compressible and incompressible simulations. Values of the constants are given for the air at sea level and at \SI{298.15}{\kelvin}.}
        \begin{tabular}{cccc}
            \toprule
             Fields & Internal & Aerofoil & Freestream \\
             \midrule
             $U$ & $U_\infty$ & \emph{noSlip} & \emph{freestreamVelocity} \\
             $p$ & 0\tnote{$\star$} & \emph{zeroGradient} & \emph{freestreamPressure} \\
             $\nu_t$ & $\nu$ & \emph{nutLowReWallFunction} & \emph{freestream} \\
             $\Tilde{\nu}$ & $4\nu$ & \emph{fixedValue} & \emph{freestream} \\
             $k$ & $0.001U_\infty^2/Re_L$ & \emph{fixedValue} & \emph{freesteam} \\
             $\omega$ & $5U_\infty/L$ & \emph{omegaWallFunction} & \emph{freestream} \\
             $T$ & \SI{298.15}{\kelvin} & \emph{zeroGradient} & \emph{freestream} \\
             $\alpha_t$ & $\nu_t/Pr_t$ & \emph{compressible::alphatWallFunction} & \emph{calculated} \\
             \bottomrule
        \end{tabular}
        \begin{tablenotes}
            \item [$\star$] This value has to be set to an absolute pressure value, in our case \SI{1.013e5}{\pascal}, for the compressible case.
        \end{tablenotes}
        \label{tab:boundaries}
    \end{threeparttable}
\end{table}

\begin{table}
    \caption{Definition of the quantities involved in Table \ref{tab:boundaries} and their values for the air at sea level and at a temperature of \SI{298.15}{\kelvin} (\SI{25}{\degreeCelsius}).}
    \centering
    \begin{tabular}{ccc}
        \toprule
         Quantity & Definition & Value \\
         \midrule
        $\rho$ & Density of the fluid & $\SI{1.184}{\kilogram\per\cubic\meter}$ \\
        $\nu$ & Kinematic viscosity of the fluid & $\SI{1.56e-5}{\square\meter\per\second}$ \\
        $L$ & Length of the domain & $\SI{400}{\meter}$ \\
        $U_\infty$ & Velocity at the inlet & - \\
        $Re_L$ & Reynolds number computed with $L$ & $U_\infty L/\nu$ \\
        $Pr_t$ & Turbulent Prandtl number (constant) & $0.85$ \\
        \bottomrule
    \end{tabular}
    \label{tab:bound_quant}
\end{table}

\begin{table}
    \caption{Definition of the boundary conditions involved in Table \ref{tab:boundaries} and values we use when asked by OpenFOAM. The values given for \emph{fixedValue} are the values of $k$ and $\Tilde{\nu}$ at the surface of the airfoil. The quantity $\beta_1 = 0.075$ is a constant of the $k-\omega$ model \cite{komega} and $\Delta y = \SI{2}{\micro\meter}$ the height of the first cells of the boundary layer.}
    \centering
    \begin{tabular}{ccc}
        \toprule
         Boundary condition & Definition & Value \\
         \midrule
         \emph{fixedValue} & Set the quantity to a constant & \SI{0}{\joule}/\SI{0}{\square\meter\per\second} \vspace{0.2cm} \\
        \emph{calculated} & Derived from other quantities & Internal field value \vspace{0.2cm} \\
        \emph{noSlip} & Set the velocity to 0 & - \vspace{0.2cm} \\
        & Set the field at the &  \\
        \emph{zeroGradient} & boundary to the value & - \\
        & of the internal field & \vspace{0.2cm} \\
         & Mixed boundary condition & \\
        \emph{freestream} & between \emph{fixedValue} and & Internal field value\\ 
        & \emph{zeroGradient} depending on & \\
        & the direction of the flux & \vspace{0.2cm} \\
        & Same as \emph{freestream} but & \\
        \emph{freestreamVelocity} & switches in accordance with & Internal field value\\ 
        & \emph{freestreamPressure} & \vspace{0.2cm} \\
        & Same as \emph{freestream} but & \\
        \emph{freestreamPressure} & switches in accordance with & Internal field value\\ 
        & \emph{freestreamVelocity} & \vspace{0.2cm} \\
        \emph{nutLowReWallFunction} & Set the turbulent viscosity to 0 & \SI{0}{\square\meter\per\second} \vspace{0.2cm} \\
        \emph{omegaWallFunction} & For low Reynolds simulation, & $\frac{6\nu}{\beta_1\Delta y^2}$ \\
        & equivalent to \emph{fixedValue} & \vspace{0.2cm} \\
        \emph{compressible::alphatWallFunction} & Equivalent to \emph{fixedValue} & \SI{0}{\square\meter\per\second} \\
        & with a value of $\nu_t/Pr_t$ & \\
        \bottomrule
    \end{tabular}
    \label{tab:bound_cond}
\end{table}

We do not present here the discretization schemes chosen for the simulations, nor the linear solver and hyperparameters of the SIMPLE algorithm. You can find them in the \emph{fvSchemes} and \emph{fvSolution} dictionnaries respectively. We just mention that we used the SIMPLEC \cite{SIMPLEC} algorithm for the incompressible case and the classical SIMPLE \cite{SIMPLE} one for the compressible setup as the SIMPLEC was not stable in this case.

\section{Simulation Validation}\label{ap:validation}
In this section, we test our mesh and boundary conditions on two different problems, with two turbulence models and in the compressible and incompressible settings, in order to validate the choice made in this work. To do so, we use the experimental data produced by the NASA and available on the Turbulence Modeling Resource (TMR) website of the Langley Research Center \cite{TMR} for the NACA 0012 and 4412.

\paragraph{NACA 0012 airfoil.} We compare our results with experimental data for the force coefficients done on the NACA 0012 \cite{NACA0012-1, NACA0012-2}. We restricted our study to the case of a Reynolds of 6 million for different angle of attacks (see Table XIII \cite{NACA0012-2}). In our simulations, we run an incompressible solver with the properties of the air at \SI{298.15}{\kelvin} and at sea level (see Table \ref{tab:bound_quant}) which gives an inlet velocity $U_\infty$ of \SI{93.6}{\meter\per\second} with a characteristic length equal to the chord of the airfoil (in our case \SI{1}{\meter}). The celerity of sound in this medium is taken to be \SI{346.1}{\meter\per\second}, which gives a Mach number ($Ma := U_\infty/c$) of roughly 0.27. We often set a limit a 0.3 for the Mach number in order to run incompressible simulations, and as we are close to this limit, we run incompressible and compressible simulations for an additional restricted set of angle of attacks of \SI{0}{\degree} and \SI{10}{\degree}. The pressure coefficient at the surface of the airfoil are compared to another set of experimental data (Table II of \cite{NACA0012-1}) done at Mach 0.3 and Reynolds 6 million for this two angles of attack (more precisely at angle \SI{0.0169}{\degree} and \SI{10.0254}{\degree}). Moreover, we tried with the Spalart-Allmaras and $k-\omega$ SST models of turbulence.

The pressure coefficient $c_p$ at the surface of the airfoil is a dimensionless coefficient defined as:
\begin{align}
    c_p := \frac{\Bar{p}-\Bar{p}_\infty}{q_\infty}, \qquad q_\infty := \frac{1}{2}U_\infty^2 A
\end{align}
where $\Bar{p}$ is the mean-field reduced pressure, $\Bar{p}_\infty$ the far field pressure (set to 0 in the incompressible case), $U_\infty$ is the magnitude of the inlet velocity and $A$ is the characteristic area of the problem, we take here $A = \SI{1}{\square\meter}$.

\begin{figure}
    \centering
    \includegraphics[width = \linewidth]{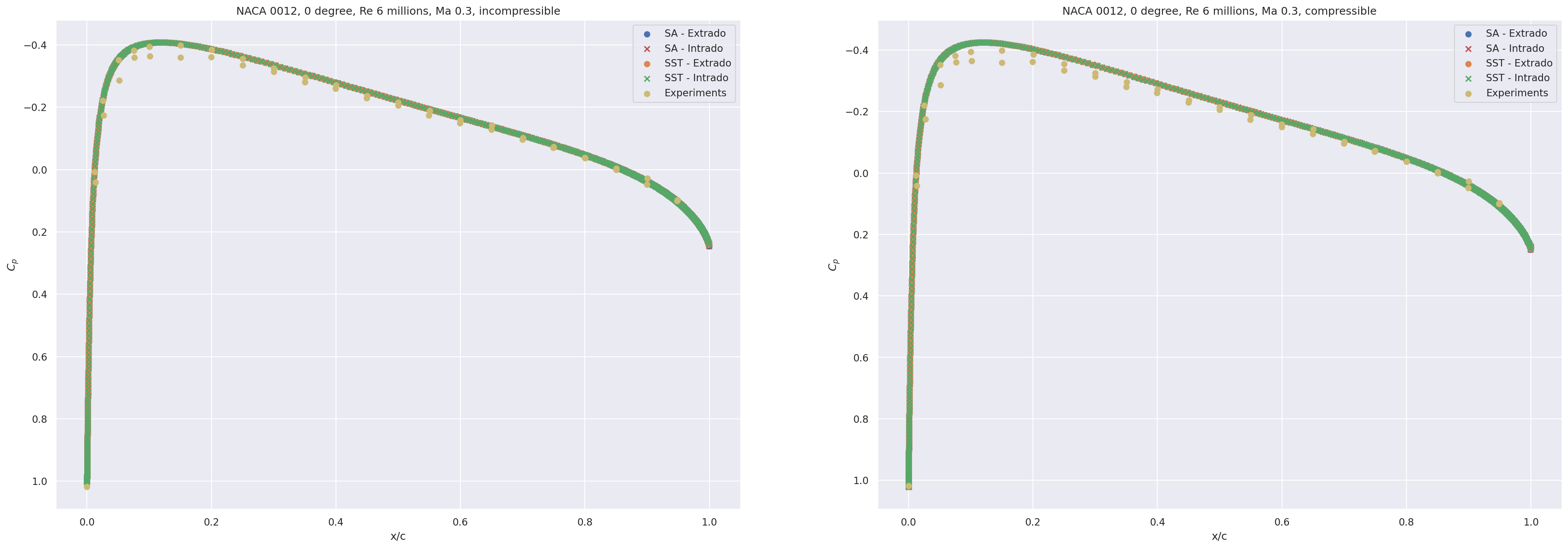}
    \caption{Pressure coefficient at the surface of the airfoil for the NACA 0012 at an angle of attack of \SI{0}{\degree} in the incompressible (left) and compressible (right) cases for the Spalart-Allamaras,  $k-\omega$ SST models and the experiments with respect to the abscissas in chord length. The points on the upper and lower surfaces are given in different colors for the simulations.}
    \label{fig:cp_0012_0}
\end{figure}

\begin{figure}
    \centering
    \includegraphics[width = \linewidth]{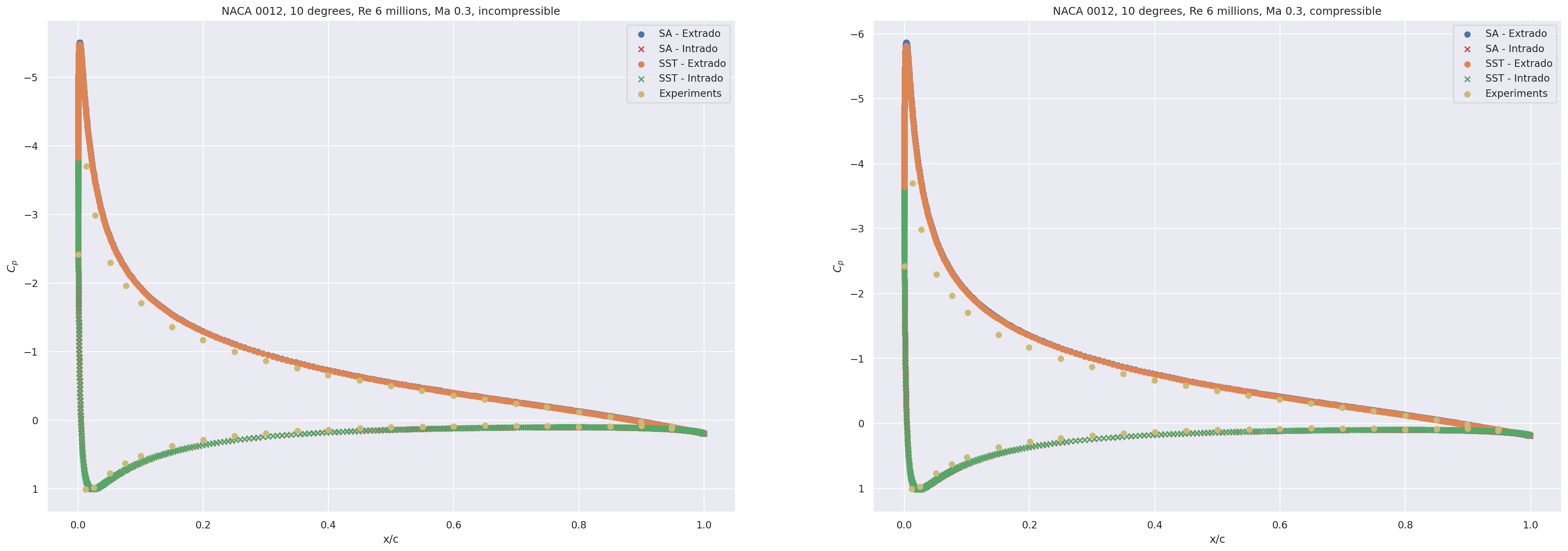}
    \caption{Pressure coefficient at the surface of the airfoil for the NACA 0012 at an angle of attack of \SI{10}{\degree} in the incompressible (left) and compressible (right) cases for the Spalart-Allamaras,  $k-\omega$ SST models and the experiments with respect to the abscissas in chord length. The points on the upper and lower surfaces are given in different colors for the simulations.}
    \label{fig:cp_0012_10}
\end{figure}

In Figure \ref{fig:cp_0012_0} and \ref{fig:cp_0012_10}, the surface pressure coefficient is given and we see no significant difference between the two models nor between the compressible and incompressible cases. All the simulations are in good agreement with the experiments.

\begin{figure}
    \centering
    \includegraphics[width = \linewidth]{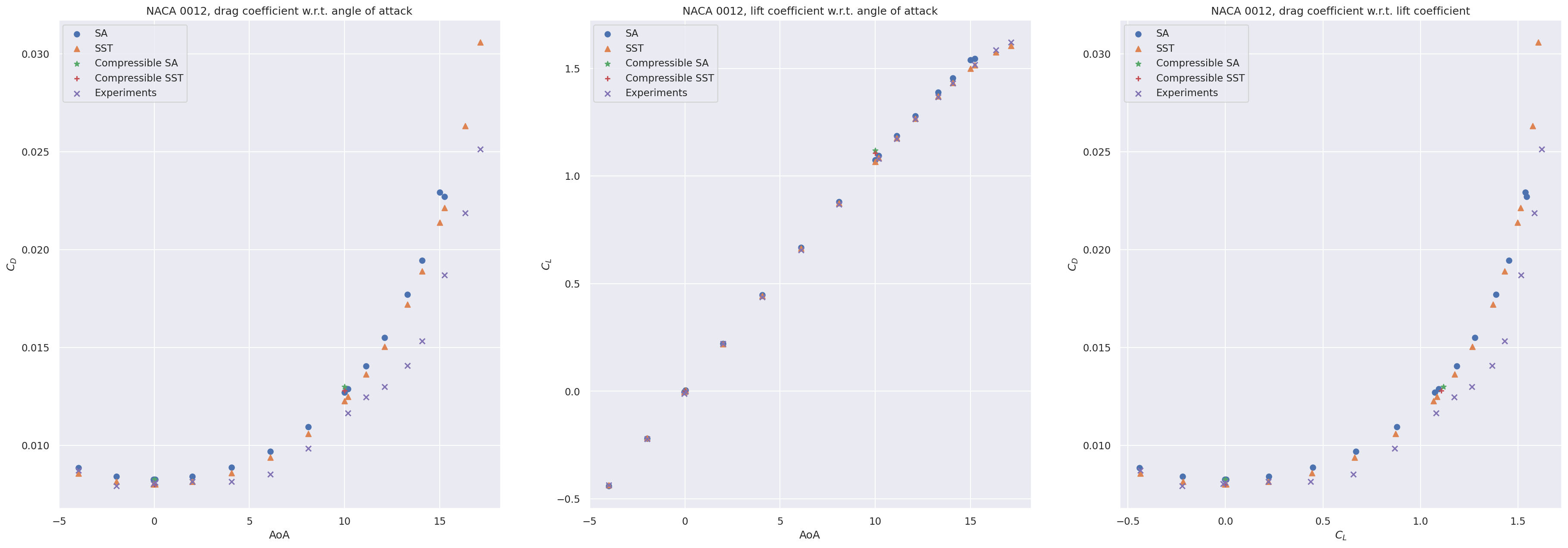}
    \caption{Drag and lift coefficients with respect to angle of attacks and to each other in the case of the NACA 0012. The compressible simulations have only be done at \SI{0}{\degree} and \SI{10}{\degree}.}
    \label{fig:force_0012}
\end{figure}

In Figure \ref{fig:force_0012} are displayed the drag and lift coefficients with respect to angle of attacks and the drag coefficient with respect to the lift coefficient for the two models and the experiments. In the compressible case, only \SI{0}{\degree} and \SI{10}{\degree} have been simulated for time and stability reasons, no significant differences are present with the incompressible simulations. In the incompressible case, a missing point in the plot means that the simulation was unstable and we did not manage to make it converge correctly. We can see that both compressible and incompressible solver gives a slightly over estimated drag with respect to experiments, this is in agreement with the TMR. We also see that the $k-\omega$ SST model is more stable than the Spalart-Allmaras model, we noticed a faster convergence for the first too. Finally, the $k-\omega$ SST model fits better the experiments than the Spalart-Allamaras model. In total, both model are in good agreement with experimental data but the $k-\omega$ SST model looks more stable, faster to converge and more accurate than the Spalart-Allamaras.

From this point, we only run incompressible simulations as this validation case showed no distinctions between compressible and incompressible simulations. We now test our setup on another validation case in order to chose between the two turbulence models.

\paragraph{NACA 4412 airfoil.} In this setup, the experimental data \cite{NACA4412} are done with a NACA 4412 at an angle of attack of \SI{13.87}{\degree} and a Reynolds number of 1.52 million. The values of the experimental data are the one given on the NACA 4412 page of the TMR. The values found on this website are slightly different from the one found in the original papers, moreover, the normalization factor for the pressure coefficient is computed with a reference velocity $U_{ref}$ of roughly $0.93U_\infty$ but, as in the NASA simulations, the results better fit when using a normalization factor computed with $U_\infty$. This is underlined on the TMR page and incite us to take this validation case only as a qualitative validation.

\begin{figure}
    \centering
    \includegraphics[width = \linewidth]{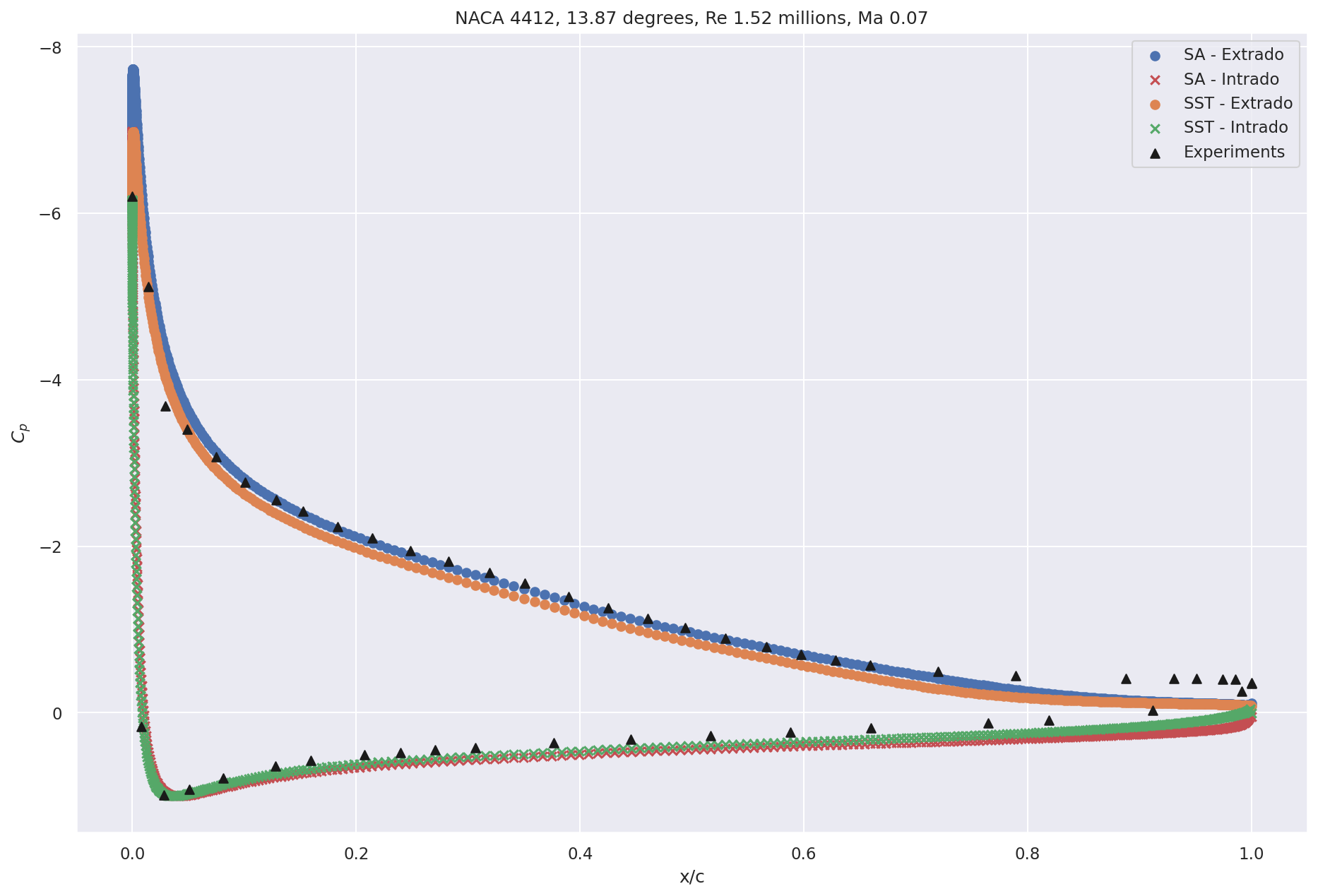}
    \caption{Pressure coefficient at the surface of a NACA 4412 with an angle of attack of \SI{13.87}{\degree} and a Reynolds of 1.52 million. The normalization of the pressure coefficient is computed with the magnitude of the inlet velocity $U_\infty$. Are displayed the experimental data, the Spalart-Allmaras and $k-\omega$ SST models incompressible results.}
    \label{fig:cp_4412}
\end{figure}

In Figure \ref{fig:cp_4412}, the pressure coefficient is given with a normalization factor computed with the magnitude of the inlet velocity $U_\infty$. Both turbulence models results are in good agreement with the experiments. 

\begin{figure}
    \centering
    \includegraphics[height = 0.9\textheight]{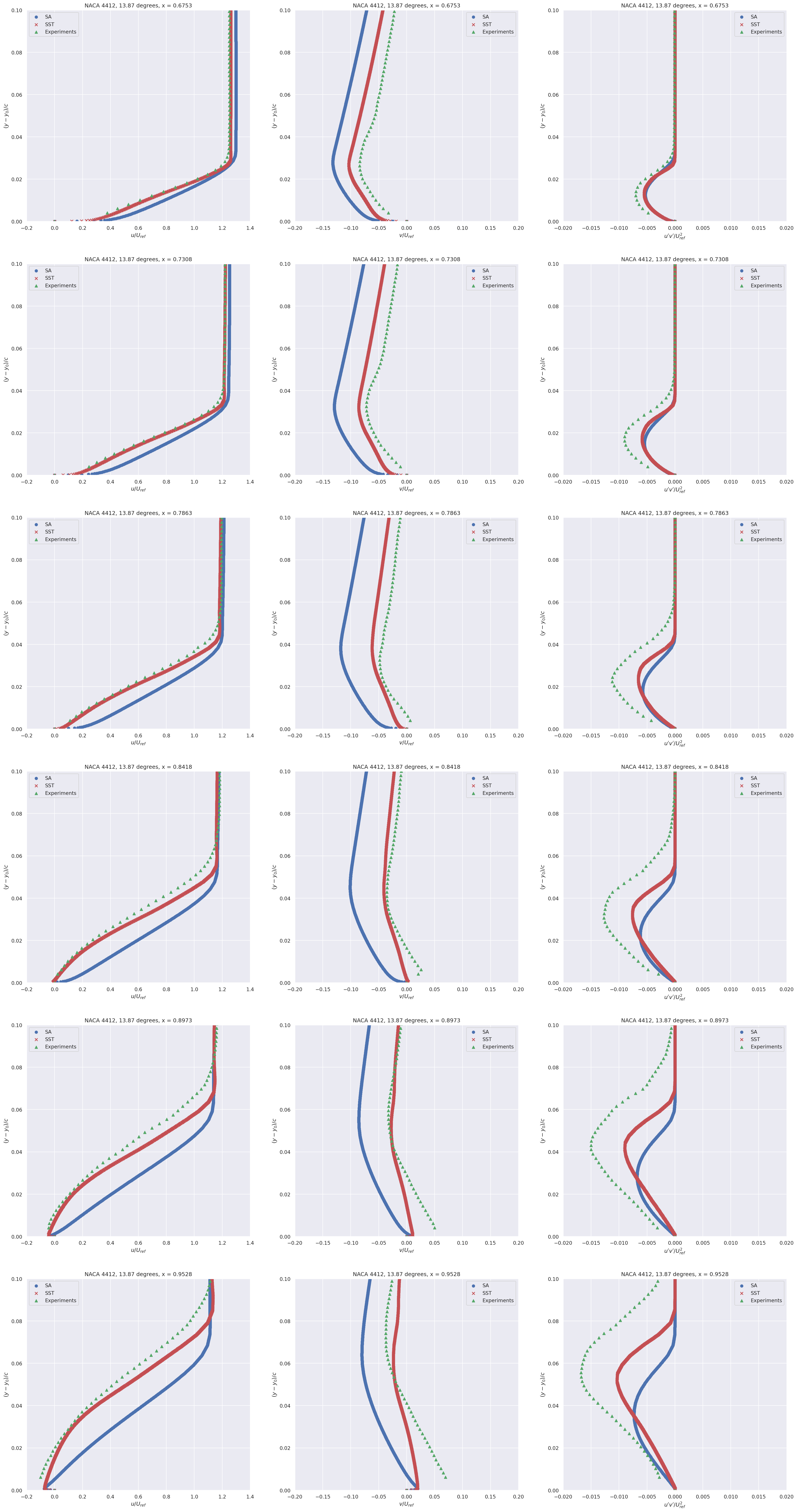}
    \caption{Boundary layer velocity components and shear Reynolds stress for different point at the surface of the NACA 4412 at a Reynolds number of 1.52 million. Each quantity is normalized either by $U_{ref}$ or $U_{ref}^2$. The ordinate has to be understand as the distance to the given point at the surface of the airfoil following the normal direction.}
    \label{fig:bl_4412}
\end{figure}

In Figure \ref{fig:bl_4412}, we look at the boundary layer of the airfoil at different abscissas. Here the $x$ and $y$ components of the velocity (denoted by $u$ and $v$ respectively) are normalized by $U_{ref}$ and the term $u'v'$, corresponding to the shear stress term of the Reynolds stress tensor, is normalized by $U_{ref}^2$. We start each plot at a given point at the surface of the airfoil and take the direction of the normal of the airfoil at this point. Hence, the name $(y - y_0)/c$ for the ordinate of the plot has to be understand as the distance to the airfoil in the normal direction in chord length. Both turbulence models have difficulties to predict correctly the experimental data, this behaviour has already been pointed out in the TMR study of the NACA 4412 and our results are in good agreement with theirs. Moreover, the $k-\omega$ SST model seems to give more realistic results than the Spalart-Allmaras one.

In total, our simulations on the NACA 0012 and 4412 are in good (at least qualitatively) agreement with the experiments. The incompressible $k-\omega$ SST model setup seems the best candidate for fast, stable and high fidelity simulations. In this work, we keep this setup to generate the dataset.

\section{Models architecture}\label{ap:models}
For all of the tasks, the same architecture is used in addition with the same hyperparameters. Each model is preceded by an encoder and followed by a decoder both defined as MLP with ReLU activation function, no batch normalization, and with $7-64-64-8$ and $8-64-64-4$ neurons respectively, meaning an dimension of encoding of 8. Those encoder and decoder are trained together with the chosen model.

\paragraph{Multi-Layer Perceptron.} The first baseline is another MLP with ReLU activation function and batch normalization before the activation. It has $8-64-64-64-8$ neurons.

\paragraph{GraphSAGE.} The GraphSAGE acts on a radius graph of 32000 nodes and radii \SI{5}{\centi\meter}. It is defined with 3 hidden layers and 64 hidden features per node.

\paragraph{PointNet.} The PointNet is copied from the segmentation task of \cite{qi2016pointnet}. We chose 8 neurons as a base number and we did not include any batch normalization nor dropout as it was performing badly with.

\paragraph{Graph U-Net.} For the Graph U-Net, we defined it with five scales, downsampling by half at each scale and multiplying by two the number of features at each scales. The radii of the radius graphs are \SI{5}{\centi\meter}, \SI{20}{\centi\meter}, \SI{50}{\centi\meter}, \SI{1}{\meter} and \SI{10}{\meter}. The last radii is chosen such that the graph at the coarsest scale is fully connected. Each of those radius graphs have a limit of 64 neighbors per node. For the downsampling, we did not use the gPool method presented in the historical paper \cite{gunet} and replaced it by a random downsampling over the remaining nodes, recreating a radius graph afterwards. This leads to better results. On the upward pass, we chose to aggregate the different informations from the skip connection and the preceding scale by concatenating the features. Finally, we chose to start with 8 features at the finest scale.

The learning rate for all of those experiments is set with a one-cycle cosine \cite{onecyclelr}  rate of maximum 0.001, simulations are fed one by one to the different models during training (\emph{i.e.} 32000 nodes with an associated radius graph when needed) and the number of epochs is chosen such that for each task, we have the same number of gradient updates:
\begin{itemize}
    \item \emph{Full data regime:} 400 epochs
    \item \emph{Scarce data regime:} 1600 epochs
    \item \emph{Reynolds extrapolation regime:} 635 epochs
    \item \emph{Angle of attack extrapolation regime:} 398 epochs
\end{itemize}

Ultimately, the different models are trained on 90\% of the predefined training set of those different regime, the last 10\% have been used as a validation set.

\section{Additional Results}\label{ap:results}
% In this section, we present the results of the full and scarce data regimes.
In this section, we treat the results of the three remaining machine learning tasks. Those tasks are not less important than the full data regime treated in the main paper, they are actually more important than the latter. Those regime better represent the challenges of real-life problems as the data is often lacking and extrapolation is often sought. We first give the missing results in the full data regime and then present the three other regimes.

\begin{figure}
    \centering
    \includegraphics[width = \linewidth]{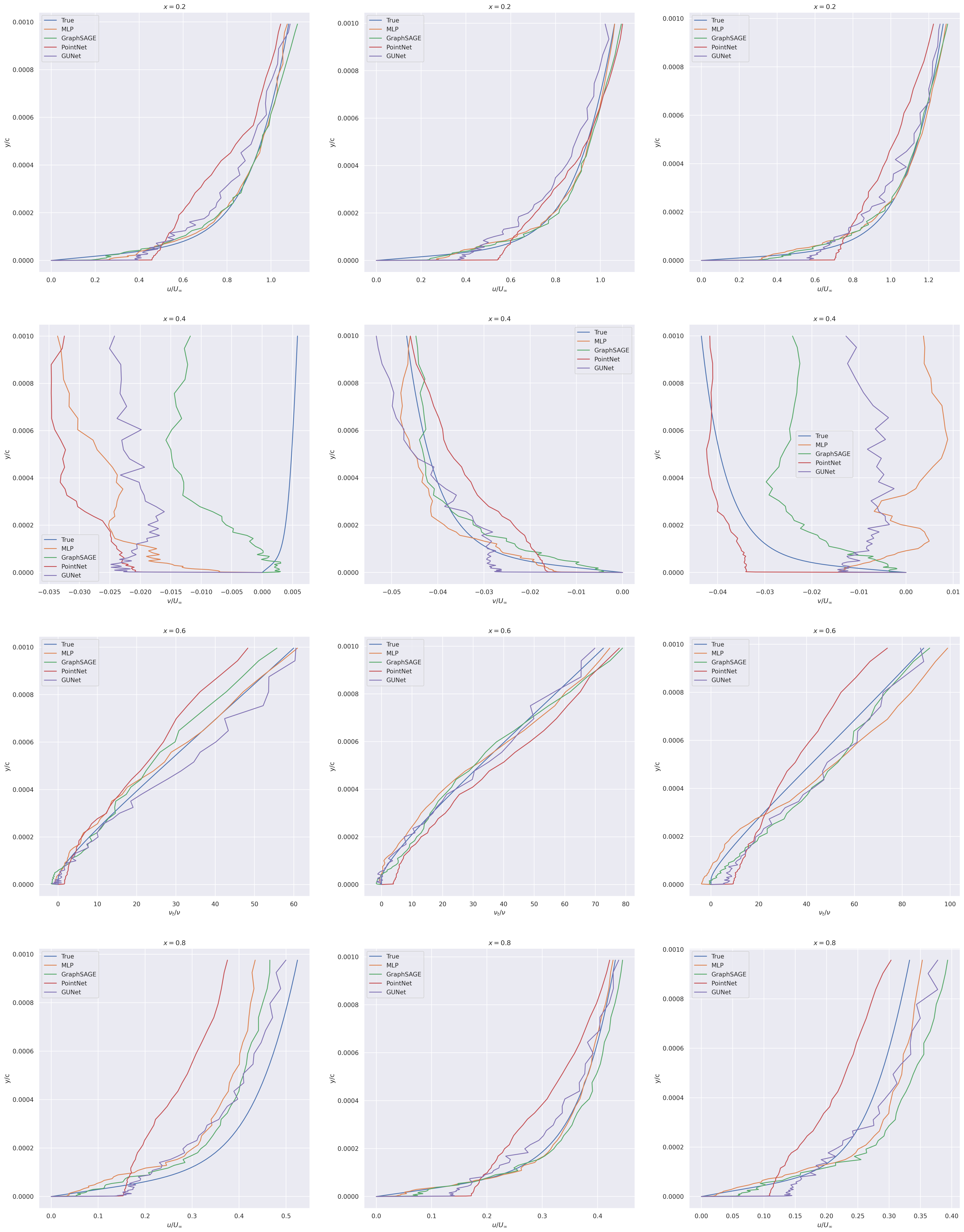}
    \caption{Comparison of the predicted boundary layers profiles on three random test geometries at different abscissas in the full data regime with respect to the true ones. Each column of plots represent a different airfoil and each line of plots represent a different abscissas. The $x$ and $y$ component of the velocity are denoted by $u$ and $v$ respectively and the turbulent viscosity is denoted by $\nu_t$. Each quantity is normalized either by $u_\infty$ the inlet velocity magnitude or $\nu$ the fluid viscosity.}
    \label{fig:full_bl}
\end{figure}

\begin{figure}
    \centering
    \includegraphics[width = \linewidth]{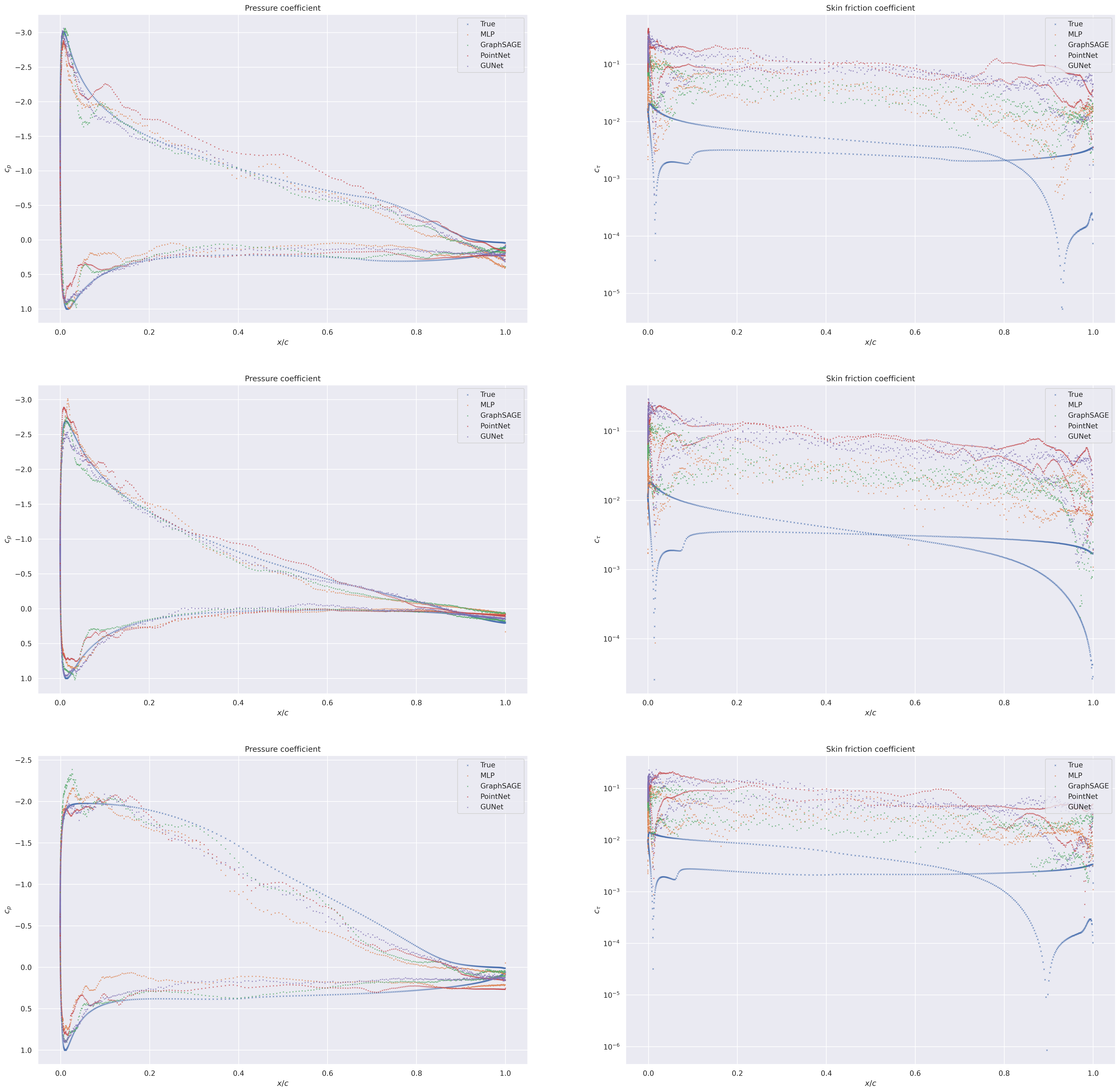}
    \caption{Comparison of the predicted surface coefficients profiles on three random test geometries in the full data regime with respect to the true one. (left) Surface coefficient $c_p$ (right) Skin friction coefficient $c_\tau$. Each line of plots represents a different airfoil. Skin friction coefficient plots are given in log scale.}
    \label{fig:full_surface}
\end{figure}

\paragraph{Full data regime.} In Figure~\ref{fig:full_surface} we show the pressure and skin friction coefficients distributions over the surface of three randomly chose airfoils in the test set. In Figure~\ref{fig:full_bl}, we present the $x$ and $y$ components of the velocity distribution in the boundary layer at the upper surface of the same three airfoils. Those velocity profiles are given at four abscissas: $x = \SI{0.2}{\meter}$, $x = \SI{0.4}{\meter}$, $x = \SI{0.6}{\meter}$ and $x = \SI{0.8}{\meter}$.

As we said in Section~\ref{sec:results}, the wall shear stress is largely overestimated, this behaviour is understood via the boundary layer velocity profiles where the first inferred point after the surface is badly approximated. This leads to high errors on the drag coefficient.

\paragraph{Scarce data regime.} In this regime, we test on the same test set as the full data regime but we trained with only two hundred simulations instead of eight hundreds.

\begin{figure}
  \centering
  \includegraphics[width = \linewidth]{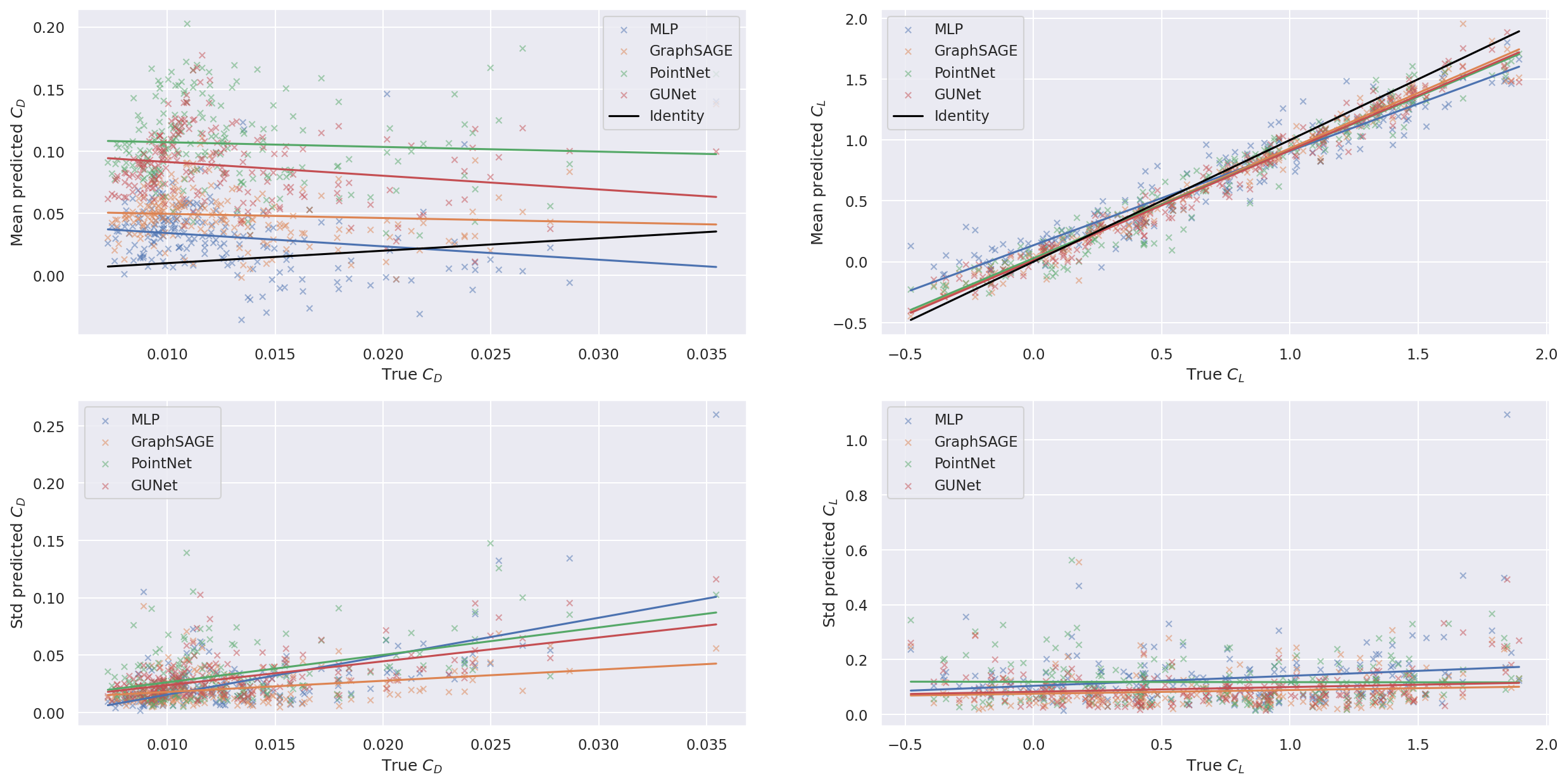}
  \caption{Predicted drag (left) and lift (right) coefficients with respect to the true ones in the scarce data regime. The mean (top) and standard deviation (bottom) of each point on the five copy of the trained models are separated for sake of readability. A linear regression is done for each point cloud in order to highlight linear trends. On the top plots, the Identity graph is given in black for comparison.}
  \label{fig:coefs_scarce}
\end{figure}

\begin{figure}
  \centering
  \includegraphics[width = \linewidth]{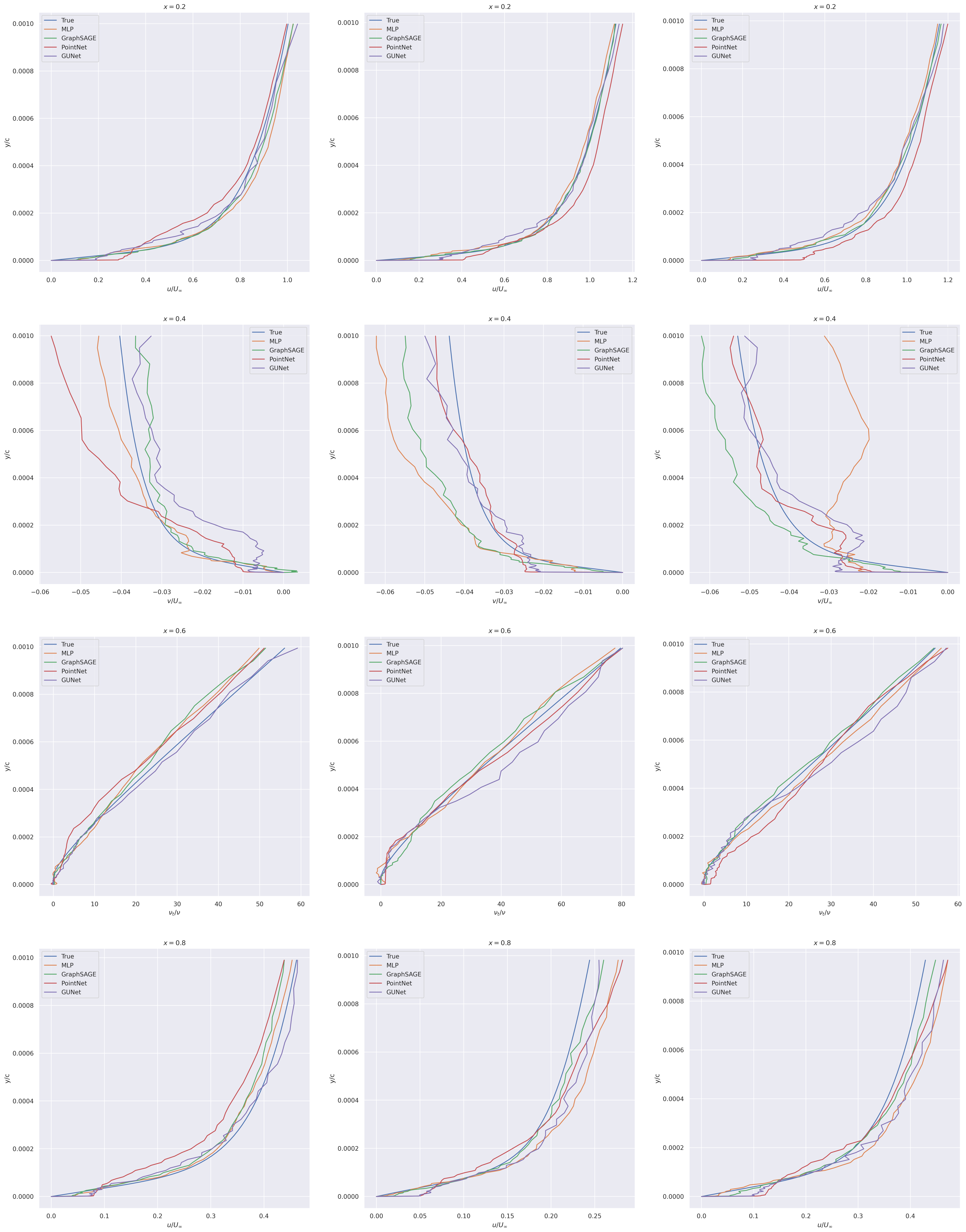}
  \caption{Comparison of the predicted boundary layers profiles on three random test geometries at different abscissas in the scarce data regime with respect to the true ones. Each column of plots represent a different airfoil and each line of plots represent a different abscissas. The $x$ and $y$ component of the velocity are denoted by $u$ and $v$ respectively and the turbulent viscosity is denoted by $\nu_t$. Each quantity is normalized either by $u_\infty$ the inlet velocity magnitude or $\nu$ the fluid viscosity.}
  \label{fig:bl_scarce}
\end{figure}

\begin{figure}
  \centering
  \includegraphics[width = \linewidth]{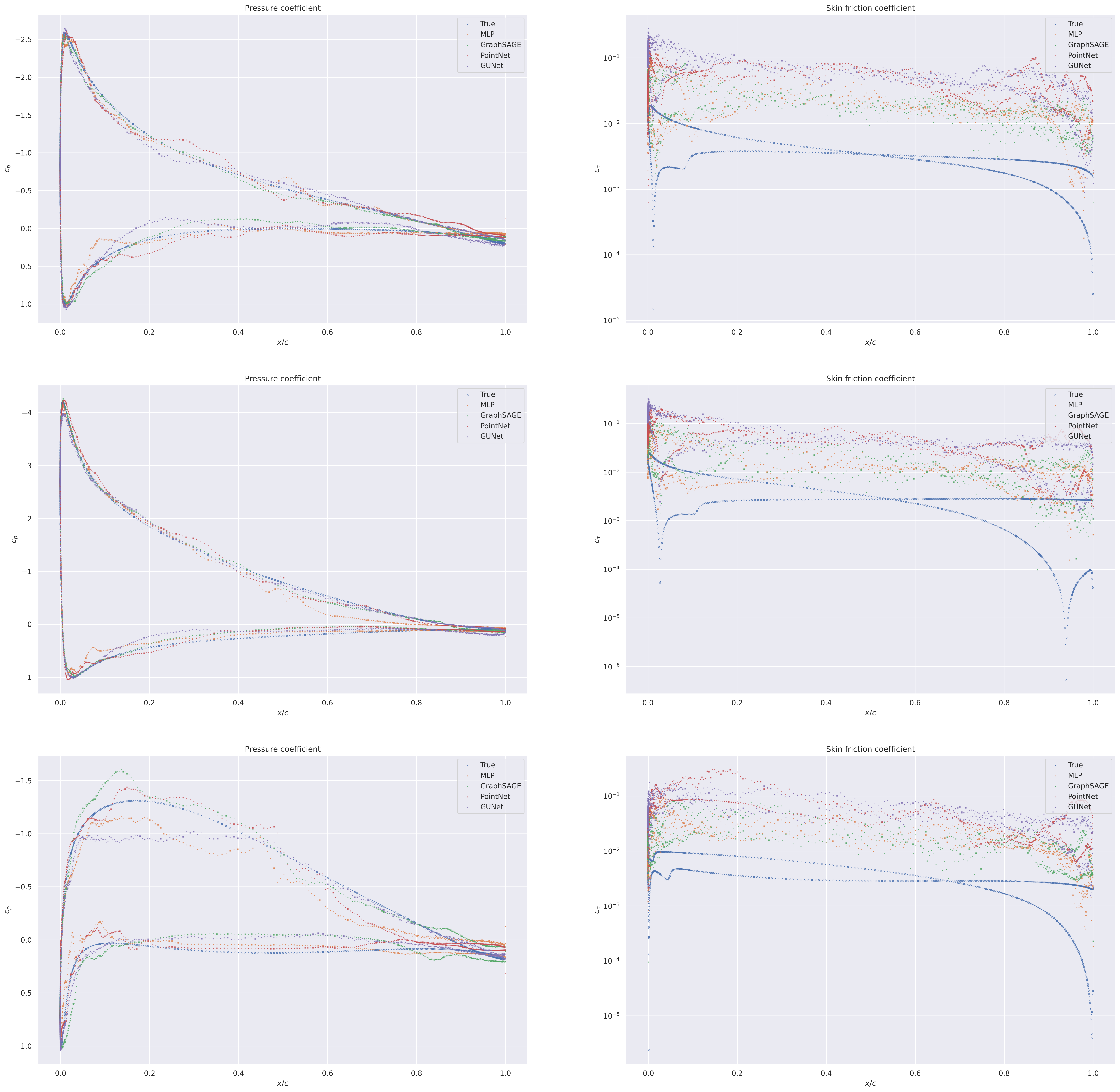}
  \caption{Comparison of the predicted surface coefficients profiles on three random test geometries in the scarce data regime with respect to the true one. (left) Surface coefficient $c_p$ (right) Skin friction coefficient $c_\tau$. Each line of plots represents a different airfoil. Skin friction coefficient plots are given in log scale.}
  \label{fig:surf_scarce}
\end{figure}

In Table~\ref{tab:MSE_scarce}, we give the MSE over the volume and at the surface of airfoils for the different regressed fields. In Table~\ref{tab:spear_scarce} we give the mean relative errors on the force coefficient and the Spearman's rank correlation coefficient. In Figure~\ref{fig:coefs_scarce} we plot the predicted force coefficients with respect to the true coefficients. In Figure~\ref{fig:bl_scarce}, we plot the velocity and turbulent viscosity profiles in the boundary layer for randomly chosen test geometries and in Figure~\ref{fig:surf_scarce} the surface coefficients for the same geometries. 

\begin{table}
  \caption{Mean squared error on the different normalized fields for an MLP and standard GDL baselines on the test set in the scarce data regime. Only the reduced pressure is given on the surface as the other quantities are null via the boundary conditions. Those quantities are directly regressed by the models.}
  \label{tab:MSE_scarce}
  \centering
  \begin{tabular}{cccccc}
    \toprule
    Model & \multicolumn{4}{c}{Volume} & Surface  \\
    & $\Bar{u}_x$ ($\times 10^{-2}$) & $\Bar{u}_y$ ($\times 10^{-2}$) & $\Bar{p}$ ($\times 10^{-2}$) & $\nu_t$ ($\times 10^{-2}$) & $\Bar{p}$ ($\times 10^{-1}$) \\
    \midrule
    MLP & 1.65$\pm$0.03 & \textbf{1.45}$\pm$\textbf{0.07} & 3.90$\pm$0.57 & 5.01$\pm$0.76 & 2.19$\pm$0.53 \\
    GraphSAGE & \textbf{1.46}$\pm$\textbf{0.13} & 1.45$\pm$0.12 & 4.70$\pm$0.80 & 6.11$\pm$0.79 & 1.95$\pm$0.34 \\
    PointNet & 3.11$\pm$0.30 & 2.78$\pm$0.39 & \textbf{3.29}$\pm$\textbf{1.05} & 5.58$\pm$2.36 & 1.83$\pm$0.41 \\
    Graph U-Net & 1.75$\pm$0.19 & 1.83$\pm$0.18 & 3.39$\pm$0.84 & \textbf{4.30}$\pm$\textbf{1.00} & \textbf{1.47}$\pm$\textbf{0.35} \\
    \bottomrule
  \end{tabular}
\end{table}

\begin{table}
  \caption{Relative errors (Spearman's rank correlation) for the predicted drag coefficient $C_D$ ($\rho_D$) and lift coefficient $C_L$ ($\rho_L$) in the scarce data regime. We want the Spearman's correlation to be close to one. Those quantities are computed as a post processing from the unnormalized regressed fields.}
  \label{tab:spear_scarce}
  \centering
  \begin{tabular}{ccccccc}
    \toprule
    Model & \multicolumn{2}{c}{Relative error} & \multicolumn{2}{c}{Spearman's correlation}  \\
     & $C_D$ & $C_L$ & $\rho_D$ & $\rho_L$ \\
    \midrule
    MLP & \textbf{2.95}$\pm$\textbf{0.14} & 0.66$\pm$0.16 & -0.24$\pm$0.08 & 0.923$\pm$0.026 \\
    GraphSAGE & 3.50$\pm$1.00 & \textbf{0.39}$\pm$\textbf{0.10}& -0.14$\pm$0.18 & \textbf{0.981}$\pm$\textbf{0.006} \\
    PointNet & 8.35$\pm$1.39 & 0.59$\pm$0.13 & -0.05$\pm$0.27 & 0.949$\pm$0.019 \\
    Graph U-Net & 6.87$\pm$1.80 & 0.42$\pm$0.13 & -0.10$\pm$0.23 & 0.976$\pm$0.009 \\
    \bottomrule
  \end{tabular}
\end{table}

\paragraph{Reynolds extrapolation regime.} In this regime, we test on out of distribution Reynolds number.

\begin{figure}
  \centering
  \includegraphics[width = \linewidth]{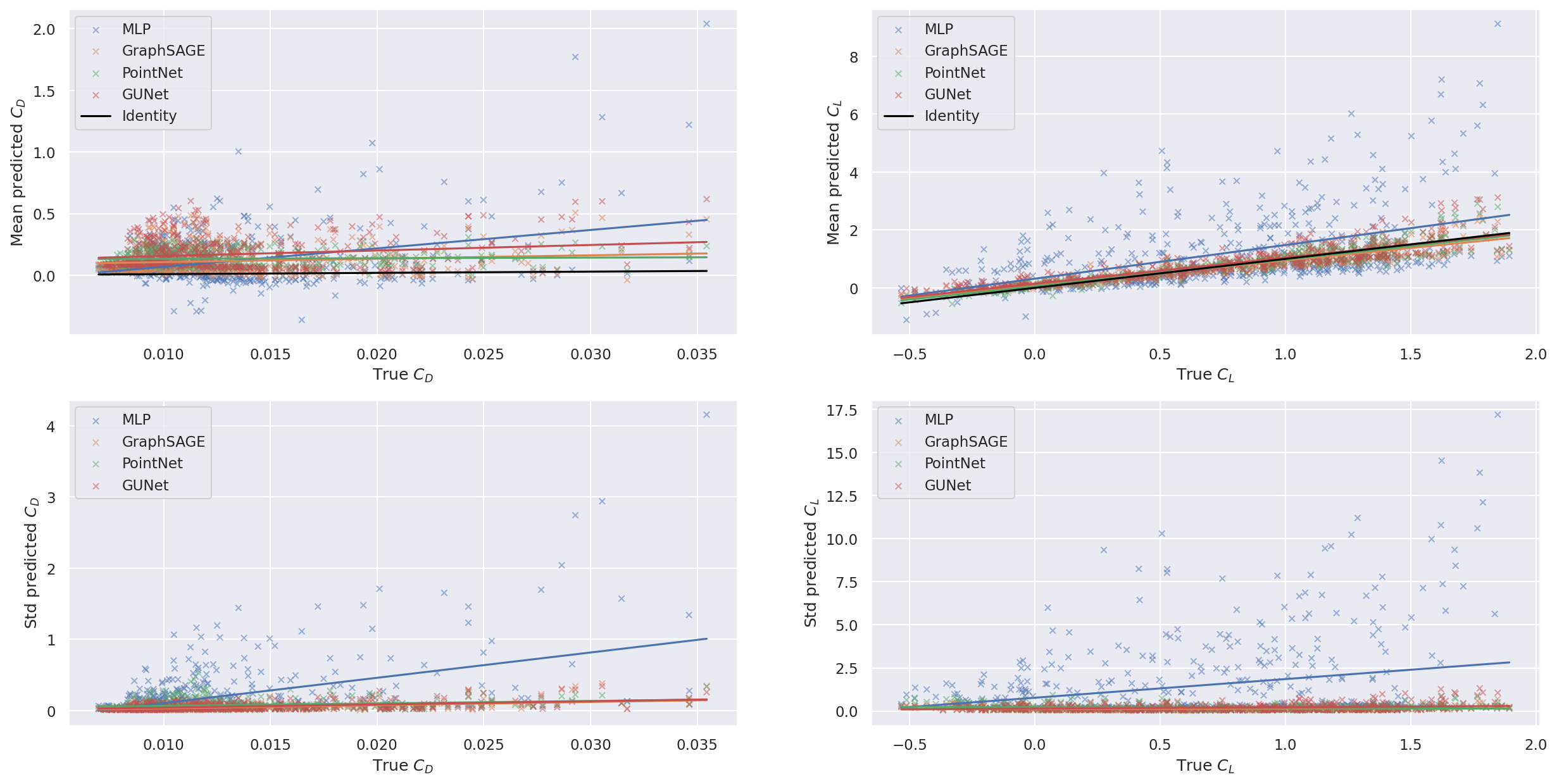}
  \caption{Predicted drag (left) and lift (right) coefficients with respect to the true ones in the Reynolds extrapolation regime. The mean (top) and standard deviation (bottom) of each point on the five copy of the trained models are separated for sake of readability. A linear regression is done for each point cloud in order to highlight linear trends. On the top plots, the Identity graph is given in black for comparison.}
  \label{fig:coefs_reynolds}
\end{figure}

\begin{figure}
  \centering
  \includegraphics[width = \linewidth]{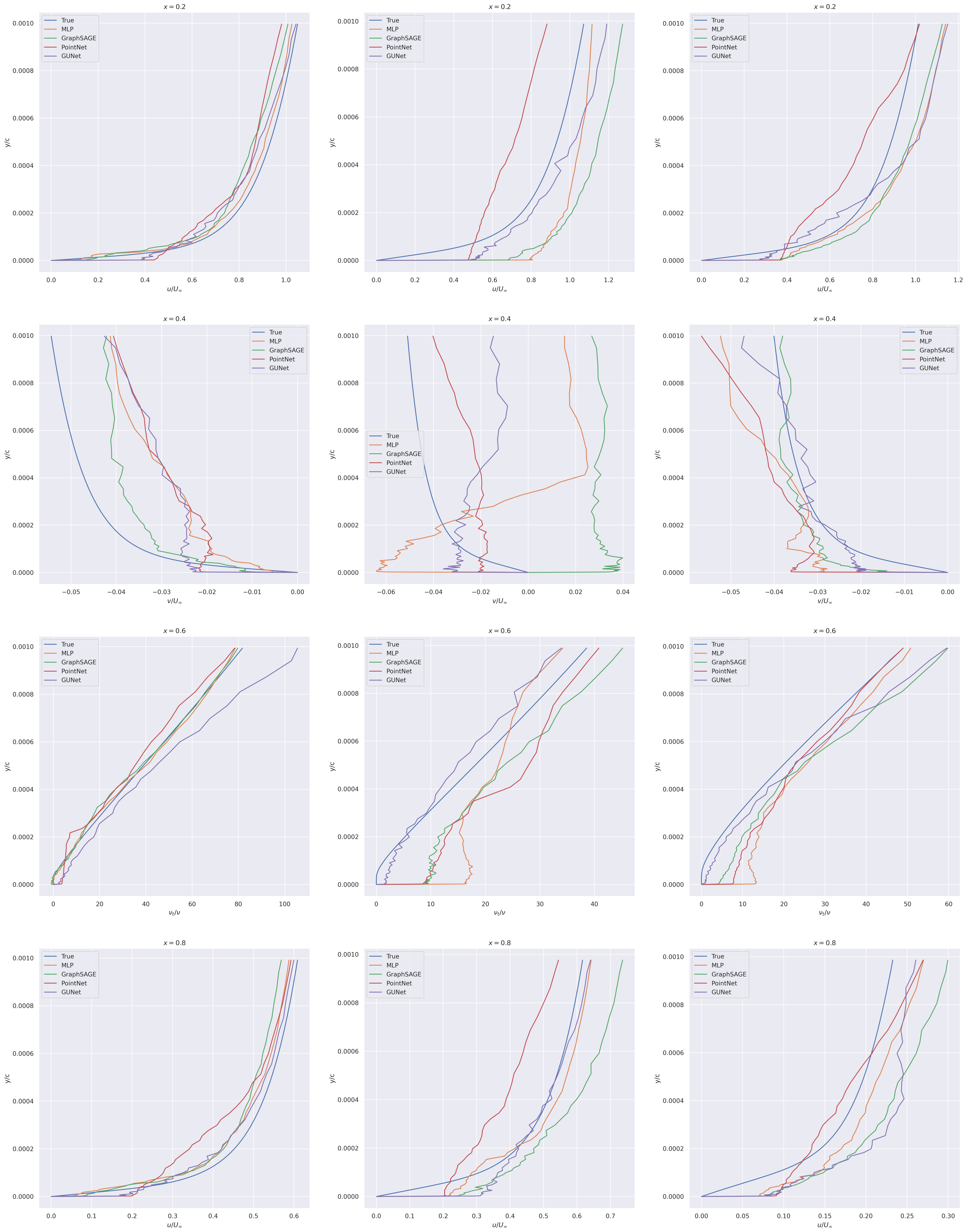}
  \caption{Comparison of the predicted boundary layers profiles on three random test geometries at different abscissas in the Reynolds extrapolation regime with respect to the true ones. Each column of plots represent a different airfoil and each line of plots represent a different abscissas. The $x$ and $y$ component of the velocity are denoted by $u$ and $v$ respectively and the turbulent viscosity is denoted by $\nu_t$. Each quantity is normalized either by $u_\infty$ the inlet velocity magnitude or $\nu$ the fluid viscosity.}
  \label{fig:bl_reynolds}
\end{figure}

\begin{figure}
  \centering
  \includegraphics[width = \linewidth]{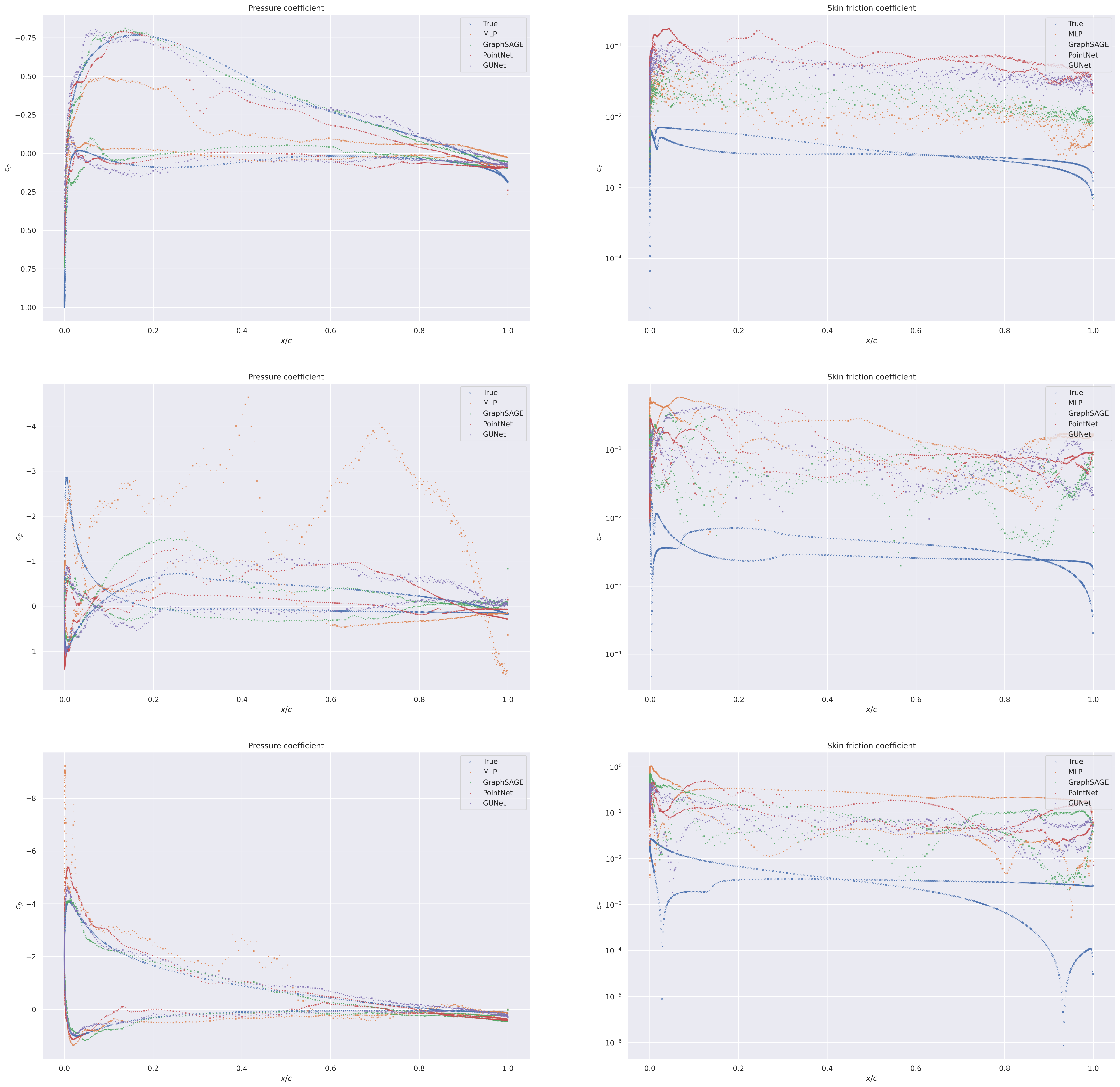}
  \caption{Comparison of the predicted surface coefficients profiles on three random test geometries in the Reynolds extrapolation regime with respect to the true one. (left) Surface coefficient $c_p$ (right) Skin friction coefficient $c_\tau$. Each line of plots represents a different airfoil. Skin friction coefficient plots are given in log scale.}
  \label{fig:surf_reynolds}
\end{figure}

In Table~\ref{tab:MSE_reynolds}, we give the MSE over the volume and at the surface of airfoils for the different regressed fields. In Table~\ref{tab:spear_reynolds} we give the mean relative errors on the force coefficient and the Spearman's rank correlation coefficient. In Figure~\ref{fig:coefs_reynolds} we plot the predicted force coefficients with respect to the true coefficients. In Figure~\ref{fig:bl_reynolds}, we plot the velocity and turbulent viscosity profiles in the boundary layer for randomly chosen test geometries and in Figure~\ref{fig:surf_reynolds} the surface coefficients for the same geometries. 

\begin{table}
  \caption{Mean squared error on the different normalized fields for an MLP and standard GDL baselines on the test set in the Reynolds extrapolation regime. Only the reduced pressure is given on the surface as the other quantities are null via the boundary conditions. Those quantities are directly regressed by the models.}
  \label{tab:MSE_reynolds}
  \centering
  \begin{tabular}{cccccc}
    \toprule
    Model & \multicolumn{4}{c}{Volume} & Surface  \\
    & $\Bar{u}_x$ ($\times 10^{-2}$) & $\Bar{u}_y$ ($\times 10^{-2}$) & $\Bar{p}$ ($\times 10^{-2}$) & $\nu_t$ ($\times 10^{-1}$) & $\Bar{p}$ ($\times 10^{-1}$) \\
    \midrule
    MLP & 9.51$\pm$1.27 & 4.92$\pm$0.80 & 4.30$\pm$0.19 & 1.31$\pm$0.34 & 20.9$\pm$35.5 \\
    GraphSAGE & \textbf{7.56}$\pm$\textbf{1.05} & \textbf{3.50}$\pm$\textbf{0.61} & \textbf{3.83}$\pm$\textbf{0.25} & 1.69$\pm$0.38 & \textbf{1.80}$\pm$\textbf{0.34} \\
    PointNet & 9.42$\pm$1.08 & 7.13$\pm$0.80 & 4.01$\pm$0.74 & \textbf{1.27}$\pm$\textbf{0.44} & 2.01$\pm$0.76 \\
    Graph U-Net & 8.38$\pm$1.82 & 5.25$\pm$1.36 & 4.48$\pm$0.40 & 1.28$\pm$0.31 & 2.06$\pm$0.44 \\
    \bottomrule
  \end{tabular}
\end{table}

\begin{table}
  \caption{Relative errors (Spearman's rank correlation) for the predicted drag coefficient $C_D$ ($\rho_D$) and lift coefficient $C_L$ ($\rho_L$) in the Reynolds extrapolation regime. We want the Spearman's correlation to be close to one. Those quantities are computed as a post processing from the unnormalized regressed fields.}
  \label{tab:spear_reynolds}
  \centering
  \begin{tabular}{ccccccc}
    \toprule
    Model & \multicolumn{2}{c}{Relative error} & \multicolumn{2}{c}{Spearman's correlation}  \\
     & $C_D$ & $C_L$ & $\rho_D$ & $\rho_L$ \\
    \midrule
    MLP & 13.4$\pm$7.2 & 3.33$\pm$3.29 & -0.15$\pm$0.15 & 0.642$\pm$0.274 \\
    GraphSAGE & \textbf{8.97}$\pm$\textbf{1.28} & \textbf{0.62}$\pm$\textbf{0.12} & 0.01$\pm$0.06 & \textbf{0.927}$\pm$\textbf{0.027} \\
    PointNet & 11.6$\pm$2.8 & 0.90$\pm$0.33 & 0.01$\pm$0.24 & 0.898$\pm$0.056 \\
    Graph U-Net & 13.3$\pm$1.8 & 0.87$\pm$0.37 & 0.03$\pm$0.12 & 0.904$\pm$0.064 \\
    \bottomrule
  \end{tabular}
\end{table}

\paragraph{Angle of attack extrapolation regime.} In this regime, we test on out of distribution angle of attacks.

\begin{figure}
  \centering
  \includegraphics[width = \linewidth]{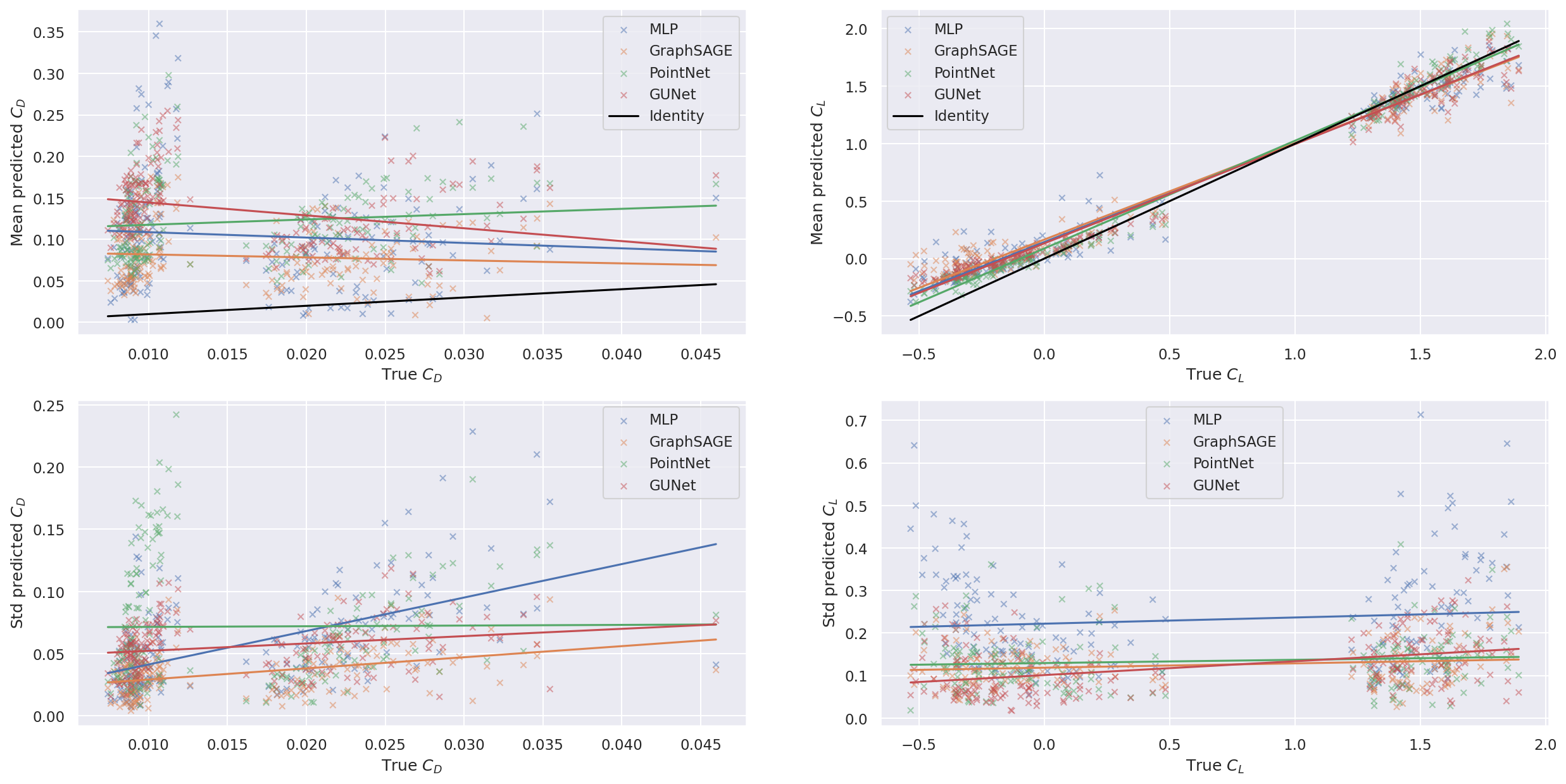}
  \caption{Predicted drag (left) and lift (right) coefficients with respect to the true ones in the angle of attack extrapolation regime. The mean (top) and standard deviation (bottom) of each point on the five copy of the trained models are separated for sake of readability. A linear regression is done for each point cloud in order to highlight linear trends. On the top plots, the Identity graph is given in black for comparison.}
  \label{fig:coefs_aoa}
\end{figure}

\begin{figure}
  \centering
  \includegraphics[width = \linewidth]{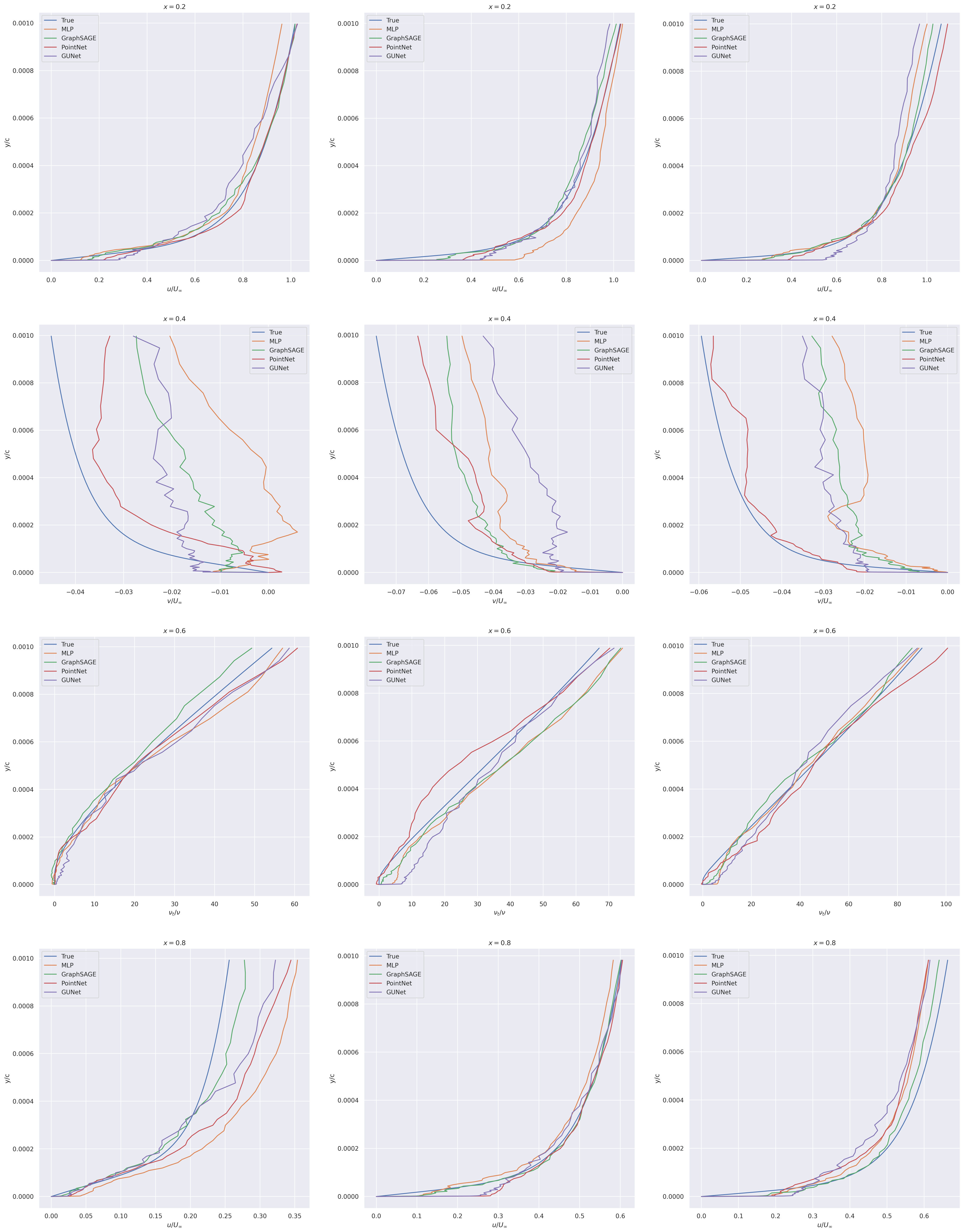}
  \caption{Comparison of the predicted boundary layers profiles on three random test geometries at different abscissas in the angle of attack extrapolation regime with respect to the true ones. Each column of plots represent a different airfoil and each line of plots represent a different abscissas. The $x$ and $y$ component of the velocity are denoted by $u$ and $v$ respectively and the turbulent viscosity is denoted by $\nu_t$. Each quantity is normalized either by $u_\infty$ the inlet velocity magnitude or $\nu$ the fluid viscosity.}
  \label{fig:bl_aoa}
\end{figure}

\begin{figure}
  \centering
  \includegraphics[width = \linewidth]{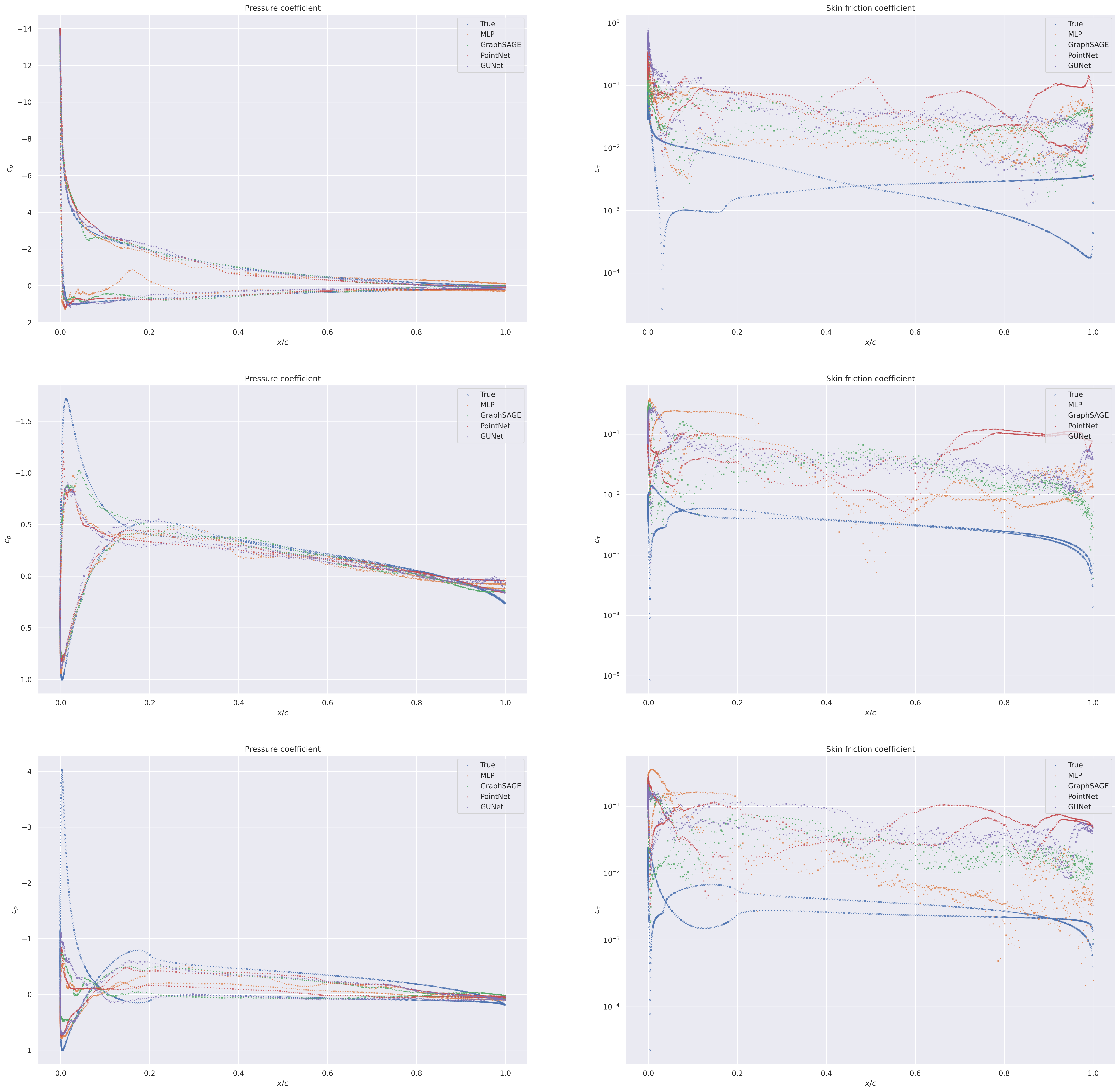}
  \caption{Comparison of the predicted surface coefficients profiles on three random test geometries in the angle of attack extrapolation regime with respect to the true one. (left) Surface coefficient $c_p$ (right) Skin friction coefficient $c_\tau$. Each line of plots represents a different airfoil. Skin friction coefficient plots are given in log scale.}
  \label{fig:surf_aoa}
\end{figure}

In Table~\ref{tab:MSE_aoa}, we give the MSE over the volume and at the surface of airfoils for the different regressed fields. In Table~\ref{tab:spear_aoa} we give the mean relative errors on the force coefficient and the Spearman's rank correlation coefficient. In Figure~\ref{fig:coefs_aoa} we plot the predicted force coefficients with respect to the true coefficients. In Figure~\ref{fig:bl_aoa}, we plot the velocity and turbulent viscosity profiles in the boundary layer for randomly chosen test geometries and in Figure~\ref{fig:surf_aoa} the surface coefficients for the same geometries. 

\begin{table}
  \caption{Mean squared error on the different normalized fields for an MLP and standard GDL baselines on the test set in the angle of attack extrapolation regime. Only the reduced pressure is given on the surface as the other quantities are null via the boundary conditions. Those quantities are directly regressed by the models.}
  \label{tab:MSE_aoa}
  \centering
  \begin{tabular}{cccccc}
    \toprule
    Model & \multicolumn{4}{c}{Volume} & Surface  \\
    & $\Bar{u}_x$ ($\times 10^{-2}$) & $\Bar{u}_y$ ($\times 10^{-1}$) & $\Bar{p}$ ($\times 10^{-1}$) & $\nu_t$ ($\times 10^{-1}$) & $\Bar{p}$ ($\times 10^{-1}$) \\
    \midrule
    MLP & 6.96$\pm$0.54 & 1.06$\pm$0.17 & 1.17$\pm$0.25 & 5.43$\pm$0.24 & 8.76$\pm$1.59 \\
    GraphSAGE & \textbf{4.43}$\pm$\textbf{0.33} & \textbf{0.94}$\pm$\textbf{0.22} & \textbf{1.09}$\pm$\textbf{0.22} & 5.18$\pm$0.37 & 7.64$\pm$0.95 \\
    PointNet & 8.68$\pm$2.34 & 1.58$\pm$0.54 & 1.62$\pm$0.66 & \textbf{4.63}$\pm$\textbf{0.40} & \textbf{5.85}$\pm$\textbf{0.36} \\
    Graph U-Net & 5.69$\pm$0.71 & 1.03$\pm$0.28 & 1.49$\pm$0.75 & 5.35$\pm$0.59 & 6.97$\pm$2.32 \\
    \bottomrule
  \end{tabular}
\end{table}

\begin{table}
  \caption{Relative errors (Spearman's rank correlation) for the predicted drag coefficient $C_D$ ($\rho_D$) and lift coefficient $C_L$ ($\rho_L$) in the angle of attack extrapolation regime. We want the Spearman's correlation to be close to one. Those quantities are computed as a post processing from the unnormalized regressed fields.}
  \label{tab:spear_aoa}
  \centering
  \begin{tabular}{ccccccc}
    \toprule
    Model & \multicolumn{2}{c}{Relative error} & \multicolumn{2}{c}{Spearman's correlation}  \\
     & $C_D$ & $C_L$ & $\rho_D$ & $\rho_L$ \\
    \midrule
    MLP & 8.00$\pm$0.85 & 1.06$\pm$0.29 & 0.04$\pm$0.17 & 0.861$\pm$0.049 \\
    GraphSAGE & \textbf{5.59}$\pm$\textbf{1.09} & 0.82$\pm$0.30 & 0.05$\pm$0.17 & 0.908$\pm$0.019 \\
    PointNet & 8.99$\pm$3.44 & 0.72$\pm$0.15 & 0.12$\pm$0.30 & \textbf{0.936}$\pm$\textbf{0.022} \\
    Graph U-Net & 10.2$\pm$3.4 & \textbf{0.69}$\pm$\textbf{0.14} & -0.20$\pm$0.13 & 0.934$\pm$0.022 \\
    \bottomrule
  \end{tabular}
\end{table}

\paragraph{Summary.} All the three latter regimes are obviously more difficult than the full data one. The MSE scores on the scarce data regime, the Reynolds and angle of attack extrapolation regimes are order of magnitude greater than the ones on the full data regime. This can lead to higher relative errors on force coefficients and lower score on Spearman's correlation but is not directly linked to (as in the scarce data regime). In any case, the wall shear stress is never well predict as the velocity profiles in the boundary layer are not accurate close to the geometries and the pressure coefficient suffers from out of distribution conditions depending on the models. In total, there is a lot of progress possible with those four settings and we hope that more recent architectures and technologies will better perform on the proposed metrics.

In Table~\ref{tab:MSE_comparison} we give the MSE scores of all of the models on all of the tasks and in Table~\ref{tab:spear_comparison} their scores on the force coefficients in a more readable way.

\begin{table}
  \caption{Comparison of the mean squared error on the normalized fields for an MLP and standard GDL baselines on the different task for the associated test set. Only the reduced pressure is given on the surface as the other quantities are null via the boundary conditions. Those quantities are directly regressed by the models. The field denoted by $\Bar{p}_s$ is the mean field reduced pressure at the surface of airfoils.}
  \label{tab:MSE_comparison}
  \centering
  \begin{tabular}{cccccc}
    \toprule
    Field & Model & \multicolumn{4}{c}{Task}  \\
     & & Full & Scarce & Reynolds & AoA \\
    \midrule
    \multirow{4}{*}{$\Bar{u}_x$ ($\times 10^{-2}$)} & MLP & 0.949 $\pm$ 0.058 & 1.647 $\pm$ 0.032 & 9.505 $\pm$ 1.275 & 6.965 $\pm$ 0.545 \\
     & GraphSAGE & \textbf{0.832 $\pm$ 0.015} & \textbf{1.457 $\pm$ 0.125} & \textbf{7.558 $\pm$ 1.046} & \textbf{4.435 $\pm$ 0.334} \\
     & PointNet & 3.500 $\pm$ 1.044 & 3.111 $\pm$ 0.303 & 9.422 $\pm$ 1.082 & 8.680 $\pm$ 2.337 \\
     & GUNet & 1.517 $\pm$ 0.343 & 1.749 $\pm$ 0.190 & 8.383 $\pm$ 1.815 & 5.689 $\pm$ 0.708 \\
    \midrule
    \multirow{4}{*}{$\Bar{u}_y$ ($\times 10^{-2}$)} & MLP & \textbf{0.978 $\pm$ 0.172} & \textbf{1.451 $\pm$ 0.071} & 4.924 $\pm$ 0.800 & 10.630 $\pm$ 1.651 \\
     & GraphSAGE & 0.994 $\pm$ 0.052 & 1.454 $\pm$ 0.123 & \textbf{3.498 $\pm$ 0.613} & \textbf{9.400 $\pm$ 2.167} \\
     & PointNet & 3.645 $\pm$ 1.261 & 2.776 $\pm$ 0.395 & 7.129 $\pm$ 0.801 & 15.796 $\pm$ 5.392 \\
     & GUNet & 2.028 $\pm$ 0.391 & 1.825 $\pm$ 0.182 & 5.250 $\pm$ 1.362 & 10.342 $\pm$ 2.788 \\
    \midrule
    \multirow{4}{*}{$\Bar{p}$ ($\times 10^{-2}$)} & MLP & 0.737 $\pm$ 0.131 & 3.904 $\pm$ 0.570 & 4.300 $\pm$ 0.188 & 11.711 $\pm$ 2.518 \\
     & GraphSAGE & 0.661 $\pm$ 0.050 & 4.696 $\pm$ 0.804 & \textbf{3.826 $\pm$ 0.248} & \textbf{10.908 $\pm$ 2.164} \\
     & PointNet & 1.151 $\pm$ 0.230 & \textbf{3.294 $\pm$ 1.052} & 4.011 $\pm$ 0.744 & 16.237 $\pm$ 6.601 \\
     & GUNet & \textbf{0.657 $\pm$ 0.080} & 3.388 $\pm$ 0.844 & 4.483 $\pm$ 0.401 & 14.887 $\pm$ 7.502 \\
    \midrule
    \multirow{4}{*}{$\nu_t$ ($\times 10^{-1}$)} & MLP & 0.190 $\pm$ 0.010 & 0.501 $\pm$ 0.076 & 1.312 $\pm$ 0.344 & 5.433 $\pm$ 0.237 \\
     & GraphSAGE & 0.160 $\pm$ 0.021 & 0.611 $\pm$ 0.079 & 1.694 $\pm$ 0.383 & 5.178 $\pm$ 0.365 \\
     & PointNet & 0.292 $\pm$ 0.048 & 0.558 $\pm$ 0.236 & \textbf{1.273 $\pm$ 0.443} & \textbf{4.632 $\pm$ 0.398} \\
     & GUNet & \textbf{0.146 $\pm$ 0.014} & \textbf{0.433 $\pm$ 0.100} & 1.283 $\pm$ 0.310 & 5.348 $\pm$ 0.589 \\
    \midrule
    \multirow{4}{*}{$\Bar{p}_{s}$ ($\times 10^{-1}$)} & MLP & 1.130 $\pm$ 0.141 & 2.192 $\pm$ 0.529 & 20.898 $\pm$ 35.537 & 8.762 $\pm$ 1.589 \\
     & GraphSAGE & 0.662 $\pm$ 0.103 & 1.945 $\pm$ 0.336 & \textbf{1.797 $\pm$ 0.338} & 7.638 $\pm$ 0.945 \\
     & PointNet & 0.925 $\pm$ 0.259 & 1.827 $\pm$ 0.413 & 2.013 $\pm$ 0.758 & \textbf{5.846 $\pm$ 0.361} \\
     & GUNet & \textbf{0.386 $\pm$ 0.071} & \textbf{1.473 $\pm$ 0.347} & 2.059 $\pm$ 0.442 & 6.967 $\pm$ 2.317 \\
    \bottomrule
  \end{tabular}
\end{table}

\begin{table}
  \caption{Comparison of the relative errors (Spearman's rank correlation) for the predicted drag coefficient $C_D$ ($\rho_D$) and lift coefficient $C_L$ ($\rho_L$) on the four different task for the associated test set. We want the Spearman's correlation to be close to one. Those quantities are computed as a post processing from the unnormalized regressed fields.}
  \label{tab:spear_comparison}
  \centering
  \begin{tabular}{cccccc}
    \toprule
    Field & Model & \multicolumn{4}{c}{Task}  \\
     & & Full & Scarce & Reynolds & AoA \\
    \midrule
    \multirow{4}{*}{$C_D$} & MLP & 4.289 $\pm$ 0.679 & \textbf{2.950 $\pm$ 0.144} & 13.397 $\pm$ 7.154 & 8.003 $\pm$ 0.848 \\
     & GraphSAGE & \textbf{4.050 $\pm$ 0.704} & 3.504 $\pm$ 0.998 & \textbf{8.971 $\pm$ 1.278} & \textbf{5.589 $\pm$ 1.090} \\
     & PointNet & 14.637 $\pm$ 3.668 & 8.350 $\pm$ 1.387 & 11.558 $\pm$ 2.783 & 8.991 $\pm$ 3.436 \\
     & GUNet & 10.385 $\pm$ 1.895 & 6.871 $\pm$ 1.801 & 13.268 $\pm$ 1.818 & 10.238 $\pm$ 3.394 \\
    \midrule
    \multirow{4}{*}{$C_L$} & MLP & 0.769 $\pm$ 0.108 & 0.662 $\pm$ 0.161 & 3.330 $\pm$ 3.287 & 1.061 $\pm$ 0.288 \\
     & GraphSAGE & \textbf{0.517 $\pm$ 0.162} &  \textbf{0.385 $\pm$ 0.097} & \textbf{0.616 $\pm$ 0.124} & 0.818 $\pm$ 0.300 \\
     & PointNet & 0.742 $\pm$ 0.186 & 0.587 $\pm$ 0.135 & 0.897 $\pm$ 0.326 & 0.716 $\pm$ 0.145 \\
     & GUNet & 0.489 $\pm$ 0.105 & 0.418 $\pm$ 0.129 & 0.868 $\pm$ 0.369 & \textbf{0.693 $\pm$ 0.136} \\
    \midrule
    \multirow{4}{*}{$\rho_D$} & MLP & -0.117 $\pm$ 0.256 & -0.242 $\pm$ 0.078 & -0.146 $\pm$ 0.153 & 0.038 $\pm$ 0.174 \\
     & GraphSAGE & -0.303 $\pm$ 0.124 & -0.139 $\pm$ 0.175 & 0.013 $\pm$ 0.064 & 0.055 $\pm$ 0.171 \\
     & PointNet & -0.022 $\pm$ 0.097 & -0.050 $\pm$ 0.272 & 0.006 $\pm$ 0.241 & 0.122 $\pm$ 0.300 \\
     & GUNet & -0.138 $\pm$ 0.258 & -0.095 $\pm$ 0.232 & 0.028 $\pm$ 0.116 & -0.195 $\pm$ 0.134 \\
    \midrule
    \multirow{4}{*}{$\rho_L$} & MLP & 0.913 $\pm$ 0.018 & 0.923 $\pm$ 0.026 & 0.642 $\pm$ 0.274 & 0.861 $\pm$ 0.049 \\
     & GraphSAGE & 0.965 $\pm$ 0.011 & \textbf{0.981 $\pm$ 0.006} & \textbf{0.927 $\pm$ 0.027} & 0.908 $\pm$ 0.019 \\
     & PointNet & 0.938 $\pm$ 0.023 & 0.949 $\pm$ 0.019 & 0.898 $\pm$ 0.056 & \textbf{0.936 $\pm$ 0.022} \\
     & GUNet &\textbf{ 0.967 $\pm$ 0.019} & 0.976 $\pm$ 0.009 & 0.904 $\pm$ 0.064 & 0.934 $\pm$ 0.022 \\
    \bottomrule
  \end{tabular}
\end{table}

\clearpage
\section{Results for loss \ref{eq:global_loss}}\label{ap:corrected_results}

In this section, we gives the results of the proposed models trained with the loss \ref{eq:global_loss} in the same way as the previous section.

\begin{figure}
    \centering
    \includegraphics[width = \linewidth]{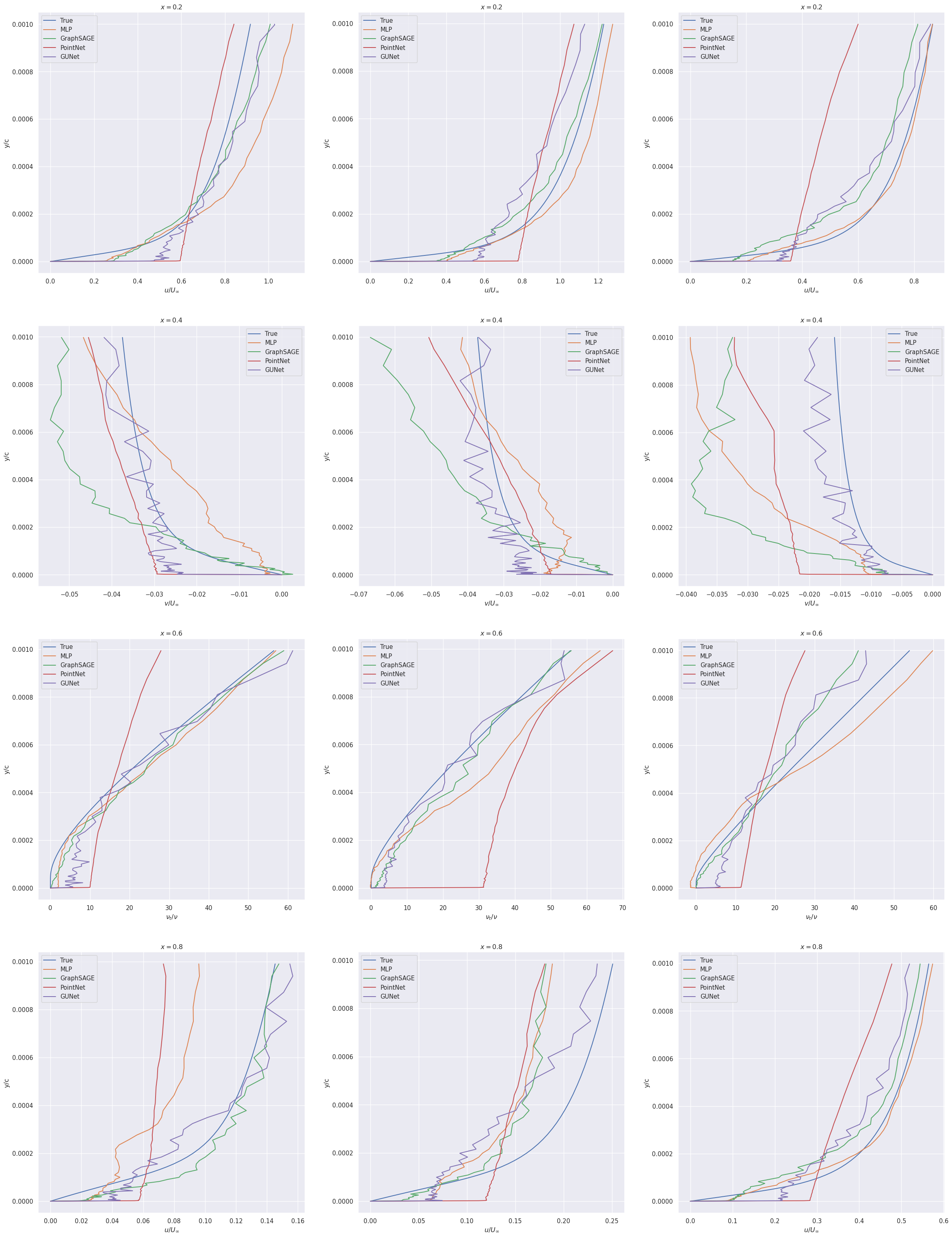}
    \caption{Comparison of the predicted boundary layers profiles on three random test geometries at different abscissas in the full data regime with respect to the true ones. Each column of plots represent a different airfoil and each line of plots represent a different abscissas. The $x$ and $y$ component of the velocity are denoted by $u$ and $v$ respectively and the turbulent viscosity is denoted by $\nu_t$. Each quantity is normalized either by $u_\infty$ the inlet velocity magnitude or $\nu$ the fluid viscosity.}
    \label{fig:bl_full_corr}
\end{figure}

\begin{figure}
    \centering
    \includegraphics[width = \linewidth]{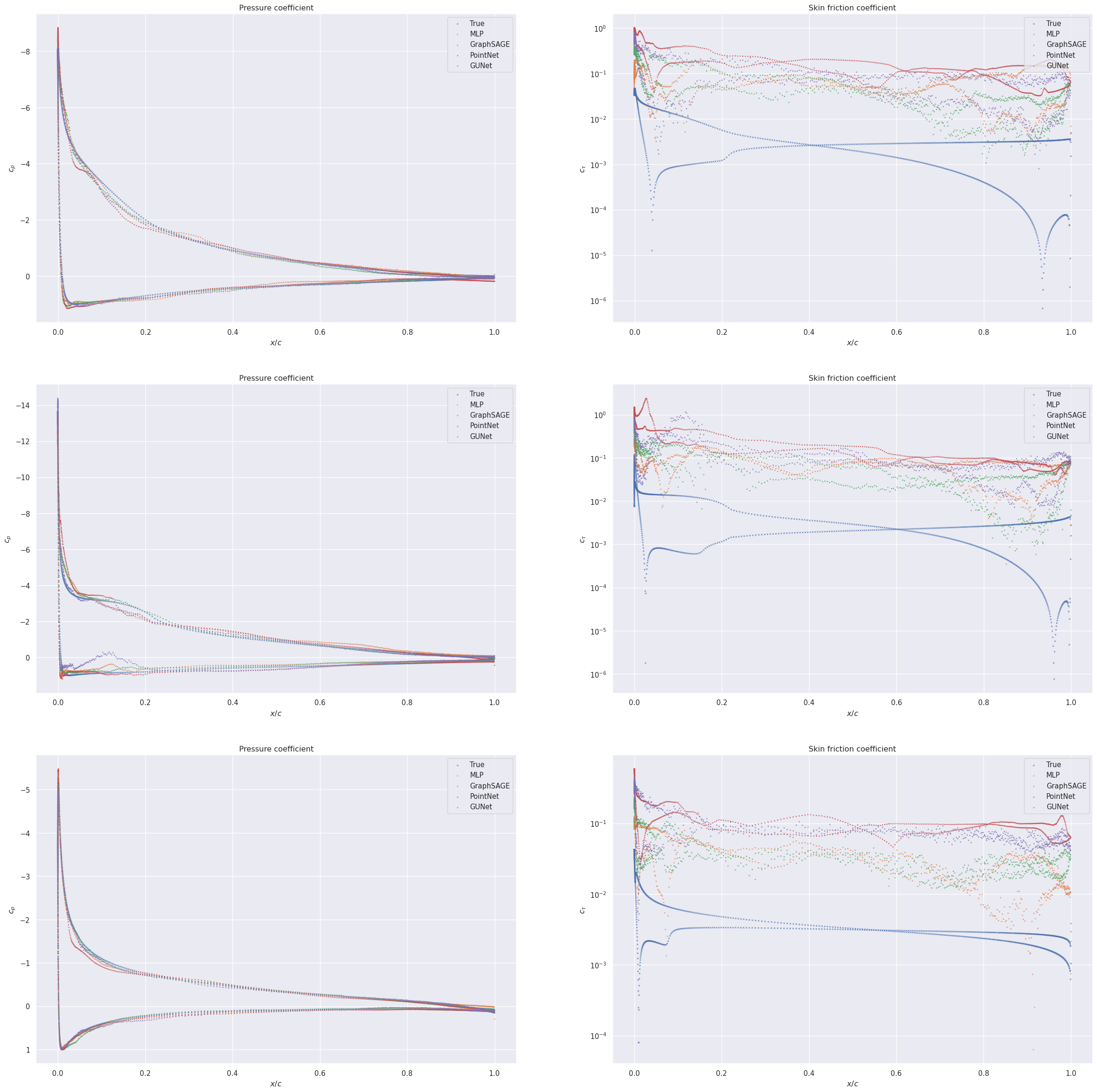}
    \caption{Comparison of the predicted surface coefficients profiles on three random test geometries in the full data regime with respect to the true one. (left) Surface coefficient $c_p$ (right) Skin friction coefficient $c_\tau$. Each line of plots represents a different airfoil. Skin friction coefficient plots are given in log scale.}
    \label{fig:surf_full_corr}
\end{figure}

\paragraph{Full data regime.} In Table~\ref{tab:MSE_full_corr}, we give the MSE over the volume and at the surface of airfoils for the different regressed fields. In Table~\ref{tab:spear_full_corr} we give the mean relative errors on the force coefficient and the Spearman's rank correlation coefficient. In Figure~\ref{fig:surf_full_corr} we show the pressure and skin friction coefficients distributions over the surface of three randomly chosen airfoils in the test set. In Figure~\ref{fig:bl_full_corr}, we present the $x$ and $y$ components of the velocity distribution in the boundary layer at the upper surface of the same three airfoils. Those velocity profiles are given at four abscissas: $x = \SI{0.2}{\meter}$, $x = \SI{0.4}{\meter}$, $x = \SI{0.6}{\meter}$ and $x = \SI{0.8}{\meter}$.

\begin{table}
  \caption{Mean squared error on the different normalized fields for an MLP and standard GDL baselines on the test set in the full data regime. Only the reduced pressure is given on the surface as the other quantities are null via the boundary conditions. Those quantities are directly regressed by the models.}
  \label{tab:MSE_full_corr}
  \centering
  \begin{tabular}{cccccc}
    \toprule
    Model & \multicolumn{4}{c}{Volume} & Surface  \\
    & $\Bar{u}_x$ ($\times 10^{-2}$) & $\Bar{u}_y$ ($\times 10^{-2}$) & $\Bar{p}$ ($\times 10^{-2}$) & $\nu_t$ ($\times 10^{-2}$) & $\Bar{p}$ ($\times 10^{-2}$) \\
    \midrule
    MLP & 1.63$\pm$0.19 & \textbf{1.09$\pm$0.38} & 0.81$\pm$0.06 & 2.59$\pm$0.19 & 2.00$\pm$0.41 \\
    GraphSAGE & \textbf{1.58$\pm$0.16} & 1.41$\pm$0.24 & 0.87$\pm$0.12 & 2.11$\pm$0.10 & 1.84$\pm$0.37 \\
    PointNet & 6.63$\pm$1.10 & 6.01$\pm$1.13 & 2.53$\pm$0.50 & 5.26$\pm$1.89 & 9.96$\pm$0.52 \\
    Graph U-Net & 2.81$\pm$0.34 & 2.95$\pm$0.40 & \textbf{0.76$\pm$0.06} & \textbf{1.78$\pm$0.09} & \textbf{1.44$\pm$0.02} \\
    \bottomrule
  \end{tabular}
\end{table}

\begin{table}
  \caption{Relative errors (Spearman's rank correlation) for the predicted drag coefficient $C_D$ ($\rho_D$) and lift coefficient $C_L$ ($\rho_L$) in the full data regime. We want the Spearman's correlation to be close to one. Those quantities are computed as a post processing from the unnormalized regressed fields.}
  \label{tab:spear_full_corr}
  \centering
  \begin{tabular}{ccccccc}
    \toprule
    Model & \multicolumn{2}{c}{Relative error} & \multicolumn{2}{c}{Spearman's correlation}  \\
     & $C_D$ & $C_L$ & $\rho_D$ & $\rho_L$ \\
    \midrule
    MLP & \textbf{6.18$\pm$0.90} & 0.21$\pm$0.03 & \textbf{0.25$\pm$0.09} & 0.9932$\pm$0.0017 \\
    GraphSAGE & 7.37$\pm$1.21 & \textbf{0.15$\pm$0.03} & 0.19$\pm$0.07 & \textbf{0.9964$\pm$0.0007} \\
    PointNet & 17.4$\pm$1.4 & 0.20$\pm$0.0.03 & 0.07$\pm$0.06 & 0.9919$\pm$0.0017 \\
    Graph U-Net & 13.3$\pm$0.9 & 0.17$\pm$0.02 & 0.09$\pm$0.05 & 0.9949$\pm$0.0011 \\
    \bottomrule
  \end{tabular}
\end{table}

\begin{figure}
  \centering
  \includegraphics[width = \linewidth]{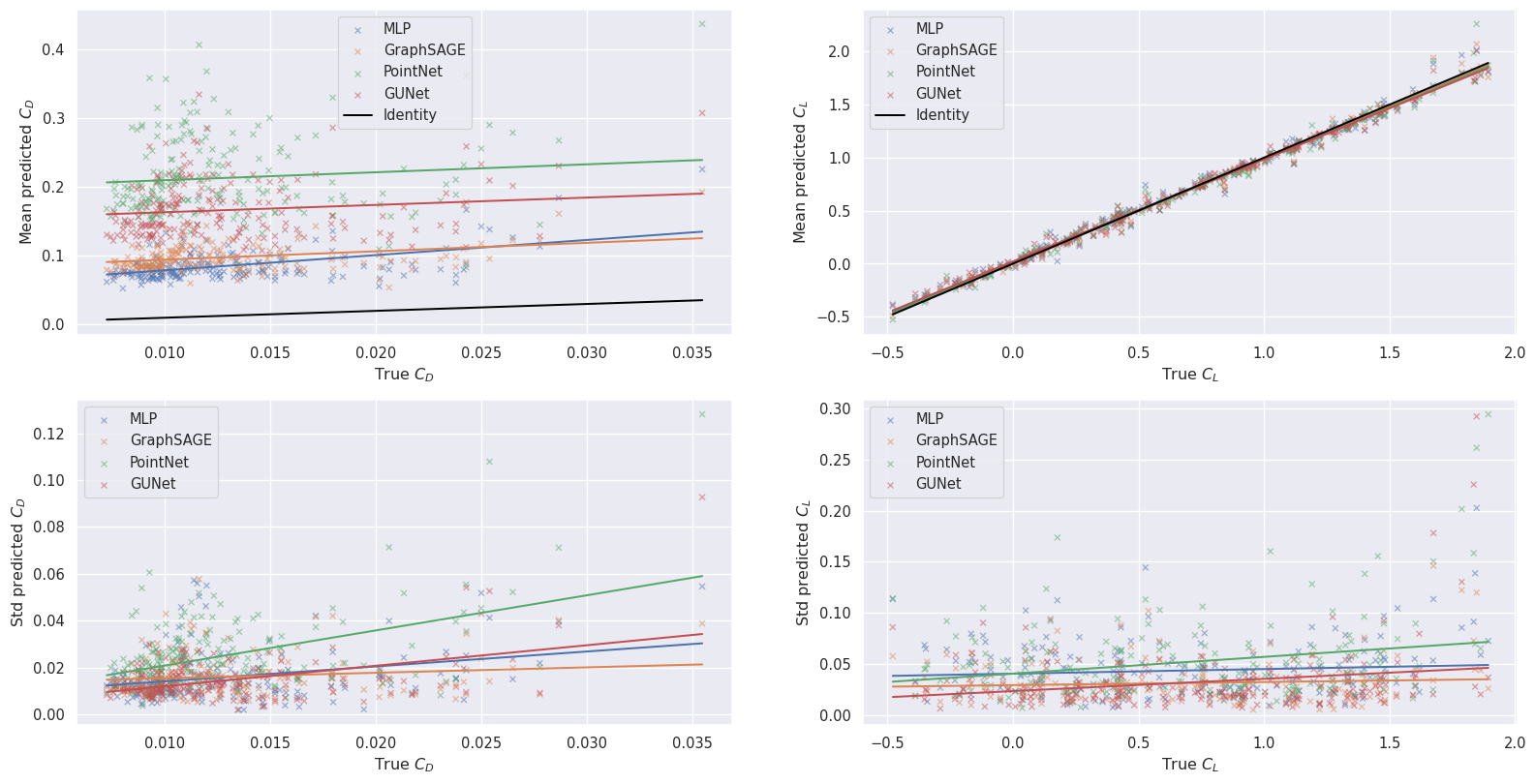}
  \caption{Predicted drag (left) and lift (right) coefficients with respect to the true ones in the full data regime. The mean (top) and standard deviation (bottom) of each point on the five copy of the trained models are separated for sake of readability. A linear regression is done for each point cloud in order to highlight linear trends. On the top plots, the Identity graph is given in black for comparison.}
  \label{fig:coefs_full_corr}
\end{figure}

With this weighted loss, the pressure at the surface is better regressed but the wall shear stress is still largely overestimated, this behaviour is understood via the boundary layer velocity profiles where the first inferred point after the surface is badly approximated. This leads to high errors on the drag coefficient.

\paragraph{Scarce data regime.} In this regime, we test on the same test set as the full data regime but we trained with only two hundred simulations instead of eight hundreds.

\begin{figure}
  \centering
  \includegraphics[width = \linewidth]{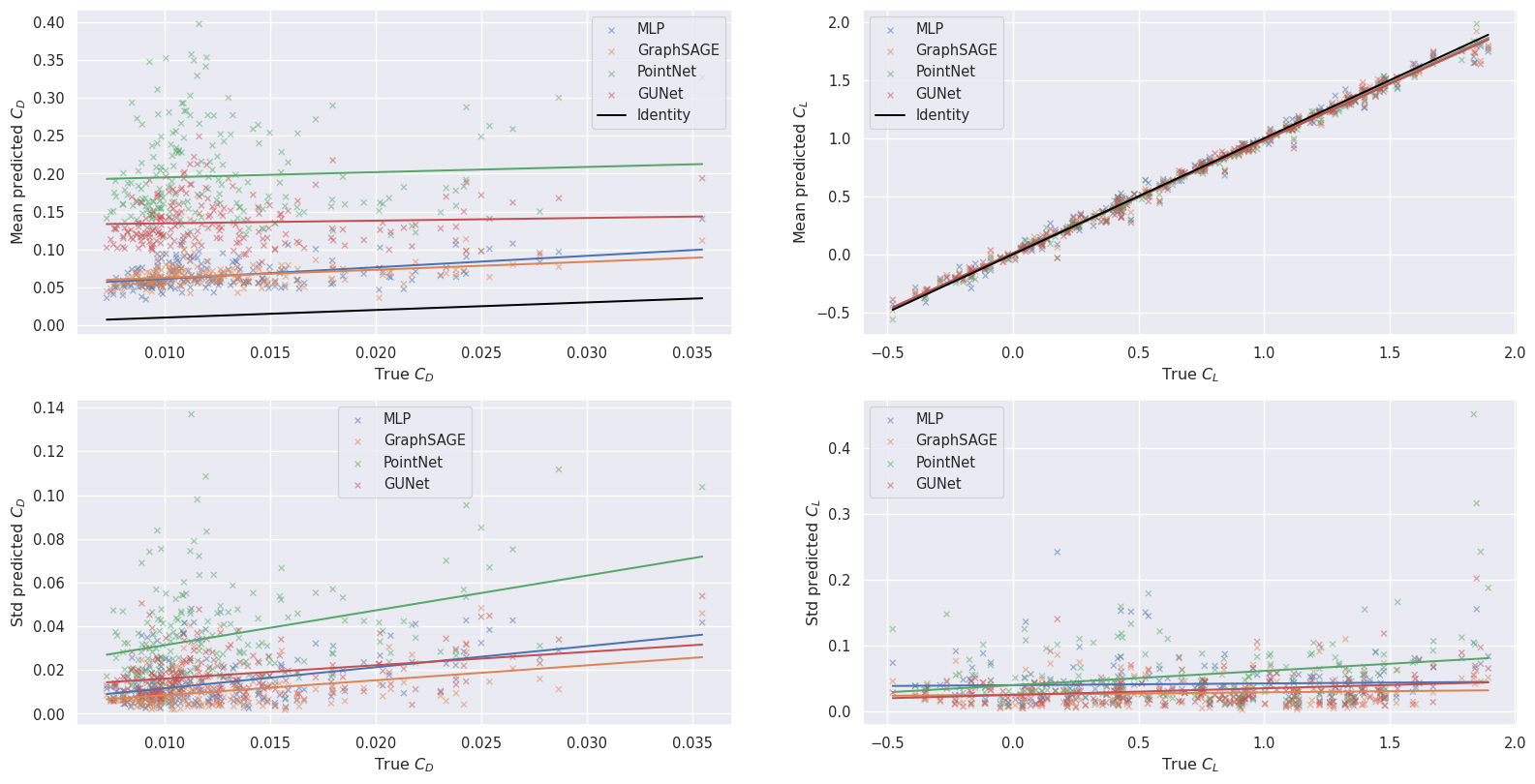}
  \caption{Predicted drag (left) and lift (right) coefficients with respect to the true ones in the scarce data regime. The mean (top) and standard deviation (bottom) of each point on the five copy of the trained models are separated for sake of readability. A linear regression is done for each point cloud in order to highlight linear trends. On the top plots, the Identity graph is given in black for comparison.}
  \label{fig:coefs_scarce_corr}
\end{figure}

\begin{figure}
  \centering
  \includegraphics[width = \linewidth]{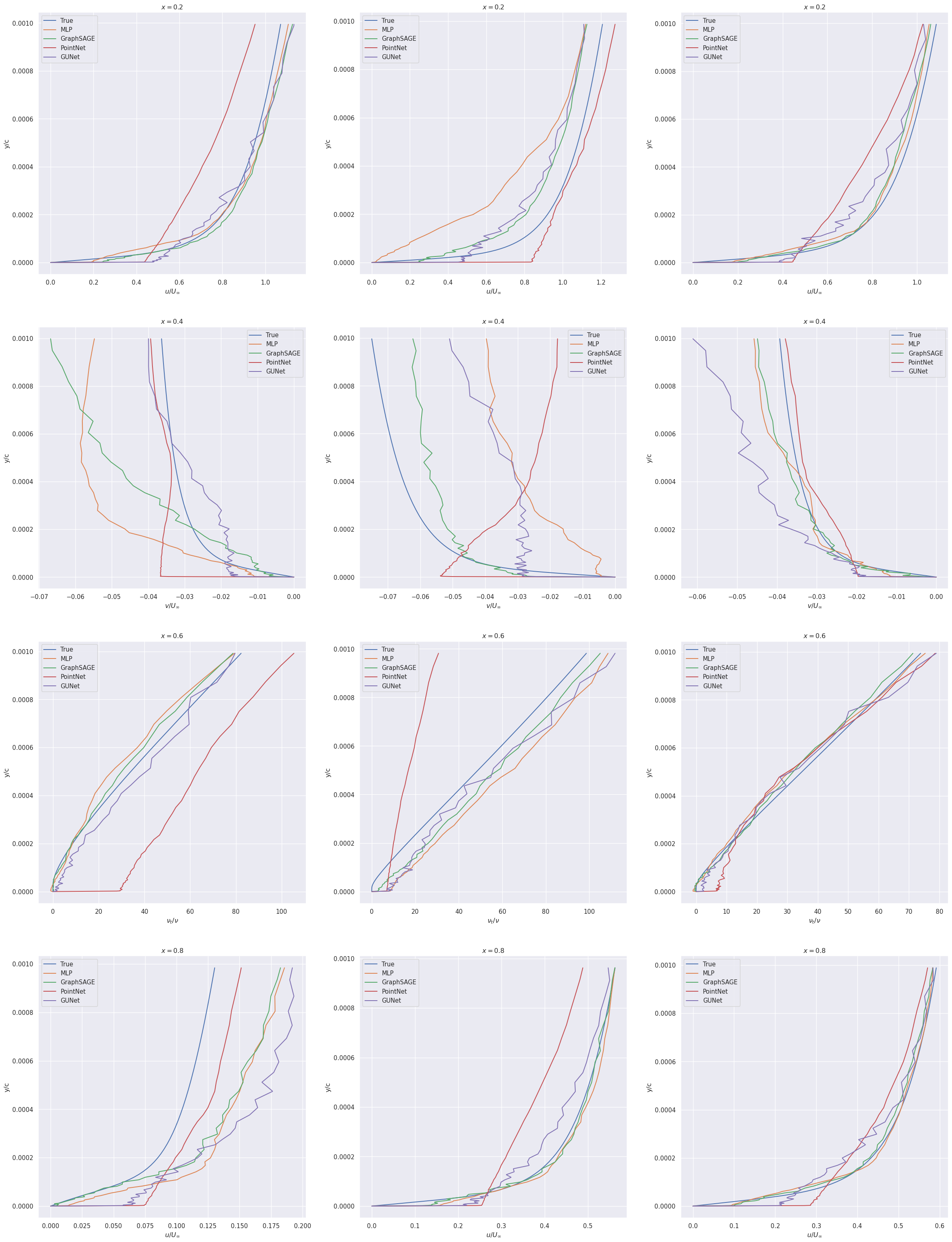}
  \caption{Comparison of the predicted boundary layers profiles on three random test geometries at different abscissas in the scarce data regime with respect to the true ones. Each column of plots represent a different airfoil and each line of plots represent a different abscissas. The $x$ and $y$ component of the velocity are denoted by $u$ and $v$ respectively and the turbulent viscosity is denoted by $\nu_t$. Each quantity is normalized either by $u_\infty$ the inlet velocity magnitude or $\nu$ the fluid viscosity.}
  \label{fig:bl_scarce_corr}
\end{figure}

\begin{figure}
  \centering
  \includegraphics[width = \linewidth]{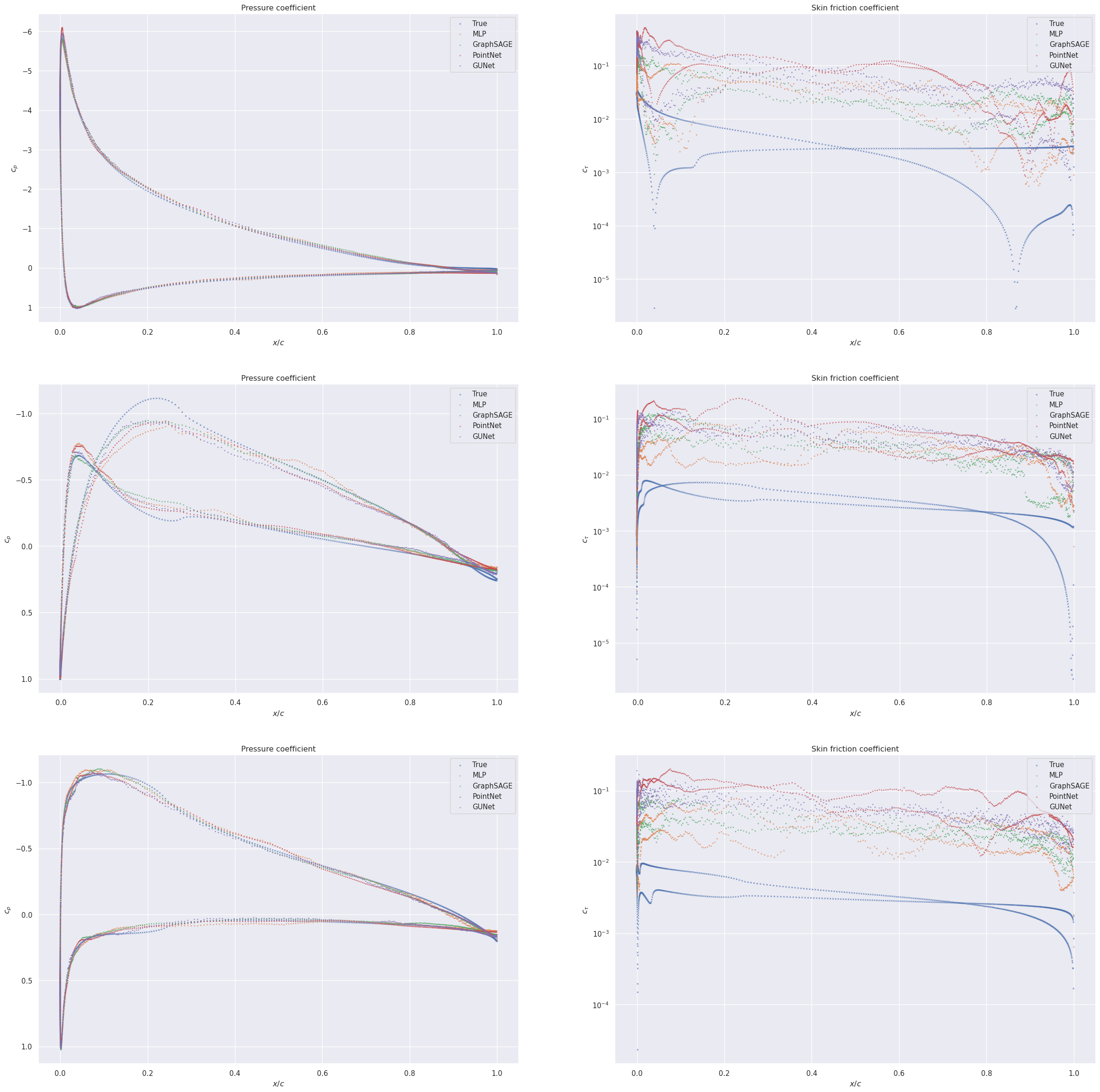}
  \caption{Comparison of the predicted surface coefficients profiles on three random test geometries in the scarce data regime with respect to the true one. (left) Surface coefficient $c_p$ (right) Skin friction coefficient $c_\tau$. Each line of plots represents a different airfoil. Skin friction coefficient plots are given in log scale.}
  \label{fig:surf_scarce_corr}
\end{figure}

In Table~\ref{tab:MSE_scarce_corr}, we give the MSE over the volume and at the surface of airfoils for the different regressed fields. In Table~\ref{tab:spear_scarce_corr} we give the mean relative errors on the force coefficient and the Spearman's rank correlation coefficient. In Figure~\ref{fig:coefs_scarce_corr} we plot the predicted force coefficients with respect to the true coefficients. In Figure~\ref{fig:bl_scarce_corr}, we plot the velocity and turbulent viscosity profiles in the boundary layer for randomly chosen test geometries and in Figure~\ref{fig:surf_scarce_corr} the surface coefficients for the same geometries. 

\begin{table}
  \caption{Mean squared error on the different normalized fields for an MLP and standard GDL baselines on the test set in the scarce data regime. Only the reduced pressure is given on the surface as the other quantities are null via the boundary conditions. Those quantities are directly regressed by the models.}
  \label{tab:MSE_scarce_corr}
  \centering
  \begin{tabular}{cccccc}
    \toprule
    Model & \multicolumn{4}{c}{Volume} & Surface  \\
    & $\Bar{u}_x$ ($\times 10^{-2}$) & $\Bar{u}_y$ ($\times 10^{-2}$) & $\Bar{p}$ ($\times 10^{-2}$) & $\nu_t$ ($\times 10^{-2}$) & $\Bar{p}$ ($\times 10^{-1}$) \\
    \midrule
    MLP & 2.32$\pm$0.15 & \textbf{1.81$\pm$0.17} & 4.25$\pm$0.43 & 6.20$\pm$0.95 & 1.21$\pm$0.10 \\
    GraphSAGE & \textbf{1.87$\pm$0.14} & 1.87$\pm$0.19 & 4.85$\pm$0.25 & \textbf{5.18$\pm$0.78} & 1.40$\pm$0.07 \\
    PointNet & 7.52$\pm$1.55 & 6.29$\pm$0.80 & 7.51$\pm$3.13 & 7.16$\pm$2.69 & 1.86$\pm$0.44 \\
    Graph U-Net & 2.65$\pm$0.19 & 2.67$\pm$0.21 & \textbf{2.88$\pm$0.46} & 5.38$\pm$1.26 & \textbf{0.84$\pm$0.10} \\
    \bottomrule
  \end{tabular}
\end{table}

\begin{table}
  \caption{Relative errors (Spearman's rank correlation) for the predicted drag coefficient $C_D$ ($\rho_D$) and lift coefficient $C_L$ ($\rho_L$) in the scarce data regime. We want the Spearman's correlation to be close to one. Those quantities are computed as a post-processing from the unnormalized regressed fields.}
  \label{tab:spear_scarce_corr}
  \centering
  \begin{tabular}{ccccccc}
    \toprule
    Model & \multicolumn{2}{c}{Relative error} & \multicolumn{2}{c}{Spearman's correlation}  \\
     & $C_D$ & $C_L$ & $\rho_D$ & $\rho_L$ \\
    \midrule
    MLP & \textbf{4.54$\pm$0.26} & 0.20$\pm$0.03 & \textbf{0.25$\pm$0.06} & 0.9927$\pm$0.0019 \\
    GraphSAGE & 4.59$\pm$0.40 & \textbf{0.15$\pm$0.01} & 0.25$\pm$0.07 & \textbf{0.9961$\pm$0.0007} \\
    PointNet & 16.0$\pm$1.9 & 0.20$\pm$0.05 & 0.05$\pm$0.07 & 0.9920$\pm$0.0025 \\
    Graph U-Net & 10.7$\pm$1.2 & 0.15$\pm$0.01 & 0.07$\pm$0.02 & 0.9940$\pm$0.0003 \\
    \bottomrule
  \end{tabular}
\end{table}

\paragraph{Reynolds extrapolation regime.} In this regime, we test on out of distribution Reynolds number.

\begin{figure}
  \centering
  \includegraphics[width = \linewidth]{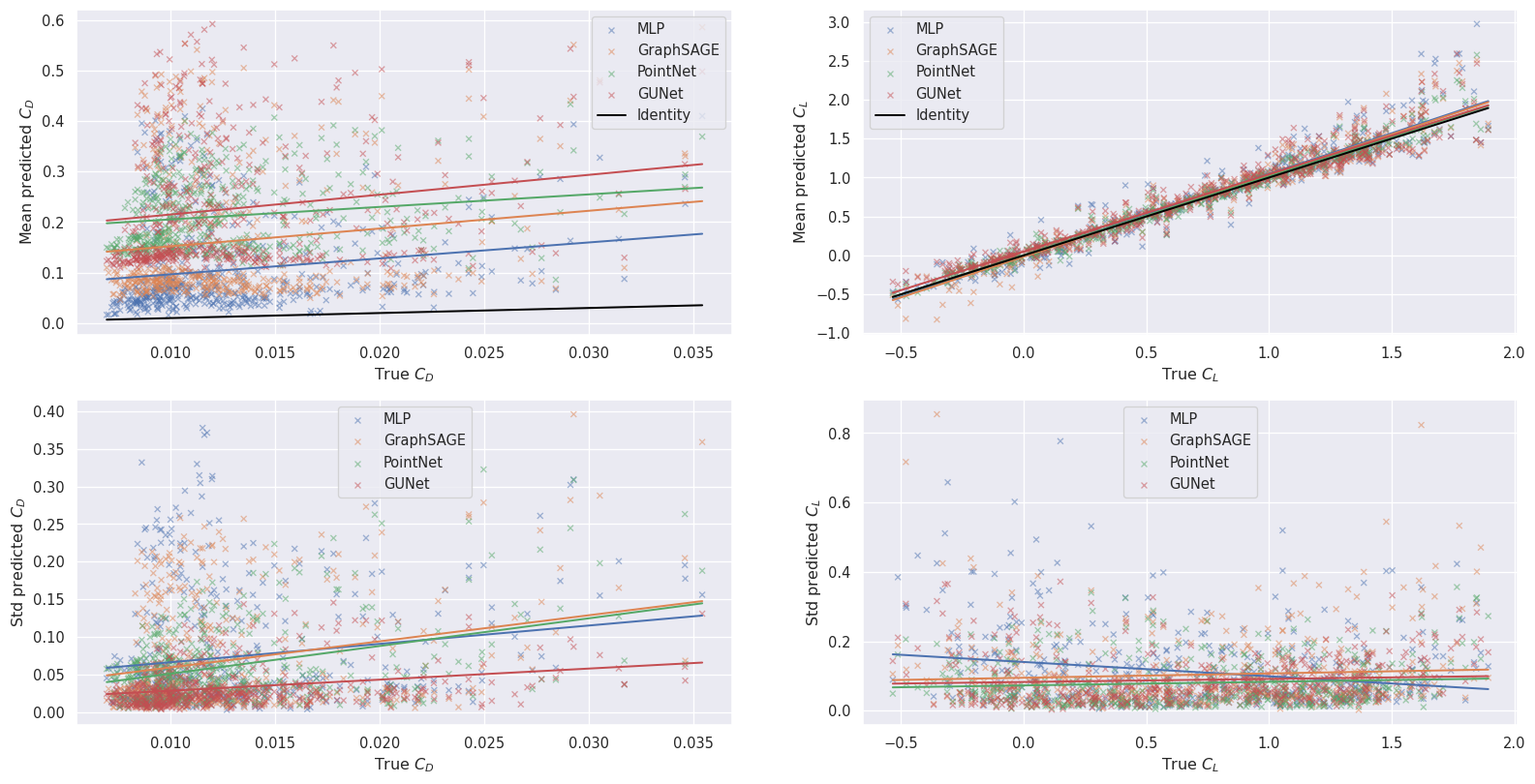}
  \caption{Predicted drag (left) and lift (right) coefficients with respect to the true ones in the Reynolds extrapolation regime. The mean (top) and standard deviation (bottom) of each point on the five copy of the trained models are separated for sake of readability. A linear regression is done for each point cloud in order to highlight linear trends. On the top plots, the Identity graph is given in black for comparison.}
  \label{fig:coefs_reynolds_corr}
\end{figure}

\begin{figure}
  \centering
  \includegraphics[width = \linewidth]{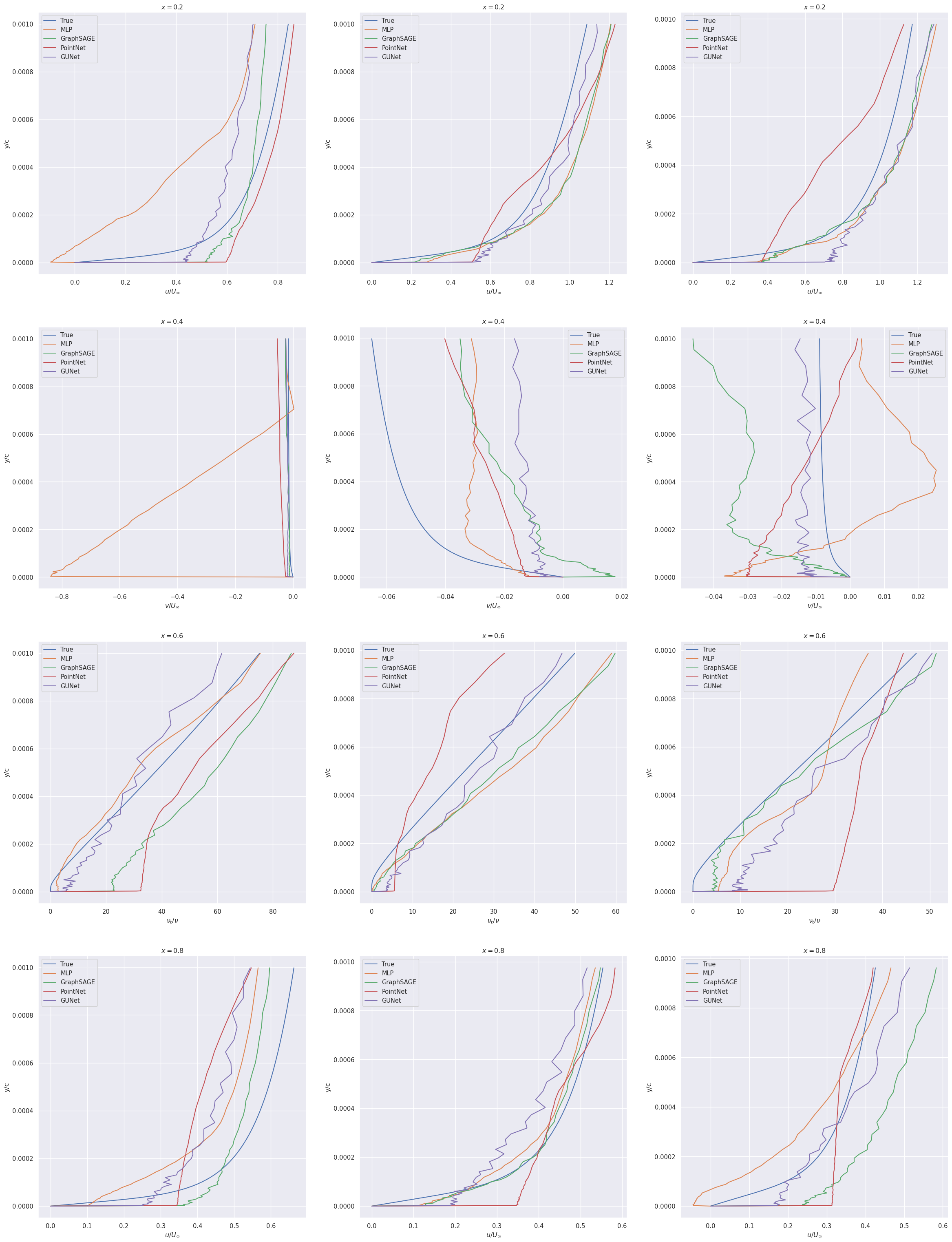}
  \caption{Comparison of the predicted boundary layers profiles on three random test geometries at different abscissas in the Reynolds extrapolation regime with respect to the true ones. Each column of plots represent a different airfoil and each line of plots represent a different abscissas. The $x$ and $y$ component of the velocity are denoted by $u$ and $v$ respectively and the turbulent viscosity is denoted by $\nu_t$. Each quantity is normalized either by $u_\infty$ the inlet velocity magnitude or $\nu$ the fluid viscosity.}
  \label{fig:bl_reynolds_corr}
\end{figure}

\begin{figure}
  \centering
  \includegraphics[width = \linewidth]{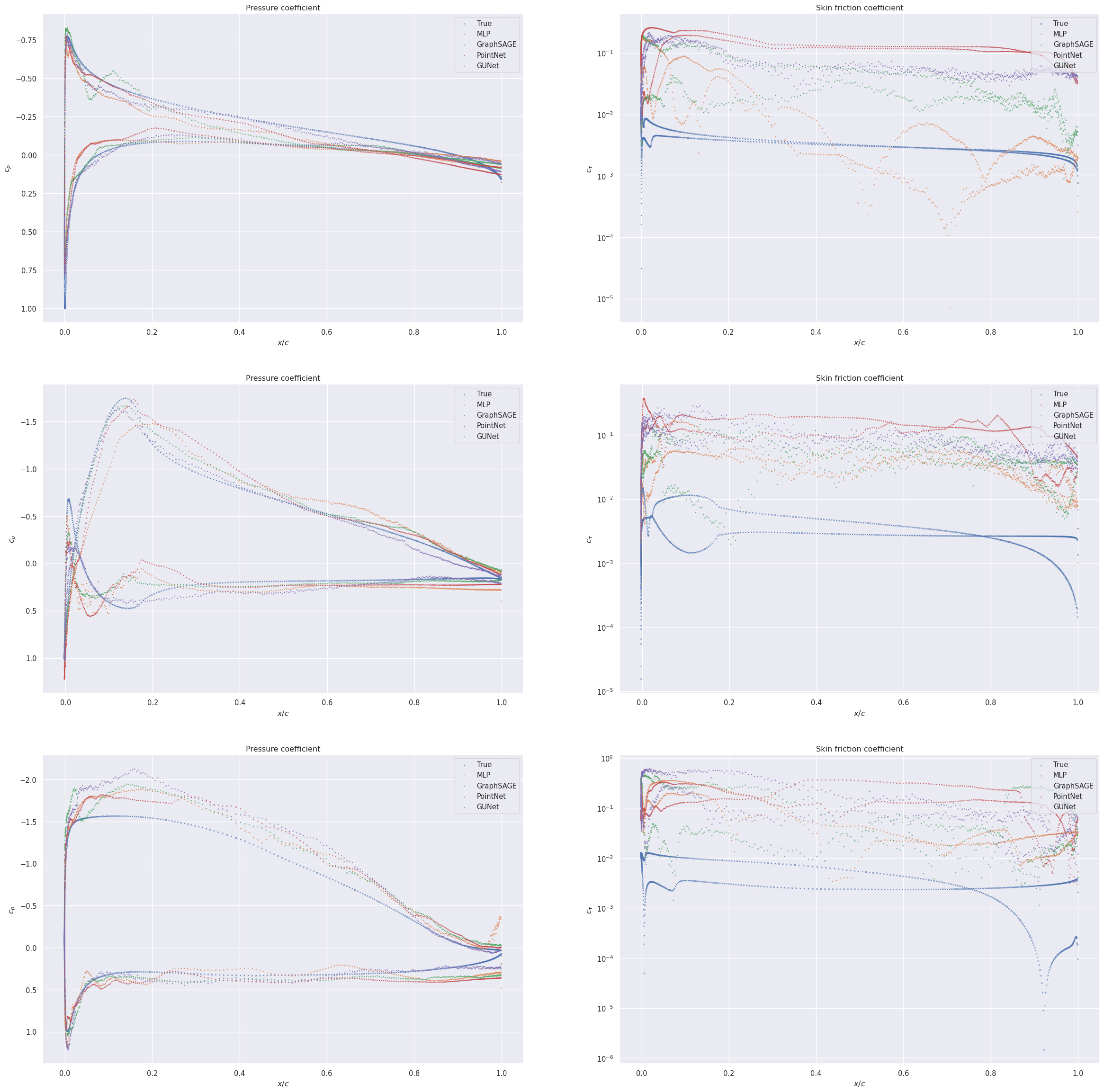}
  \caption{Comparison of the predicted surface coefficients profiles on three random test geometries in the Reynolds extrapolation regime with respect to the true one. (left) Surface coefficient $c_p$ (right) Skin friction coefficient $c_\tau$. Each line of plots represents a different airfoil. Skin friction coefficient plots are given in log scale.}
  \label{fig:surf_reynolds_corr}
\end{figure}

In Table~\ref{tab:MSE_reynolds_corr}, we give the MSE over the volume and at the surface of airfoils for the different regressed fields. In Table~\ref{tab:spear_reynolds_corr} we give the mean relative errors on the force coefficient and the Spearman's rank correlation coefficient. In Figure~\ref{fig:coefs_reynolds_corr} we plot the predicted force coefficients with respect to the true coefficients. In Figure~\ref{fig:bl_reynolds_corr}, we plot the velocity and turbulent viscosity profiles in the boundary layer for randomly chosen test geometries and in Figure~\ref{fig:surf_reynolds_corr} the surface coefficients for the same geometries. 

\begin{table}
  \caption{Mean squared error on the different normalized fields for an MLP and standard GDL baselines on the test set in the Reynolds extrapolation regime. Only the reduced pressure is given on the surface as the other quantities are null via the boundary conditions. Those quantities are directly regressed by the models.}
  \label{tab:MSE_reynolds_corr}
  \centering
  \begin{tabular}{cccccc}
    \toprule
    Model & \multicolumn{4}{c}{Volume} & Surface  \\
    & $\Bar{u}_x$ ($\times 10^{-1}$) & $\Bar{u}_y$ ($\times 10^{-2}$) & $\Bar{p}$ ($\times 10^{-2}$) & $\nu_t$ ($\times 10^{-1}$) & $\Bar{p}$ ($\times 10^{-1}$) \\
    \midrule
    MLP & 1.24$\pm$0.21 & 6.73$\pm$3.13 & 6.69$\pm$1.48 & 1.38$\pm$0.35 & 1\textbf{.15$\pm$0.22} \\
    GraphSAGE & 1.19$\pm$0.26 & 6.67$\pm$1.97 & 7.98$\pm$5.23 & 1.51$\pm$0.48 & 1.25$\pm$0.42 \\
    PointNet & 1.32$\pm$0.28 & 11.3$\pm$6.0 & 8.38$\pm$4.93 & 1.38$\pm$0.98 & 1.40$\pm$0.32 \\
    Graph U-Net & \textbf{1.02$\pm$0.10} & \textbf{5.72$\pm$0.51} & \textbf{5.38$\pm$1.28} & \textbf{1.02$\pm$0.34} & 1.17$\pm$0.25 \\
    \bottomrule
  \end{tabular}
\end{table}

\begin{table}
  \caption{Relative errors (Spearman's rank correlation) for the predicted drag coefficient $C_D$ ($\rho_D$) and lift coefficient $C_L$ ($\rho_L$) in the Reynolds extrapolation regime. We want the Spearman's correlation to be close to one. Those quantities are computed as a post processing from the unnormalized regressed fields.}
  \label{tab:spear_reynolds_corr}
  \centering
  \begin{tabular}{ccccccc}
    \toprule
    Model & \multicolumn{2}{c}{Relative error} & \multicolumn{2}{c}{Spearman's correlation}  \\
     & $C_D$ & $C_L$ & $\rho_D$ & $\rho_L$ \\
    \midrule
    MLP & \textbf{8.29$\pm$3.05} & 0.62$\pm$0.19 & 0.16$\pm$0.14 & 0.958$\pm$0.022 \\
    GraphSAGE & 12.8$\pm$4.0 & 0.43$\pm$0.06 & 0.04$\pm$0.05 & 0.971$\pm$0.018 \\
    PointNet & 17.1$\pm$2.7 & \textbf{0.38$\pm$0.04} & 0.11$\pm$0.15 & \textbf{0.981$\pm$0.006} \\
    Graph U-Net & 18.1$\pm$0.7 & 0.47$\pm$0.12 & \textbf{0.19$\pm$0.13} & 0.964$\pm$0.016 \\
    \bottomrule
  \end{tabular}
\end{table}

\paragraph{Angle of attack extrapolation regime.} In this regime, we test on out of distribution angle of attacks.

\begin{figure}
  \centering
  \includegraphics[width = \linewidth]{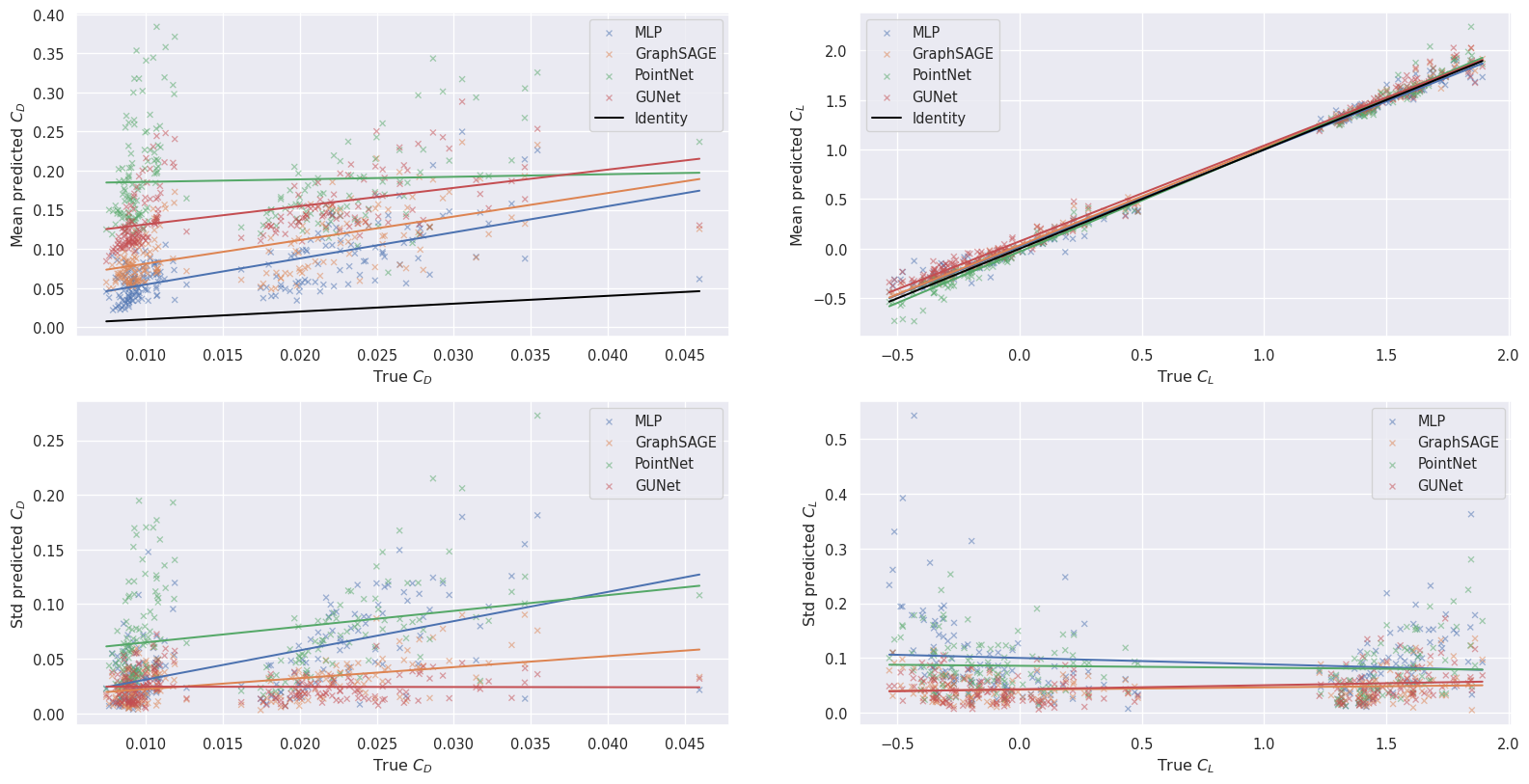}
  \caption{Predicted drag (left) and lift (right) coefficients with respect to the true ones in the angle of attack extrapolation regime. The mean (top) and standard deviation (bottom) of each point on the five copy of the trained models are separated for sake of readability. A linear regression is done for each point cloud in order to highlight linear trends. On the top plots, the Identity graph is given in black for comparison.}
  \label{fig:coefs_aoa_corr}
\end{figure}

\begin{figure}
  \centering
  \includegraphics[width = \linewidth]{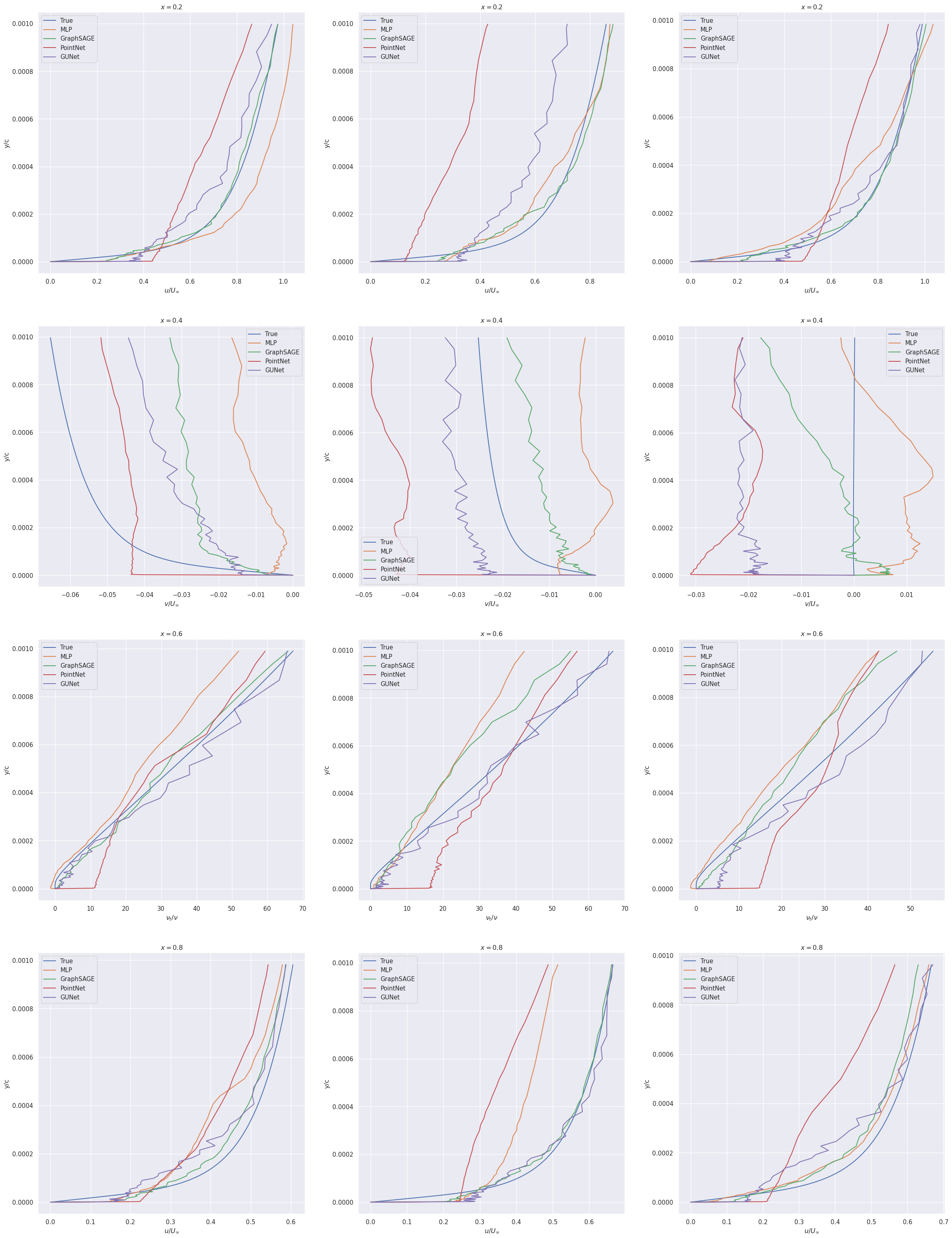}
  \caption{Comparison of the predicted boundary layers profiles on three random test geometries at different abscissas in the angle of attack extrapolation regime with respect to the true ones. Each column of plots represent a different airfoil and each line of plots represent a different abscissas. The $x$ and $y$ component of the velocity are denoted by $u$ and $v$ respectively and the turbulent viscosity is denoted by $\nu_t$. Each quantity is normalized either by $u_\infty$ the inlet velocity magnitude or $\nu$ the fluid viscosity.}
  \label{fig:bl_aoa_corr}
\end{figure}

\begin{figure}
  \centering
  \includegraphics[width = \linewidth]{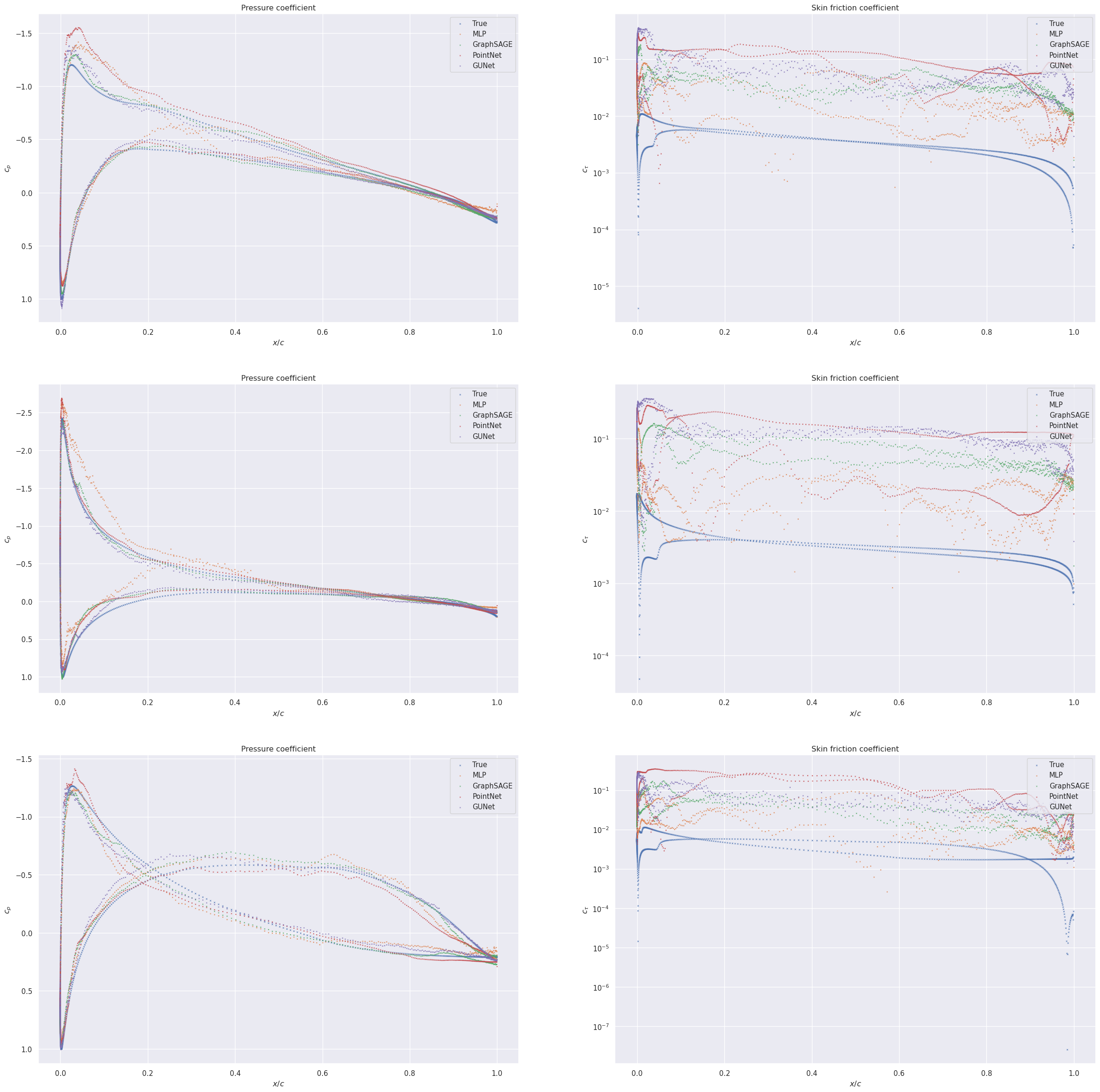}
  \caption{Comparison of the predicted surface coefficients profiles on three random test geometries in the angle of attack extrapolation regime with respect to the true one. (left) Surface coefficient $c_p$ (right) Skin friction coefficient $c_\tau$. Each line of plots represents a different airfoil. Skin friction coefficient plots are given in log scale.}
  \label{fig:surf_aoa_corr}
\end{figure}

In Table~\ref{tab:MSE_aoa_corr}, we give the MSE over the volume and at the surface of airfoils for the different regressed fields. In Table~\ref{tab:spear_aoa_corr} we give the mean relative errors on the force coefficient and the Spearman's rank correlation coefficient. In Figure~\ref{fig:coefs_aoa_corr} we plot the predicted force coefficients with respect to the true coefficients. In Figure~\ref{fig:bl_aoa_corr}, we plot the velocity and turbulent viscosity profiles in the boundary layer for randomly chosen test geometries and in Figure~\ref{fig:surf_aoa_corr} the surface coefficients for the same geometries. 

\begin{table}
  \caption{Mean squared error on the different normalized fields for an MLP and standard GDL baselines on the test set in the angle of attack extrapolation regime. Only the reduced pressure is given on the surface as the other quantities are null via the boundary conditions. Those quantities are directly regressed by the models.}
  \label{tab:MSE_aoa_corr}
  \centering
  \begin{tabular}{cccccc}
    \toprule
    Model & \multicolumn{4}{c}{Volume} & Surface  \\
    & $\Bar{u}_x$ ($\times 10^{-1}$) & $\Bar{u}_y$ ($\times 10^{-1}$) & $\Bar{p}$ ($\times 10^{-1}$) & $\nu_t$ ($\times 10^{-1}$) & $\Bar{p}$ ($\times 10^{-1}$) \\
    \midrule
    MLP & 0.87$\pm$0.23 & 0.98$\pm$0.39 & 1.31$\pm$0.30 & 4.84$\pm$0.41 & 3.31$\pm$1.00 \\
    GraphSAGE & \textbf{0.56$\pm$0.10} & \textbf{0.89$\pm$0.25} & 1.19$\pm$0.51 & \textbf{4.71$\pm$0.45} & \textbf{2.36$\pm$0.97} \\
    PointNet & 1.64$\pm$0.49 & 2.33$\pm$0.25 & 2.02$\pm$0.40 & 5.16$\pm$0.34 & 4.65$\pm$0.90 \\
    Graph U-Net & 0.71$\pm$0.13 & 0.97$\pm$0.24 & \textbf{0.98$\pm$0.18} & 5.02$\pm$0.15 & 2.39$\pm$0.61 \\
    \bottomrule
  \end{tabular}
\end{table}

\begin{table}
  \caption{Relative errors (Spearman's rank correlation) for the predicted drag coefficient $C_D$ ($\rho_D$) and lift coefficient $C_L$ ($\rho_L$) in the angle of attack extrapolation regime. We want the Spearman's correlation to be close to one. Those quantities are computed as a post processing from the unnormalized regressed fields.}
  \label{tab:spear_aoa_corr}
  \centering
  \begin{tabular}{ccccccc}
    \toprule
    Model & \multicolumn{2}{c}{Relative error} & \multicolumn{2}{c}{Spearman's correlation}  \\
     & $C_D$ & $C_L$ & $\rho_D$ & $\rho_L$ \\
    \midrule
    MLP & \textbf{4.35$\pm$1.13} & 0.41$\pm$0.10 & 0.35$\pm$0.22 & 0.957$\pm$0.036 \\
    GraphSAGE & 6.05$\pm$1.30 & \textbf{0.25$\pm$0.05} & 0.53$\pm$0.09 & \textbf{0.989$\pm$0.002} \\
    PointNet & 13.9$\pm$4.3 & 0.44$\pm$0.11 & 0.09$\pm$0.28 & 0.978$\pm$0.003 \\
    Graph U-Net & 9.81$\pm$1.28 & 0.38$\pm$0.05 & \textbf{0.55$\pm$0.06} & 0.981$\pm$0.007 \\
    \bottomrule
  \end{tabular}
\end{table}

Finally, in Table~\ref{tab:MSE_comparison_corr} we give the MSE scores of all of the models on all of the tasks and in Table~\ref{tab:spear_comparison_corr} their scores on the force coefficients in a more readable way.

\begin{table}
  \caption{Comparison of the mean squared error on the normalized fields for an MLP and standard GDL baselines on the different task for the associated test set. Only the reduced pressure is given on the surface as the other quantities are null via the boundary conditions. Those quantities are directly regressed by the models. The field denoted by $\Bar{p}_s$ is the mean field reduced pressure at the surface of airfoils.}
  \label{tab:MSE_comparison_corr}
  \centering
  \begin{tabular}{cccccc}
    \toprule
    Field & Model & \multicolumn{4}{c}{Task}  \\
     & & Full & Scarce & Reynolds & AoA \\
    \midrule
    \multirow{4}{*}{$\Bar{u}_x$ ($\times 10^{-2}$)} & MLP & 1.63 $\pm$ 0.19 & 2.32 $\pm$ 0.15 & 12.4 $\pm$ 2.1 & 8.67 $\pm$ 2.35 \\
     & GraphSAGE & \textbf{1.58 $\pm$ 0.16} & \textbf{1.87 $\pm$ 0.14} & 11.9 $\pm$ 2.6 & \textbf{5.64 $\pm$ 1.00} \\
     & PointNet & 6.63 $\pm$ 1.10 & 7.52 $\pm$ 1.55 & 13.2 $\pm$ 2.8 & 16.4 $\pm$ 4.94 \\
     & GUNet & 2.81 $\pm$ 0.34 & 2.65 $\pm$ 0.19 & \textbf{10.2 $\pm$ 1.0} & 7.13 $\pm$ 1.29 \\
    \midrule
    \multirow{4}{*}{$\Bar{u}_y$ ($\times 10^{-2}$)} & MLP & \textbf{1.09 $\pm$ 0.38} & \textbf{1.81 $\pm$ 0.17} & 6.73 $\pm$ 0.31 & 9.82 $\pm$ 3.86 \\
     & GraphSAGE & 1.41 $\pm$ 0.24 & 1.87 $\pm$ 0.19 & 6.68 $\pm$ 1.97 & 8.88 $\pm$ 2.51 \\
     & PointNet & 6.01 $\pm$ 1.13 & 6.29 $\pm$ 0.80 & 11.3 $\pm$ 6.0 & 23.3 $\pm$ 4.9 \\
     & GUNet & 2.95 $\pm$ 0.40 & 2.67 $\pm$ 0.21 & \textbf{5.72 $\pm$ 0.51} & \textbf{9.68 $\pm$ 2.37} \\
    \midrule
    \multirow{4}{*}{$\Bar{p}$ ($\times 10^{-2}$)} & MLP & 0.81 $\pm$ 0.06 & 4.25 $\pm$ 0.43 & 6.69 $\pm$ 1.48 & 13.1 $\pm$ 3.0 \\
     & GraphSAGE & 0.87 $\pm$ 0.12 & 4.85 $\pm$ 0.25 & 7.98 $\pm$ 5.23 & 11.9 $\pm$ 5.1 \\
     & PointNet & 2.53 $\pm$ 0.50 & 7.51 $\pm$ 3.13 & 8.38 $\pm$ 4.93 & 20.2 $\pm$ 4.0 \\
     & GUNet & \textbf{0.76 $\pm$ 0.06} & \textbf{2.88 $\pm$ 0.46} & \textbf{5.38 $\pm$ 1.28} & \textbf{9.79 $\pm$ 1.79} \\
    \midrule
    \multirow{4}{*}{$\nu_t$ ($\times 10^{-2}$)} & MLP & 2.59 $\pm$ 0.19 & 6.20 $\pm$ 0.95 & 13.8 $\pm$ 3.5 & 48.4 $\pm$ 4.1 \\
     & GraphSAGE & 2.11 $\pm$ 0.10 & \textbf{5.18 $\pm$ 0.78} & 15.1 $\pm$ 4.8 & \textbf{47.1 $\pm$ 4.5} \\
     & PointNet & 5.26 $\pm$ 1.89 & 7.16 $\pm$ 2.69 & 13.8 $\pm$ 9.8 & 51.6 $\pm$ 3.4 \\
     & GUNet & \textbf{1.78 $\pm$ 0.09} & 5.38 $\pm$ 1.26 & \textbf{10.2 $\pm$ 3.4} & 50.2 $\pm$ 1.5 \\
    \midrule
    \multirow{4}{*}{$\Bar{p}_{s}$ ($\times 10^{-2}$)} & MLP & 2.00 $\pm$ 0.41 & 12.1 $\pm$ 1.0 & \textbf{11.5 $\pm$ 2.2} & 33.1 $\pm$ 10.0 \\
     & GraphSAGE & 1.84 $\pm$ 0.37 & 14.0 $\pm$ 0.7 & 12.5 $\pm$ 4.2 & \textbf{23.6 $\pm$ 9.7} \\
     & PointNet & 9.96 $\pm$ 5.24 & 18.6 $\pm$ 4.4 & 14.0 $\pm$ 3.2 & 46.5 $\pm$ 9.0 \\
     & GUNet & \textbf{1.44 $\pm$ 0.19} & \textbf{8.41 $\pm$ 1.04} & 11.7 $\pm$ 2.5 & 23.9 $\pm$ 6.1 \\
    \bottomrule
  \end{tabular}
\end{table}

\begin{table}
  \caption{Comparison of the relative errors (Spearman's rank correlation) for the predicted drag coefficient $C_D$ ($\rho_D$) and lift coefficient $C_L$ ($\rho_L$) on the four different task for the associated test set. We want the Spearman's correlation to be close to one. Those quantities are computed as a post processing from the unnormalized regressed fields.}
  \label{tab:spear_comparison_corr}
  \centering
  \begin{tabular}{cccccc}
    \toprule
    Field & Model & \multicolumn{4}{c}{Task}  \\
     & & Full & Scarce & Reynolds & AoA \\
    \midrule
    \multirow{4}{*}{$C_D$} & MLP & \textbf{6.178 $\pm$ 0.900} & \textbf{4.540 $\pm$ 0.256} & \textbf{8.293 $\pm$ 3.049} & \textbf{4.355 $\pm$ 1.128} \\
     & GraphSAGE & 7.366 $\pm$ 1.212 & 4.587 $\pm$ 0.396 & 12.794 $\pm$ 3.980 & 6.047 $\pm$ 1.304 \\
     & PointNet & 17.392 $\pm$ 1.373 & 16.048 $\pm$ 1.928 & 17.111 $\pm$ 2.684 & 13.846 $\pm$ 4.316 \\
     & GUNet & 13.320 $\pm$ 0.924 & 10.726 $\pm$ 1.154 & 18.103 $\pm$ 0.675 & 9.814 $\pm$ 1.281 \\
    \midrule
    \multirow{4}{*}{$C_L$} & MLP & 0.211 $\pm$ 0.028 & 0.199 $\pm$ 0.031 & 0.621 $\pm$ 0.191 & 0.413 $\pm$ 0.096 \\
     & GraphSAGE & \textbf{0.148 $\pm$ 0.026} &  \textbf{0.150 $\pm$ 0.011} & 0.433 $\pm$ 0.060 & \textbf{0.254 $\pm$ 0.052} \\
     & PointNet & 0.197 $\pm$ 0.028 & 0.200 $\pm$ 0.047 & \textbf{0.384 $\pm$ 0.040} & 0.442 $\pm$ 0.107 \\
     & GUNet & 0.168 $\pm$ 0.023 & 0.150 $\pm$ 0.013 & 0.466 $\pm$ 0.118 & 0.376 $\pm$ 0.047 \\
    \midrule
    \multirow{4}{*}{$\rho_D$} & MLP & \textbf{0.250 $\pm$ 0.094} & 0.248 $\pm$ 0.064 & 0.157 $\pm$ 0.143 & 0.347 $\pm$ 0.222 \\
     & GraphSAGE & 0.194 $\pm$ 0.067 & \textbf{0.254 $\pm$ 0.071} & 0.039 $\pm$ 0.046 & 0.525 $\pm$ 0.088 \\
     & PointNet & 0.074 $\pm$ 0.063 & 0.048 $\pm$ 0.077 & 0.115 $\pm$ 0.148 & 0.089 $\pm$ 0.278 \\
     & GUNet & 0.092 $\pm$ 0.052 & 0.074 $\pm$ 0.021 & \textbf{0.192 $\pm$ 0.130} & \textbf{0.552 $\pm$ 0.056} \\
    \midrule
    \multirow{4}{*}{$\rho_L$} & MLP & 0.993 $\pm$ 0.002 & 0.993 $\pm$ 0.002 & 0.958 $\pm$ 0.022 & 0.957 $\pm$ 0.036 \\
     & GraphSAGE & \textbf{0.996 $\pm$ 0.001} & \textbf{0.996 $\pm$ 0.001} & 0.971 $\pm$ 0.018 & \textbf{0.989 $\pm$ 0.002} \\
     & PointNet & 0.992 $\pm$ 0.002 & 0.992 $\pm$ 0.002 & \textbf{0.981 $\pm$ 0.006} & 0.978 $\pm$ 0.003 \\
     & GUNet & 0.995 $\pm$ 0.001 & 0.994 $\pm$ 0.0003 & 0.964 $\pm$ 0.016 & 0.982 $\pm$ 0.007 \\
    \bottomrule
  \end{tabular}
\end{table}

\end{document}